\documentclass[12pt]{iopart}
\expandafter\let\csname equation*\endcsname\relax
\expandafter\let\csname endequation*\endcsname\relax
\usepackage[utf8]{inputenc} % allow utf-8 input
\usepackage[T1]{fontenc}    % use 8-bit T1 fonts
\usepackage{url}            % simple URL typesetting
\usepackage{booktabs}       % professional-quality tables
\usepackage{amsfonts}       % blackboard math symbols
\usepackage{nicefrac}       % compact symbols for 1/2, etc.
\usepackage{microtype}      % microtypographycvsm
\usepackage{graphicx}
\usepackage{doi}
\usepackage{caption}
\usepackage{adjustbox}
\usepackage{subcaption}
\usepackage{xcolor}
\usepackage{tcolorbox}
\usepackage{makecell}
\usepackage{array}
\usepackage{xfrac}
\usepackage{cite}           % For numerical citations
\usepackage{hyperref}
\usepackage{tikz}
\usetikzlibrary{positioning, arrows.meta, shapes.geometric, fit, backgrounds}

\usepackage{amsmath}
\usepackage{amssymb}
\usepackage{mathtools}
\usepackage{cancel}

\usepackage{nkj}

\usepackage[thinc]{esdiff}

\usepackage{amsmath,amsfonts,bm}

\def\eqref#1{equation~\ref{#1}}
\def\1{\bm{1}}

\def\rvepsilon{{\boldsymbol{\varepsilon}}}

\def\rvu{{\mathbf{i}}}

\def\rvs{{\mathbf{s}}}

\def\rvu{{\mathbf{u}}}
\def\rvv{{\mathbf{v}}}
\def\rvw{{\mathbf{w}}}
\def\rvx{{\mathbf{x}}}

\def\mI{{\bm{I}}}
\def\mJ{{\bm{J}}}

\DeclareMathAlphabet{\mathsfit}{\encodingdefault}{\sfdefault}{m}{sl}
\SetMathAlphabet{\mathsfit}{bold}{\encodingdefault}{\sfdefault}{bx}{n}

\newcommand{\E}{\mathbb{E}}

\newcommand{\R}{\mathbb{R}}

\newcommand*{\dv}{\mathrm{d}}

\newcommand{\rebuttal}[1]{#1}

\begin{document}

\title[TV/SNR Diffusion]{Disentangling Total-Variance and Signal-to-Noise-Ratio Improves Diffusion Models}

\author{
Khaled Kahouli$^{1,2,*}$,
Winfried Ripken$^{1,2}$,
Stefan Gugler$^{1,2}$,
Oliver T. Unke$^{3}$,
Klaus-Robert M\"uller$^{1,2,3,4,5,*}$
and Shinichi Nakajima$^{1,2,6,*}$
}

\address{$^1$ BIFOLD – Berlin Institute for the Foundations of Learning and Data, Berlin, Germany}
\address{$^2$ Machine Learning Group, Technische Universit\"at Berlin, Berlin, Germany}
\address{$^3$ Google DeepMind}
\address{$^4$ Department of Artificial Intelligence, Korea University, Seoul, Republic of Korea}
\address{$^5$ Max-Planck Institute for Informatics, Saarbrücken, Germany}
\address{$^6$ RIKEN Center for Advanced Intelligence Project, Tokyo, Japan}
\address{$^*$ Authors to whom any correspondence should be addressed.}

\ead{khaled.kahouli@tu-berlin.de, klaus-robert.mueller@tu-berlin.de, nakajima@tu-berlin.de}

\begin{abstract}
Diffusion models have been widely adopted as a powerful generative framework across numerous scientific domains, yet their inherently iterative sampling process remains a significant computational bottleneck. 
While this is typically mitigated by reducing the number of reverse diffusion steps, sample quality becomes highly sensitive to the chosen noise schedule, i.e.~the manner in which noise is introduced and the signal is reduced at each step.
Although prior work has improved on the original variance-preserving and variance-exploding schedules, these approaches passively adjust the total variance, without direct independent control over it. 
In this work, we propose a novel total-variance/signal-to-noise-ratio disentangled (TV/SNR) framework, where TV and SNR can be controlled independently. 
Our approach reveals that schedules with exponentially exploding TV can be consistently improved by adopting a constant TV schedule while preserving the underlying SNR schedule. 
Furthermore, we demonstrate that generalizing the SNR schedule of optimal transport flow matching significantly enhances generation efficiency. 
Crucially, these gains remain consistent across various reverse diffusion solvers and \rebuttal{reduce the search space to two independent and interpretable parameters}, enabling fast and robust sampling for complex scientific applications, such as molecular structure generation.
\end{abstract}

\vspace{2pc}

\section{Introduction}

\begin{figure*}[t]
    \centering
    \includegraphics[width=\textwidth]{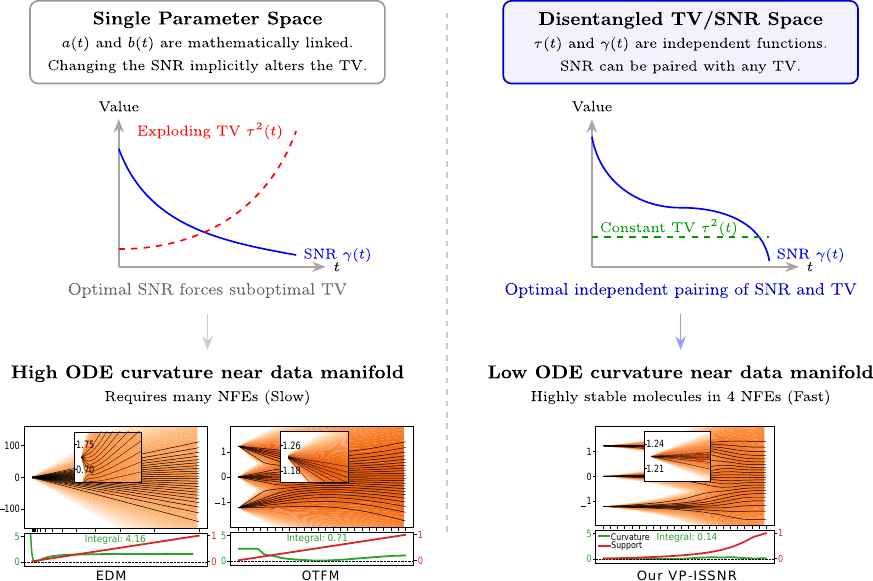}
   \caption{\textbf{Overview of our TV/SNR disentangled framework.}
    \textbf{Top:} In standard diffusion models (\textbf{left}), the signal parameter $a(t)$ and the noise parameter $b(t)$ from equation~\ref{eq:PerturbationKernel} are mathematically coupled. Consequently, optimizing the signal-to-noise-ratio (SNR) schedule often forces a suboptimal Total Variance (TV), as defined in equation~\ref{eq:snr-tv}, inducing high trajectory curvature. Our disentangled framework (\textbf{right}) breaks this dependency by controlling the diffusion process via two separate, independent schedules. This enables the design of an optimal Inverse Sigmoid SNR (ISSNR) schedule that can be independently paired with a constant TV schedule, resulting in our VP-ISSNR.
    \textbf{Bottom:} Numerical analysis of the simulated reverse ODE trajectories for a toy problem confirms this mechanism. Established methods, such as EDM (Variance-Exploding) and OTFM (Variance-Modulated), exhibit high trajectory curvature (green curves) near the data manifold ($t \to 0$), where the support of the marginal distribution (red curves) is small. In contrast, our decoupled VP-ISSNR schedule achieves significantly lower trajectory curvature (total integrated curvature is shown inside the plots). This mitigates discretization errors and enables the generation of stable molecular structures in as few as 4 NFEs. \rebuttal{A two-dimensional example with a mixture of 12 Gaussians forming two rings is given in figure~\ref{fig:toy_example_2d}, and shows the same trends.}}
    % }
    \vspace{-3mm}
    \label{fig:overview}
\end{figure*}

With the development of diffusion models~\cite{diff_mod_sohl,DDPM_Ho,song2020score,song2021score}, generative modeling has witnessed great progress in recent years. Beyond their capabilities in domains such as image synthesis \cite{diffbeatsgan, nichol2022glide,rombach2022high, DiT, SiT} and audio generation \cite{kong2021diffwave, chen2020wavegrad, audioLDM}, they are also increasingly used in the physical sciences. In computational chemistry, for example, diffusion models have become an efficient alternative for tasks such as molecule generation~\cite{gebauer2022inverse,edm, diffbridges, mdm, geoldm, moldiff, vignac2023midi, eqgat-diff}, structural relaxation \cite{kahouli2024morered}, conformer search~\cite{geodiff}, and amortized sampling from energy functions~\cite{zhang2021path, akhound2024iterated, vargas2023denoising, kahouli2025controlvariatescorematching}.

Despite their superior generative capabilities, diffusion models are computationally expensive during inference. Generating samples requires iteratively solving a reverse stochastic process, where a learned score function, parameterized by a deep neural network, must be evaluated at each step. Consequently, reducing the sampling cost without degrading sample quality requires solving the reverse process with minimal score function evaluations while avoiding high discretization errors. To address this challenge, previous work has explored both training-based improvements~\cite{imp_ddpm, watson2022learning, salimans2022progressive} and advanced sampling techniques, including higher-order solvers that replace conventional first-order methods~\cite{dpm-solver, liu2022pseudo, dockhorn2022genie, jolicoeur-martineau2022gotta, lu_dpm-solver_2023, zhang2023fast, unipc}. More recently, the field has seen significant advances through distillation and one-shot generation techniques, such as Consistency Models and Flow Maps ~\cite{cm, kim2024consistency, lu2025simplifying, zhou2025inductive, frans2025one, yang2024consistencyflowmatchingdefining, boffi2025buildconsistencymodellearning, mean_flow}, which attempt to map noise directly to the data in a single or very few steps. Although these distillation methods shift the solver inside the model and require expensive model retraining to achieve few-step generation, optimizing the underlying noise schedule represents an orthogonal, training-free approach that inherently smooths the sampling trajectory for any given solver.

\rebuttal{Previous work has also studied the impact of the noise schedule on performance \cite{chen2023importance, lin2024common}}. The seminal work of \cite{diff_edm} introduced EDM, which modifies the reverse process to follow straighter trajectories and employs a non-uniform time grid to reduce discretization errors. Building on the concept of straight paths in probability flows, conditional flow matching was later proposed as a simulation-free approach to train continuous normalizing flows by fixing the forward conditional path~\cite{liu2023flow, lipman2023flow}, achieving promising results across different data modalities, including images~\cite{SiT} and molecules~\cite{klein2023equivariant, song2023equivariant, irwin2024efficient, et-flow, frank2025sampling}. When defining these conditional paths to be Gaussian, flow models and diffusion models can be viewed as equivalent formulations of the same underlying generative process. The primary distinction lies in the training objective; diffusion models estimate the score function, whereas conditional flow matching targets the corresponding velocity field, which is a function of the score. Of particular interest is the special case of using optimal transport to define the conditional probability path as a linear interpolation, imposing straighter ODE trajectories \cite{lipman2023flow, albergo2023building, pmlr-v202-pooladian23a, tong2024improvingflow}. This case can be viewed as a specific parameterization of diffusion models governed by a distinct variance-modulating (VM) noise schedule, which we will refer to as Optimal Transport Flow Matching (OTFM).

Traditionally, diffusion processes are parameterized by a time-dependent perturbation kernel that scales the data signal and injects Gaussian noise. The noise schedule dictates how these signal and noise components are balanced over time and can be broadly categorized into variance-preserving (VP)~\cite{DDPM_Ho} and variance-exploding (VE)~\cite{song2019score, song2021score} variants. For example, EDM relies on a specific VE schedule to achieve good performance in image generation, while OTFM uses a specific VM schedule. 
Crucially, these prior works mathematically couple the total variance (TV) and the signal-to-noise-ratio (SNR), and passively adjust them together, lacking direct independent control over them, as shown in table~\ref{tab:ExistingTVSNR}. 
In this paper, we provide a more systematic approach to noise scheduling by introducing a total-variance/signal-to-noise-ratio (TV/SNR) disentangled framework. Unlike previous approaches, our new formulation explicitly decouples these two parameters into separate functions, as conceptually illustrated in figure~\ref{fig:overview}. This ensures they can be controlled independently. The resulting reverse process is subsequently solved using the corresponding ODE/SDE to generate samples.

Our empirical results indicate that fast sample generation of scheduling methods with exponentially growing TV can be significantly improved by modifying the TV schedule to instead follow a VP trajectory, while keeping the SNR schedule unchanged {(see figure~\ref{fig:ComparisonMolecule} top-right)}.
Based on this observation, we hypothesize that a constant TV schedule is already a sufficiently effective choice. Therefore, instead of fully exploring all possibilities of more sophisticated TV schedules, we focus on optimizing the SNR schedule. Specifically, we propose an exponential inverse sigmoid function for scheduling the SNR (ISSNR), which allows rapid SNR decay at both the beginning and the end of the diffusion process.
Our ISSNR schedule acts as a generalization of OTFM and reduces the required Number of Function Evaluations (NFEs) in molecular structure generation \rebuttal{without heavy modality-specific tuning}. For computational chemistry, minimizing the NFEs of the diffusion model translates into practical computational savings for high-throughput applications. We observe a similar trend in image generation, where our \rebuttal{VP-ISSNR} schedule\rebuttal{, with its two parameters tuned,} performs comparably to heavily optimized state-of-the-art methods. To offer insights into the mechanics of this fast sample generation, we provide a numerical analysis of the reverse ODE trajectories in section~\ref{sec:Discussion}. This analysis allows us to attribute the effectiveness of our approach to the resulting low curvature of the sampling paths close to the data space, coupled with the sufficiently large support of the diffused marginal distribution in regions with high curvature. \rebuttal{Throughout, we consider the \emph{training-free} setting, where a diffusion model is trained once, and all schedules are compared at inference-time for that same fixed model.}

In summary, the primary contribution of this work is the introduction of the novel {\em TV/SNR disentangled framework} that allows for independent control of the total variance and signal-to-noise-ratio, while unifying many existing diffusion and flow models within a single formulation (see table~\ref{tab:ExistingTVSNR}). Furthermore, we empirically demonstrate using different data modalities, images and molecules, that common VE schedules can be significantly improved by transforming them into their VP counterparts within our framework. Building on this insight, we propose the \rebuttal{VP-ISSNR} schedule, which shows improved sampling performance across various reverse diffusion solvers and different applications.

\section{Theory and methods}

\subsection{Diffusion models}

\begin{table*}[t]
    %\small
    \centering
    \caption{TV $\tau(t)$ and SNR $\gamma(t)$ schedules corresponding to commonly used diffusion processes within our TV/SNR framework. See~\ref{sec:A.SCLSNRExpressionExisting} for the derivations.}
    \renewcommand{\arraystretch}{2}
    \begin{adjustbox}{max width=\textwidth}
    \begin{tabular}{@{}lccc@{}}
        \br
        \textbf{Method} & $\tau^2(t)$ & $\gamma^2(t)$ & time grid \\
        \mr
        \makecell[l]{\textbf{SMLD} \scriptsize{\cite{song2019score}}} & $1 + \sigma_\mathrm{min}^{2}\left( \frac{\sigma_\mathrm{max}}{\sigma_\mathrm{min}}\right)^{2t}$  & $\sigma_\mathrm{min}^{-2}\left( \frac{\sigma_\mathrm{min}}{\sigma_\mathrm{max}}\right)^{2t}$ & uniform  \\
        \makecell[l]{\textbf{EDM} \scriptsize{\cite{diff_edm}}}  & $1 + \sigma^2(t)$   & $\sigma^{-2}(t)$ & eq.\ref{eq:EDMTimeGrid}  \\
        \makecell[l]{\textbf{EDM-UT} \\ \scriptsize{}}   & $1 + \left( \sigma_{\text{max}}^{\frac{1}{\rho}} + (1 - t)  \left( \sigma_{\text{min}}^{\frac{1}{\rho}} - \sigma_{\text{max}}^{\frac{1}{\rho}} \right) \right)^{2 \rho}$   & $\left( \sigma_{\text{max}}^{\frac{1}{\rho}} + (1 - t)  \left( \sigma_{\text{min}}^{\frac{1}{\rho}} - \sigma_{\text{max}}^{\frac{1}{\rho}} \right) \right)^{- 2 \rho}$ & uniform \\
        \makecell[l]{\textbf{OTFM} \scriptsize{\cite{lipman2023flow}}} &
        $1 - 2t (1 - t)$ & $(\frac{1}{t} - 1)^2$ & uniform \\
        \makecell[l]{\textbf{DDPM-linear}  \scriptsize{\cite{DDPM_Ho}}} & $1$   & $\left( e^{ \frac{1}{2} t^2 (\beta_\mathrm{max} - \beta_\mathrm{min}) + t \beta_\mathrm{min}} -1 \right)^{-1}$  & uniform \\
        \makecell[l]{\textbf{DDPM-cos} \scriptsize{\cite{imp_ddpm}}} & $1$   & $ \left( \left( \frac{\cos\left(\frac{s}{1 + s} \frac{\pi}{2}\right)}{\cos\left(\frac{t^\nu + s}{1 + s} \frac{\pi}{2}\right)} \right)^2 -1 \right)^{-1}$   & uniform \\
        \hline
        \makecell[l]{\textbf{VP-SMLD} \scriptsize{(Ours)}} & 1   & $\sigma_\mathrm{min}^{-2}\left( \frac{\sigma_\mathrm{min}}{\sigma_\mathrm{max}}\right)^{2t}$ & uniform  \\
        \makecell[l]{\textbf{VP-EDM-UT} \scriptsize{(Ours)}} & 1   & $\left( \sigma_{\text{max}}^{\frac{1}{\rho}} + (1 - t) \left( \sigma_{\text{min}}^{\frac{1}{\rho}} - \sigma_{\text{max}}^{\frac{1}{\rho}} \right) \right)^{- 2 \rho}$ & uniform \\
        \makecell[l]{\textbf{VP-OTFM}  \scriptsize{(Ours)}} & 1 & $(\frac{1-t}{t})^2$ & uniform \\
        \makecell[l]{\textbf{VP-ISSNR } \scriptsize{(Ours)}} & $1$   & $ \left(\frac{1-t }{t}\right)^{2\eta} \exp(2 \kappa)   $ & uniform \\
        \br
    \end{tabular}
    \end{adjustbox}
    \label{tab:ExistingTVSNR}
\end{table*}

Let $p_\mathrm{data}(\rvx)$ be the data distribution of interest defined on the support $\mathcal{X} \subseteq \R^d$.
We assume that the data are standardized, i.e., $\mathbb{E}_{p_\mathrm{data}}[\rvx] = \bfzero$ and
$\mathrm{Var}_{p_\mathrm{data}}[\rvx] = \bfone$.
Let $\{\rvx(t)\}_{t \in [0,1]}$ be the forward diffusion process describing the stochastic flow of the marginal probability density $\{p_t(\rvx)\}_{t \in [0,1]}$ resulting from the iterative injection of Gaussian noise (starting from $p_0(\rvx) := p_\mathrm{data}(\rvx)$ and ending at a tractable Gaussian prior $p_1(\rvx)$) using a standard Wiener process. This forward diffusion process can be modeled with a time-dependent perturbation kernel
    \begin{align}
        p(\rvx(t)|\rvx(0)) = \mathcal{N} \left(\rvx(t); a(t) \rvx(0), b^2(t) \mI \right),
        \label{eq:PerturbationKernel}
    \end{align}
where $t \in [0, 1]$, $a(t)$ controls the signal strength, $b(t)$ controls the noise level at each diffusion step, and $\bfI$ denotes the identity matrix. The scaling terms $a(t)$ and $b(t)$ are smooth non-negative functions satisfying specific boundary conditions: for small $\Delta t$, $a(\Delta t) \approx 1$ and $b(\Delta t) \approx 0$, ensuring that $\rvx(\Delta t) \approx \rvx(0)$. Additionally, $a(1) \ll b(1)$ ensures that $\rvx(1)$ follows an isotropic Gaussian distribution. In this work, we define the SNR as $\gamma(t)=a(t)/b(t)$~\cite{kingma2021}, and the TV as $\tau^2(t)=a^2(t)+b^2(t)$. 

Once we define this perturbation kernel, the marginal density path for $t \in [0, 1]$ is fixed as
 \begin{align}
    p_t(\rvx(t))  =     \int p(\rvx(t)|\rvx(0))p_\mathrm{data}(\rvx(0))\ \dv\rvx(0)\,,
    \label{eq:GenGaussPerturbKernel}
 \end{align}
 and the corresponding SDE is
 \begin{align}
    \dv \rvx & =  \rvx f(t)   \dv t + {g(t)}  \dv \rvw\,, \quad \text{where}
    \label{eq:GenForwSDE}\\
    f(t) &=     \frac{\dot{a}(t)}{a(t)} \:\: \text{and} \:\:
    g(t) =\sqrt{2 \frac{ b(t)}{ a(t)} \left(a(t) \dot{b}(t) - \dot{a}(t) b(t)\right)}\,.
    \notag
 \end{align}
  (see \ref{app:KerneltoSDE} for the full derivation). Here, $\rvw$ is the standard Wiener process and $f(t)$ and $g(t)$ are the drift and diffusion coefficients, respectively. We use Newton's notation for time derivatives, e.g., $\dot{a}(t) = \frac{\dv}{\dv t} a(t)$.

The reverse SDE for equation~\ref{eq:GenForwSDE} is given by
 \begin{align}
    \dv \tilde{\rvx} =  \rebuttal{-}  \left[ \tilde{\rvx} f(t)   - \frac{1+ \lambda^2}{2} {g(t)}^2 \nabla_{\tilde{\rvx}} \log p_t(\tilde{\rvx})\right] \dv \tilde{t} + \lambda {g(t)} \, \dv \tilde{\rvw}
    \label{eq:GenRevSDE}
 \end{align}
 where the tilde indicates time-reversal, i.e., $\tilde{t}=1-t$, and $\nabla_{\rvx} \log p_t(\rvx)$ is the score function of the marginal density at time $t$.
Equation~\ref{eq:GenRevSDE} describes the general reverse SDE for $\lambda \geq 0$, including the exact time-reversal when $\lambda=1$, and variants with less ($\lambda < 1$) or more ($\lambda > 1$) stochasticity that maintain the same marginal distribution. The extreme case where $\lambda = 0$ corresponds to the probability flow ODE \cite{zhang2021diffusion}.

Training diffusion models amounts to approximating the unknown score function of the marginal density in equation~\ref{eq:GenRevSDE} for $t \in [0, 1]$
by a parametrized neural network, $\rvs_{\theta}(\rvx(t),t) \approx \nabla_{\rvx} \log p_t(\rvx)$.
The parameters can be fitted by a re-weighted version of denoising score-matching (DSM) \cite{vincent,song2019score,song2021score}:
\begin{align}
    \theta^*
    & = \argmin_{\theta}
     \E_{p_{\mathrm{data}(\rvx(0))}} \E_{p(\rvx(t)|\rvx(0))} \bigl[ \| \rvepsilon_t - \rvepsilon_{\theta}(\rvx(t), t)\|^2   \bigr],
     \label{eq:DSM}
\end{align}
where $\rvepsilon_t \!=\! \frac{\rvx(t) - a(t) \rvx(0)}{b(t)}$ and
$\rvepsilon_{\theta}(\rvx(t), t) \!=\!- b(t) \rvs_{\theta}(\rvx(t),t)$.
The expectation over $p_\mathrm{data}(\rvx(0))$ is approximated by an average over training data samples, whereas the expectation over $p(\rvx(t)|\rvx(0))$, which corresponds to simulating the forward process, is numerically performed by applying the perturbation kernel in equation~\ref{eq:PerturbationKernel} to $\rvx(0)$.

\subsection{Improving sampling efficiency}
In addition to the distinction between training-free and training-based methods, approaches for improving the sampling efficiency of diffusion models can be roughly categorized into one of two major branches:

\paragraph{Noise scheduling optimization:} The two most popular noising schedules are variance-exploding (VE) \cite{song2019score, song2021score}
and variance-preserving (VP) \cite{DDPM_Ho} schedules.
The original VE schedule uses $a(t)=1$ and $b^2(t) = \sigma^2(t)$ to define the perturbation kernel in equation~\ref{eq:PerturbationKernel}, and controls $\sigma(t)$, e.g., by setting $\sigma^2(t) =  \sigma_\mathrm{min}^{2}\left( \frac{\sigma_\mathrm{max}}{\sigma_\mathrm{min}}\right)^{2t}$, where $\sigma_\mathrm{min}$ and $\sigma_\mathrm{max}$ are the minimum and maximum noise levels, respectively.
In contrast, the VP schedule uses $a(t) = \sqrt{\bar{\alpha}(t)}$ and $b^2(t) = 1 - a^2(t)$,
and controls $\bar{\alpha}(t)$, e.g., by
$\bar{\alpha}(t) = e^{- \beta_{\mathrm{min}} t - \frac{1}{2}(\beta_{\mathrm{max}} - \beta_{\mathrm{min}}) t^2}$, where $\beta_{\mathrm{min}}$ and $ \beta_{\mathrm{max}}$ control the start and end points of the schedule.

Karras~\etal~\cite{diff_edm} introduced the EDM framework, which optimizes the reverse process to minimize the number of function evaluations (NFEs) while preserving sample quality. They design the noising schedule using a scale factor $s(t)$ and the noise level $\sigma(t)$, defined as:
\begin{align}
    p(\rvx(t)|\rvx(0)) = \mathcal{N} \left(\rvx(t); s(t) \rvx(0), s^2(t) \sigma^2(t) \mI \right),
    \label{eq:PerturbationKernelEDM}
\end{align}
 and adopt a non-uniform time grid discretization over $N$ steps, $t_i = \sigma^{-1}(\sigma_i)$, where
 \begin{align}
    \sigma_{i < N} =
    \left(
    \sigma_{\mathrm{max}}^{1/\rho}
    + \frac{i}{N-1} \left(\sigma_{\mathrm{min}}^{1/\rho} - \sigma_{\mathrm{max}}^{1/\rho}\right)
    \right)^{\rho}
    \label{eq:EDMTimeGrid}
 \end{align}
and $\sigma_N=0$ for $\rho = 7$.
They argue that setting $s(t) = 1$ and $\sigma(t) = t$, which coincides with the DDIM sampler \cite{ddim},
leads to straight flow trajectories. Specifically, they argue that when $\lambda = 0$ in equation~\ref{eq:GenRevSDE}, this schedule minimizes the discretization error under specific assumptions.

\paragraph{High-order solvers and distillation approaches:}
Orthogonal to exploring optimized noise schedules, sampling efficiency can also be improved by adopting higher-order solvers, e.g., Heun's method \cite{diff_edm} or the DPM-solver \cite{dpm-solver, lu_dpm-solver_2023}. Alternatively, distillation and one-shot approaches, such as Consistency Models and Flow Maps ~\cite{cm, kim2024consistency, lu2025simplifying, zhou2025inductive, frans2025one, yang2024consistencyflowmatchingdefining, boffi2025buildconsistencymodellearning, mean_flow}, can be used to learn an efficient yet highly non-linear solver within the model, e.g. by distilling diffusion or flow models.

In this work, we focus exclusively on exploring optimal noise scheduling, and show that better schedules improve the performance across different choices of reverse diffusion solvers, which can be regarded as an orthogonal approach to other techniques \cite{diff_edm, dpm-solver, cm, kim2024consistency, lu2025simplifying, zhou2025inductive, frans2025one, yang2024consistencyflowmatchingdefining, boffi2025buildconsistencymodellearning, mean_flow} for improving diffusion models.

\subsection{Our TV/SNR diffusion framework}

\begin{figure*}[t]
 \centering
 \includegraphics[width=.32\textwidth]{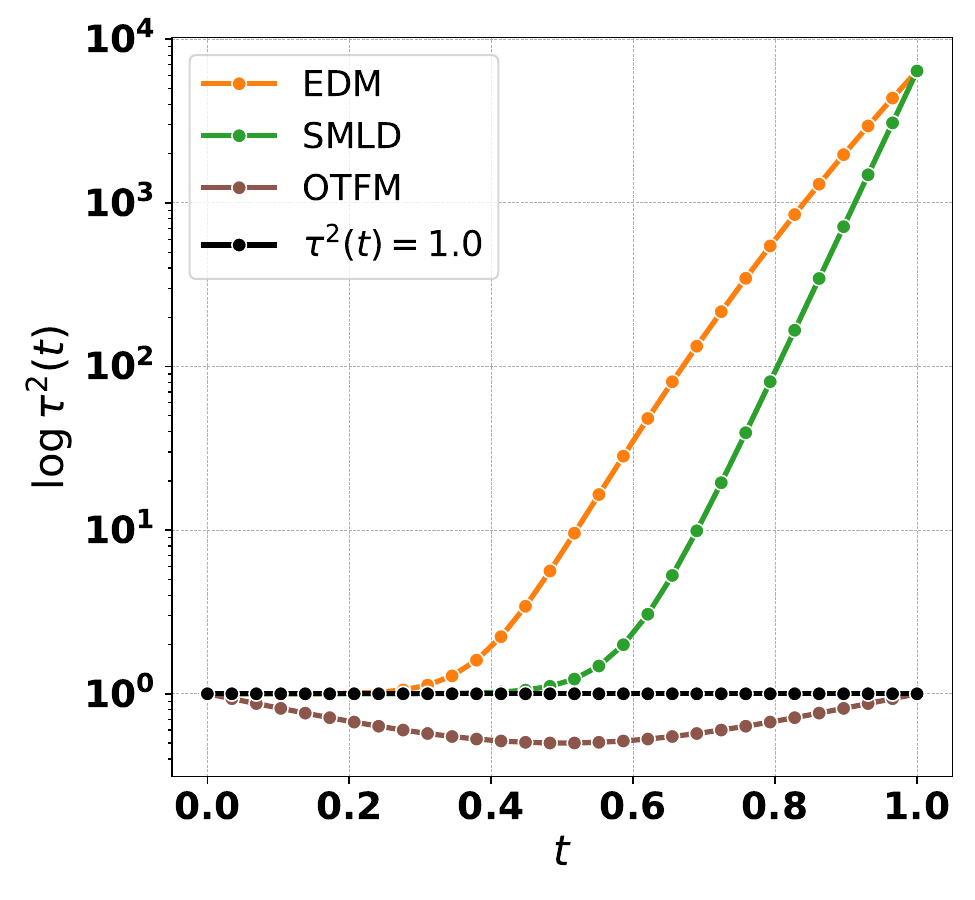}
 \includegraphics[width=.32\textwidth]{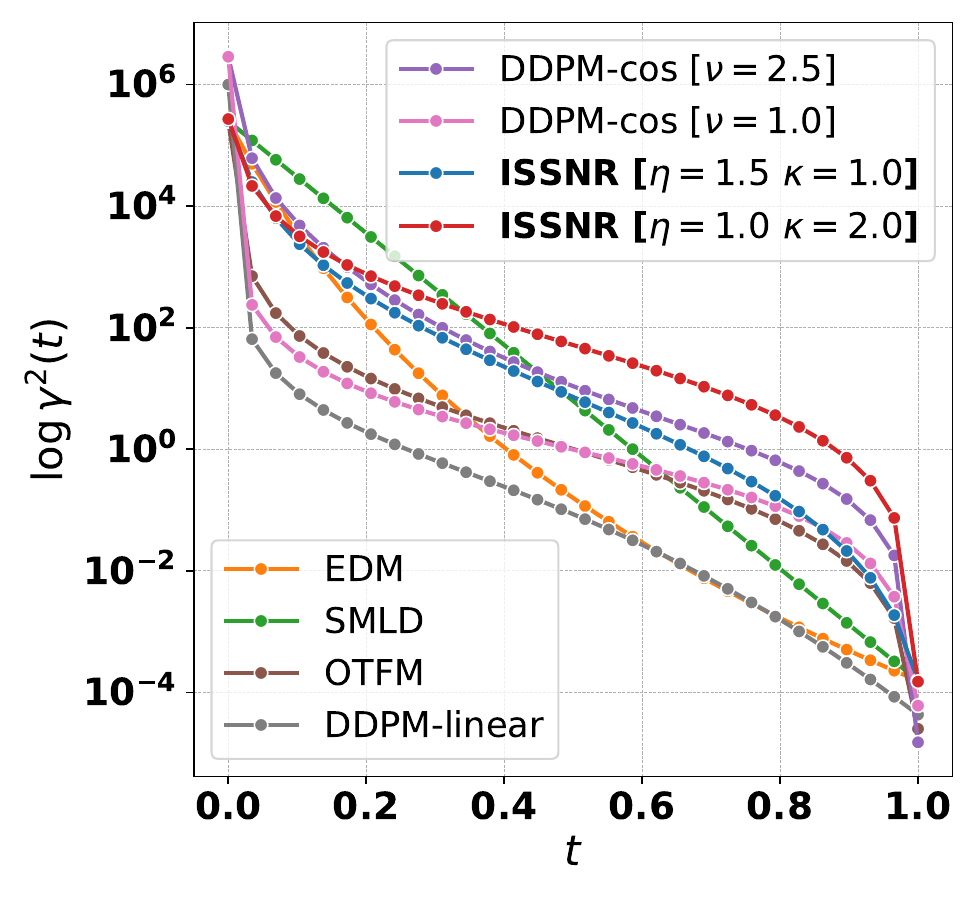}
 \includegraphics[width=.32\textwidth]{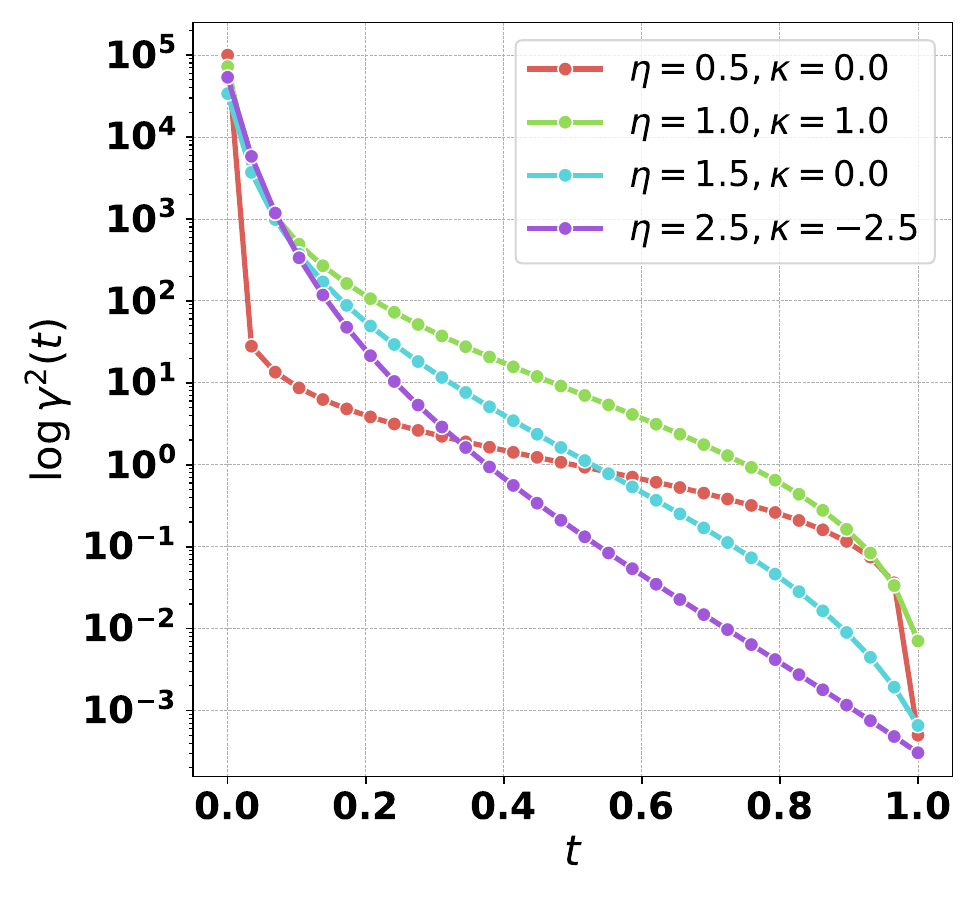}
 \vspace{-3mm}
 \caption{
  TV (\textbf{Left}) and SNR schedules (\textbf{middle}) of our ISSNR together with different established scheduling methods (see table~\ref{tab:ExistingTVSNR} for the corresponding analytic expressions). \textbf{Right:} Our proposed inverse sigmoid SNR (ISSNR) schedule.
 }
 \vspace{-3mm}
 \label{fig:ExistingTVSNR}
\end{figure*}

Although they improve performance, \rebuttal{a practical limitation of the design space of} existing diffusion formulations is the coupling of the TV and the SNR. As shown in the previous section, standard methods parameterize the forward process via a signal scale $a(t)$ and a noise scale $b(t)$. Consequently, any adjustment to these scales implicitly changes the SNR, $\gamma(t) = a(t)/b(t)$, and the TV, $\tau^2(t) = a^2(t) + b^2(t)$, simultaneously. This dependency complicates, and restricts the design space of noise schedules.

We propose a unifying framework for noise scheduling, where the total variance (TV) and the signal-to-noise-ratio (SNR) are explicitly parameterized and controlled independently. Within this framework, we cast established schedules into a shared formulation and apply simple modifications. We empirically show that for common existing schedules where the TV grows exponentially, their constant TV (i.e., VP) variants consistently improve performance. Finally, we propose the ISSNR (inverse sigmoid SNR) schedule, an SNR scheduling strategy based on the exponential of the inverse sigmoid function, to further enhance sampling stability and efficiency.

\subsubsection{Disentangled scheduling framework}

To achieve a clear separation of these variables, we reformulate the perturbation kernel in equation~\ref{eq:PerturbationKernel} as
 \begin{align}
  p(\rvx(t)|\rvx(0))
  &=
  \mathcal{N}\left(\rvx(t);  \sqrt{\frac{\tau^2(t) \gamma^2(t)}{1 + \gamma^2(t)} }\rvx(0) ,  \frac{\tau^2(t)}{1 + \gamma^2(t)} \mI \right),
  \label{eq:ProposedMarginal}
\end{align}
where $\tau(\cdot): [0, 1] \mapsto \mathbb{R}_{++}$ is a TV controlling function, and $\gamma(\cdot): [0, 1] \mapsto \mathbb{R}_{++}$ is an SNR controlling function. $\tau(t)$ can be an arbitrary positive function, while $\gamma(t)$ is monotonically decreasing from $\gamma(0) = \gamma_{\mathrm{max}}< \infty$ to $\gamma(1) = \gamma_{\mathrm{min}}>0$. By comparing equation~\ref{eq:ProposedMarginal} to equation~\ref{eq:PerturbationKernel}, it is straightforward to confirm that
\begin{equation}
  \mathrm{TV} =   a^2(t) + b^2(t) =\tau^2(t) \quad \text{and} \quad
  \mathrm{SNR} =   \frac{ a(t)}{ b(t)} =\gamma(t)\,,
  \label{eq:snr-tv}
\end{equation}
This \rebuttal{shows} that our reformulation achieves \rebuttal{an exact decoupling of the two coordinates}, where $\tau(t)$ does not affect the SNR and $\gamma(t)$ does not affect the TV, respectively, and they can be controlled independently.
The forward and reverse SDEs corresponding to the kernel in equation~\ref{eq:ProposedMarginal} are given by equations~\ref{eq:GenForwSDE} and \ref{eq:GenRevSDE}, respectively, with the drift and diffusion coefficients
 \begin{equation}
  f(t) =  \frac{\dot{\tau}(t)}{\tau(t)} + \frac{\dot{\gamma}(t)}{\gamma(t) \left( 1 + \gamma^2(t) \right)}
  \quad \text{and} \quad
  g(t) = \sqrt{\frac{-2 \tau^2(t) \dot{\gamma}(t)}{\gamma(t) \left( 1 + \gamma^2(t) \right)}}
  \label{eq:f_and_g}
 \end{equation}
 (see \ref{app:SNR-SDE} for the full derivation). In practice, we work in the logarithmic scale for numerical stability. Specifically, we use the identity $\frac{\dv}{\dv t} \log \gamma(t) = \frac{\dot{\gamma}(t)}{\gamma(t)}$, and apply the same transformation to $\tau(t)$.
 
\subsubsection{\rebuttal{Relation to existing two-function parameterizations}}
\label{sec:Parameterizations}

\rebuttal{The forward process in equation~\ref{eq:PerturbationKernel} possesses two degrees of freedom. Existing literature typically parameterizes these using either the signal and noise coefficients $(a, b)$ from equation~\ref{eq:PerturbationKernel}, or the scale and noise-level pair $(s, \sigma)$ introduced in EDM~\cite{diff_edm} (equation~\ref{eq:PerturbationKernelEDM}). For standardized data, matching the mean and variance of the perturbation kernels yields $a = s$ and $b = s\sigma$. Consequently, our proposed TV and SNR coordinates $(\tau, \gamma)$ can be mapped to the EDM variables:}
\rebuttal{
\begin{equation}
    \gamma = \frac{1}{\sigma}
    \quad \text{and} \quad
    \tau = s \sqrt{1 + \sigma^2},
    \label{eq:EDMtoTVSNR}
\end{equation}
with the inverse transformation given by $\sigma = \gamma^{-1}$ and $s = \tau \gamma / \sqrt{1+\gamma^2}$. All three parameterizations are mathematically equivalent. 
}

\rebuttal{The primary advantage of the $(\tau, \gamma)$ parameterization is the explicit decoupling of total variance and SNR. Similar to the $(a, b)$ system, in the $(s, \sigma)$ system, equation~\ref{eq:EDMtoTVSNR} demonstrates that $\tau$ depends on both $s$ and $\sigma$. Thus, modifying the noise level $\sigma(t)$ to adjust the SNR inadvertently alters the total variance unless $s(t)$ is simultaneously compensated.
By elevating TV and SNR to independent coordinates, we expose the direct relationship between the TV schedule and discretization error (section~\ref{sec:LTE}) that remains implicit in prior formulations.}

\rebuttal{While Kingma~\etal~\cite{kingma2021} established the log-SNR as the central variable for optimizing continuous-time diffusion, their formulation is restricted to the VP DDPM setting, relying on a single degree of freedom. Our framework generalizes this by introducing TV as an independent coordinate within the broader SDE setting. Similarly, Karras~\etal~\cite{diff_edm} construct their sampling time grid around $\sigma$, but retain $s$ as the second function, leaving TV control implicit.}
\subsubsection{VP variants of established non-VP schedules}

For common established schedules, including the original linear VP schedule in DDPM \cite{DDPM_Ho,song2021score} and its cosine alternative \cite{imp_ddpm}, the original VE schedule in SMLD \cite{song2019score,song2021score}, EDM \cite{diff_edm}, and FM \cite{lipman2023flow},
figure~\ref{fig:ExistingTVSNR} (left and middle) shows their corresponding TV and SNR schedules (see table~\ref{tab:ExistingTVSNR} for analytic expressions, as well as \ref{sec:A.SCLSNRExpressionExisting} for the derivations).
Since EDM uses a non-uniform time grid, we also consider EDM with a uniform time grid (EDM-UT), where we incorporate the original non-uniform time grid into the TV/SNR schedules. Note that, although EDM and EDM-UT effectively use the same schedules and their respective ODEs are equivalent in the continuous time case, they perform differently when using numerical integration with discretization to solve the ODE/SDE.
SMLD, EDM, and EDM-UT all use a VE schedule, where the TV increases exponentially to a large value $\sigma_{\mathrm{max}}^2$ when $t \to 1$.
The TV schedule of the optimal transport flow matching (OTFM) is modulated, i.e., it is a variance-modulated (VM) schedule, although it does not grow exponentially, as depicted in figure~\ref{fig:ExistingTVSNR} (left).
In section~\ref{sec:Experiments}, we examine whether exploding or modulated TV schedules are essential for achieving good performance. To this end, we introduce their VP counterparts, VP-SMLD, VP-EDM-UT, and VP-OTFM, which use a constant TV schedule and the original SNR schedules (see table~\ref{tab:ExistingTVSNR}).

\subsubsection{Variance-preserving inverse sigmoid SNR (VP-ISSNR) schedule} 

Building on these insights, we propose to schedule TV and SNR with the following functions:
    \begin{align}
        \tau^2(t) &=1,
        \label{eq:ConstantTVScheduling}\\
        \gamma^2(t)
        &=  \exp \left(2 \eta \log \left( \frac{1}{t \, (t_{\mathrm{max}} - t_{\mathrm{min}}) + t_{\mathrm{min}}} -1 \right) + 2\kappa \right) \nonumber \\
        &=
            \left( \frac{1}{t \, (t_{\mathrm{max}} - t_{\mathrm{min}}) + t_{\mathrm{min}}} -1 \right)^{2 \eta}
            \!
            \exp( 2 \kappa).
        \label{eq:SigmoidSNRScheduling}
    \end{align}
    Namely, we set TV to be constant, and schedule SNR with the exponential of the inverse sigmoid function.
    The parameters $\eta > 0$ and $\kappa \in \mathbb{R}$ control the steepness and the offset of the inverse sigmoid function, respectively, and the two constants
    $0 < t_{\mathrm{min}} \approx 0$ and $ 1 > t_{\mathrm{max}} \approx 1$ adjust the starting $\gamma_{\mathrm{max}}$ and final $\gamma_{\mathrm{min}}$ SNR values as
    \begin{align}
        \gamma^2(0) &=  \exp \left(2 \eta \log \left( \frac{1}{t_{\mathrm{min}}} -1 \right) + 2 \kappa \right) = \gamma^2_{\mathrm{max}}, \\
        \gamma^2(1) &=  \exp \left(2 \eta \log \left( \frac{1}{t_{\mathrm{max}}} -1 \right) + 2 \kappa \right) = \gamma^2_{\mathrm{min}}.
    \end{align}
    The right panel in figure~\ref{fig:ExistingTVSNR} shows the inverse sigmoid function from equation~\ref{eq:SigmoidSNRScheduling} with different parameter choices.
    With this SNR function, we can allocate more steps to specific SNR levels using $\kappa$, while $\eta$ controls the relative emphasis on the most critical SNR levels compared to other regions of the diffusion process.

\paragraph{Relation to OTFM}

Using optimal transport (OT) to define the conditional probability path in flow matching (FM) \cite{lipman2023flow} results in a linear interpolation between the prior and the target data distribution,
\begin{align}
    p_t(\rvx_t|\rvx_0) = \mathcal{N}\left(\rvx(t); (1-t) \rvx(0) ,  ((1 - t) \sigma_\mathrm{min} + t)^2 \mI \right),
    \label{eq:OTFMKernel}
\end{align}
where $\sigma_\mathrm{min}$ is chosen to be sufficiently small, ensuring that the Gaussian distribution is concentrated around the target data point $x(0)$\footnote{Note that in \cite{lipman2023flow}, the time is reversed, with $\rvx(1)$ and $\rvx(0)$ corresponding to the samples in the target and latent domain, respectively. In this paper, we always define $t$ in the forward diffusion direction.}.
Consider a generalization of equation~\ref{eq:OTFMKernel}:
\begin{align}
    p_t(\rvx_t|\rvx_0) =  \mathcal{N}\left(\rvx(t); (1-t)^\eta \rvx(0) ,  t^{2\eta} \exp(-2\kappa) \mI \right).
    \label{eq:OTFMKernelGeneral}
\end{align}
Then, the corresponding TV and SNR schedules are
\begin{equation*}
\tau^2(t) = (1-t)^{2\eta} + t^{2\eta} \exp(-2\kappa) \quad \text{and} \quad
\gamma^2(t) =
\left(\frac{1}{t} - 1\right)^{2 \eta}
\exp( 2 \kappa)\,.
\end{equation*}
Comparing these expressions to our proposed ISSNR schedules in equations~\ref{eq:ConstantTVScheduling} and \ref{eq:SigmoidSNRScheduling},
we observe that our proposed SNR schedule is a generalization of the OTFM in equation~\ref{eq:OTFMKernelGeneral} with a constant TV schedule (setting $\eta=1.0$ and $\kappa=0$ recovers the SNR schedule of OTFM). The generalization allows for more expressive control over the generated probability flow.

\section{Results and discussion}
\label{sec:Experiments}

Before applying our framework to complex data modalities, we first establish an intuition for optimal scheduling by analyzing the curvature of ODE trajectories and the support of the marginal density in a controlled toy setting. We then demonstrate the significant advantages of our approach in the primary task of molecular structure generation, before evaluating its generalization to image generation.

\subsection{Intuition from trajectory curvature and marginal support}
\label{sec:Discussion}

To understand why specific schedules yield better sampling efficiency, we must examine the paths they generate. Karras~\etal~\cite{diff_edm} argue that if ODE trajectories follow straight paths, crude time discretization should not produce substantial errors, enabling fast sample generation. They further argue from a theoretical point of view that ODE trajectories of EDM are straight, by using Tweedie's formula \cite{Efron22011},
$ \nabla_{\rvx} \log p_t(\rvx(t)) =  \frac{\textcolor{black}{a(t)} \E[\rvx(0)|\rvx(t)] - \rvx(t)}{\textcolor{black}{b^2(t)}} $.
However, this argument assumes that the conditional expectation $\E[\rvx(0)|\rvx(t)]$ approximates the data point $\rvx(0)$ that $\rvx(t)$ reaches at time $t=0$ in the reverse ODE process. This does not necessarily hold due to the interaction between trajectories; an ODE trajectory generating a particular data point never crosses the trajectory of another data point, resulting in curved backward trajectories.

To demonstrate such trajectory interactions, figure~\ref{fig:overview} (Bottom) depicts ODE trajectories of  EDM, OTFM, and our VP-ISSNR for a toy problem where the data distribution $p_\mathrm{data}(\rvx(0))$ is a mixture of three delta peaks with equal weights (uniformly spaced at $x = 0, \pm\sqrt{\sfrac{3}{2}}$, such that the mean and variance are standardized). The ODE trajectories are obtained by solving the reverse ODE with the exact score function, which can be computed analytically in this case. At first glance, the trajectories of EDM (left) seem straight; however, focusing on the neighborhood of a single delta peak (e.g., $x = \sqrt{\sfrac{3}{2}}$) close to the data space ($ t \to 0$), reveals that trajectories are highly curved (see inset).

Following Liu~\etal~\cite{liu2023flow}, we quantify the local straightness of the reverse trajectories by
\begin{align}
    \rebuttal{
  \mathcal{C}(t) = \E_{\rvx(1) \sim p_1} \bigl[ \lVert (\rvx(1)-\rvx(0)) - \dot{\rvx}(t) \rVert^2 \bigr],
    }
  \label{eq:CurvatureMetric}
\end{align}
\rebuttal{where $\rvx(0)$ and $\dot{\rvx}(t)$ are the endpoint and the instantaneous velocity of the deterministic reverse trajectory ($\lambda=0$ in equation~\ref{eq:GenRevSDE}) started from \textbf{$\rvx(1)$}. The secant $\rvx(1)-\rvx(0)$ is exactly the mean velocity $\int_0^1 \dot{\rvx}(t) \dv t$, so $\mathcal{C}(t)$ measures the squared deviation of the instantaneous velocity from the constant velocity that would traverse a perfectly straight path, and its overall value $\int_0^1 \mathcal{C}(t) \dv t$ is reported on top of the green curves in figure~\ref{fig:overview}. While $\mathcal{C}(t)$ is \emph{not} the geometric curvature of the path, $\int_0^1 \mathcal{C}(t) \dv t = 0$ if and only if the trajectories are straight lines and are traversed at constant speed, i.e. $\ddot{\rvx}(t) = \bfzero$. A geometrically straight path still has a non-zero acceleration if it is traversed at a varying speed, and since the trajectory acceleration controls the discretization error of the solver (see section~\ref{sec:LTE}), the same geometric path can be integrated with different accuracy depending on the schedule. For this reason, we quantify the deviation of the velocity rather than the geometric curvature, and we use the terms ``trajectory curvature'' and ``local curvature'' throughout as shorthand for this velocity deviation.}
Figure~\ref{fig:overview} (Bottom) (green curves) shows that our VP-ISSNR schedule leads to trajectories with smaller $\mathcal{C}(t)$ than both EDM and OTFM. For completeness, figure~\ref{fig:toy_example_3deltas_ours} in \ref{sec:TorajectoryAnalysis} shows ODE trajectories for all other schedules. \rebuttal{Moreover, we provide a two-dimensional version of the same analysis, with a mixture of 12 Gaussians forming two rings, in figure~\ref{fig:toy_example_2d} in \ref{sec:TorajectoryAnalysis}, which shows the same results.}

We observe that discretization errors around $ t \approx 0$ are more severe than errors around $ t \approx 1$, because errors that arise when the marginal $p_t(\rvx)$ has a large support (due to Gaussian diffusion) should not strongly affect sample quality at $t = 0$. This is because such errors do not push the latent sample $\rvx(t)$ into the out-of-distribution region of $p_t(\rvx)$. Instead, they steer samples onto an ``incorrect'' trajectory, potentially violating bijectiveness. However, if the remaining reverse process is solved accurately, samples following such incorrect trajectories can still reach high-quality points at $t=0$.

Thus, we hypothesize that a good schedule should i) have straight trajectories close to the data space ($ t \approx 0$), and ii) the support of the marginal density $p_t(\rvx)$ (relative to the variance at $t=1$) should grow quickly, ideally saturating before $t$ reaches $1$. The relative support $b(t)/ b(t_{\mathrm{max}})$ of the marginal distributions is quantified in Figures~\ref{fig:overview} (Bottom) and \ref{fig:toy_example_3deltas_ours} (see red curves). We observe that for schedules with exploding TV, e.g., SMLD and EDM-UT, the support of the marginal $p_t(\rvx)$ steadily increases until $t=1$ (see figure~\ref{fig:toy_example_3deltas_ours}). In contrast, schedules with non-exploding $\tau(t)$, like OTFM and our VP schedules, already reach (close to) maximum support for $t < 1$.
Furthermore, figure~\ref{fig:overview} (Bottom) demonstrates that both OTFM and EDM exhibit a high trajectory curvature near the data space ($ t \approx 0$) when the relative support is small, in contrast to our VP-ISSNR. \rebuttal{This analysis is qualitative and restricted to a setting in which the score function is available in closed form, so $\mathcal{C}(t)$ can be evaluated without conflating schedule effects with the error of the learned score model. Therefore, it provides an intuition for the mechanism that plausibly underlies the improvements reported in the following sections.}

\subsection{\rebuttal{Trajectory curvature and the local truncation error}}
\label{sec:LTE}

\rebuttal{To formalize the intuition from section~\ref{sec:Discussion}, we now connect the TV schedule and the empirical velocity deviation metric $\mathcal{C}(t)$ to the truncation error of a numerical solver by decomposing its trajectory acceleration using the disentangled coordinates of our framework. A detailed derivation is provided in~\ref{app:LTE}.}

\rebuttal{Defining the variance normalized state $\hat{\rvx}(t) = \rvx(t)/\tau(t)$, we show in~\ref{app:LTE} that the continuous ODE governing $\hat{\rvx}$ depends \emph{only} on the SNR schedule and the data distribution. Consequently, a schedule and its constant TV counterpart generate identical \emph{normalized} trajectories. In continuous time (i.e., at infinite NFE), the TV schedule has no effect on the generated samples, and any difference in sample quality between a schedule and its constant-TV counterpart arises from discretization. This aligns with our empirical observation at high NFEs (figures~\ref{fig:ComparisonMolecule} and~\ref{fig:ComparisonImage}). Therefore, we study the effect of truncation error arising from discretization.}

\rebuttal{First, by expanding the acceleration of the actual state $\rvx = \tau \hat{\rvx}$ within our framework, we obtain the exact decomposition:}
\begin{align}
    \ddot{\rvx}(t) = \underbrace{\frac{\ddot{\tau}}{\tau}\rvx(t) + 2f_\tau \tau \dot{\hat{\rvx}}(t)}_{\text{TV-induced acceleration}} \; + \; \underbrace{\tau \ddot{\hat{\rvx}}(t)}_{\text{SNR-induced acceleration}} ,
    \label{eq:ExactAccelerationDecomposition}
\end{align}
\rebuttal{where $f_\tau = \dot{\tau}/\tau$. Note that the first two terms exist if and only if the TV is dynamic (non-constant). As shown in~\ref{app:LTE}, this decomposition allows us to bound the global truncation error of a first-order Euler approximation over $N$ steps, at leading order in $1/N$, by:}
\begin{equation}
    \lVert \rvx_N - \rvx(0) \rVert \;\lesssim\; \frac{\tau(0)}{2N} \int_0^1 \lVert \hat{\mJ}(0,t) \rVert \left( \left\lvert \frac{\ddot{\tau}}{\tau} \right\rvert \lVert \hat{\rvx} \rVert + 2 \lvert f_\tau \rvert \, \lVert \dot{\hat{\rvx}} \rVert + \lVert \ddot{\hat{\rvx}} \rVert \right) \dv t ,
    \label{eq:NormalizedGlobalError}
\end{equation}
\rebuttal{where $\hat{\mJ}(0,t) = \partial \hat{\rvx}(0)/\partial \hat{\rvx}(t)$ is the Jacobian of the normalized reverse flow from $t$ to $0$, which describes the error propagation, and $\rvx_N$ is the final approximated sample after $N$ Euler steps.
This inequality bounds the truncation error, revealing an explicit dependence on the TV schedule through the dynamic coefficients $f_\tau$ and $\ddot{\tau}/\tau = \dot{f}_\tau + f_\tau^2$. The remaining factors $\lVert \hat{\mJ} \rVert$, $\lVert \hat{\rvx} \rVert$, $\lVert \dot{\hat{\rvx}} \rVert$ and $\lVert \ddot{\hat{\rvx}} \rVert$ are identical for a schedule and its constant-TV counterpart, as they depend only on the SNR and the data distribution.}

\rebuttal{Combined with the argument of section~\ref{sec:Discussion}, this shows that suppressing the acceleration is especially critical near the data manifold ($t \approx 0$), where errors are permanently imprinted onto the final sample, as there is no remaining flow to correct them.
Moreover, the TV schedule penalizes sample generation through its time derivatives, $f_\tau$ and $\ddot{\tau}/\tau$, which is consistent with our empirical observations.
For VE schedules, the rapid noise growth causes these coefficients to become large, which inflates the truncation error bound, especially at low NFEs. 
Conversely, imposing a constant TV ($\tau=1$) makes both derivatives vanish identically, eliminating the TV-induced penalty. For \emph{bounded} non-constant schedules (like OTFM), these coefficients are bounded and may partially counteract the intrinsic SNR acceleration, depending on the data distribution, which is consistent with the discrepancy of performance in OTFM between modalities. A constant TV is therefore a robust, modality-agnostic solution, whereas the performance of bounded dynamic TVs might require careful adaptation to the data distribution to gain further improvements. For non-uniform time grids, as in EDM, the bound applies in the time variable in which the grid is uniform, so that EDM and EDM-UT, which describe the same continuous process, can have different error bounds.}

\rebuttal{Finally, the trajectory acceleration also controls the empirical velocity deviation metric $\mathcal{C}(t)$. Using the Poincar\'e--Wirtinger inequality (see~\ref{app:LTE}), we establish:}
\begin{align}
    \int_0^1 \mathcal{C}(t) \dv t \;\leq\; \frac{1}{\pi^2} \, \E \int_0^1 \lVert \ddot{\rvx}(t) \rVert^2 \dv t .
    \label{eq:CurvatureLTEBound}
\end{align}
\rebuttal{This one-way inequality formalizes $\mathcal{C}(t)$ as a diagnostic tool for schedule instability, and connects it to the truncation error and TV schedule through the acceleration term. Hence, a small mean squared acceleration implies both a small bound on the sum of the local errors (by the Cauchy--Schwarz inequality) and a small integrated velocity deviation, whereas trajectories with a large integrated velocity deviation must suffer from a large mean squared acceleration. If the acceleration remains moderate along the trajectory, this also implies a large bound on the sum of the local errors, which translates into a high truncation error, to the extent that the bound is tight.}

\subsection{Application to molecular structure generation}
\label{sec:ExperimentsMolecular}

Building on the intuition that low curvature near the data manifold and rapid support growth are critical for efficient sampling, we test our framework on the complex task of molecular structure generation. The goal is to predict an equilibrium state $R$ given a molecular composition $Z$, i.e., $\rvx(0) \sim p(R|Z)$. Note the difference from the general molecule generation task, where the composition $Z$ is also predicted, and conformation search, where we condition on a molecular graph.
Our experiments systematically evaluate different scheduling techniques used in diffusion and flow matching models for molecular tasks~\cite{edm, geoldm, vignac2023midi, kahouli2024morered, song2023equivariant, eqgat-diff, et-flow}. Common schedules include DDPM-cos with $\nu=1$ and $\nu=2.5$ and OTFM. We use the QM9 dataset~\cite{qm9}, a widely used benchmark comprising $\sim 130\,000$ equilibrium molecules with up to 9 heavy atoms (C, O, N, and F) for our evaluation. Following Kahouli~\etal~\cite{kahouli2024morered}, we use a training/validation/test split of $55\,000$/$10\,000$/$10\,000$ molecules.

\begin{figure*}[t]
    \centering
    \hspace{5mm}
    \includegraphics[width=.42\linewidth]{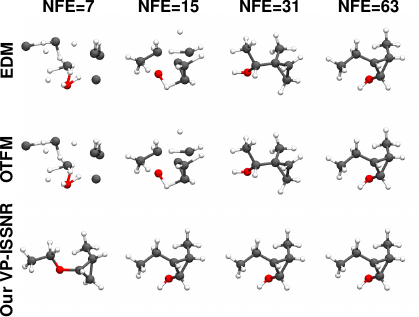}
    %\hspace{1mm}
    \includegraphics[width=.47\linewidth]{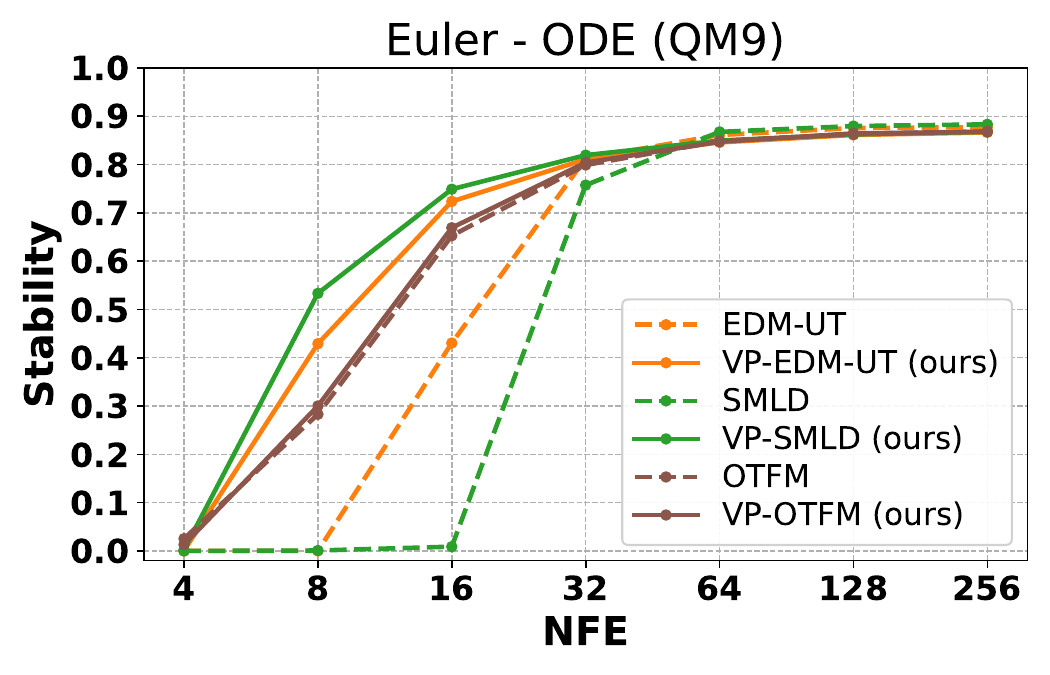}
    \includegraphics[width=.47\linewidth]{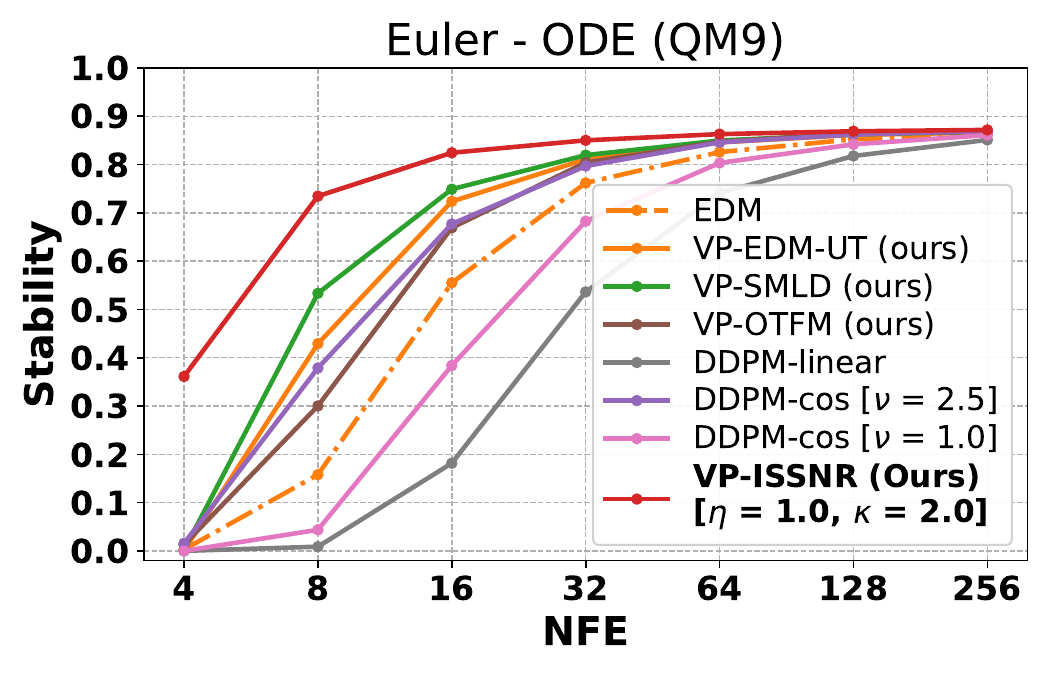}
    \includegraphics[width=.47\linewidth]{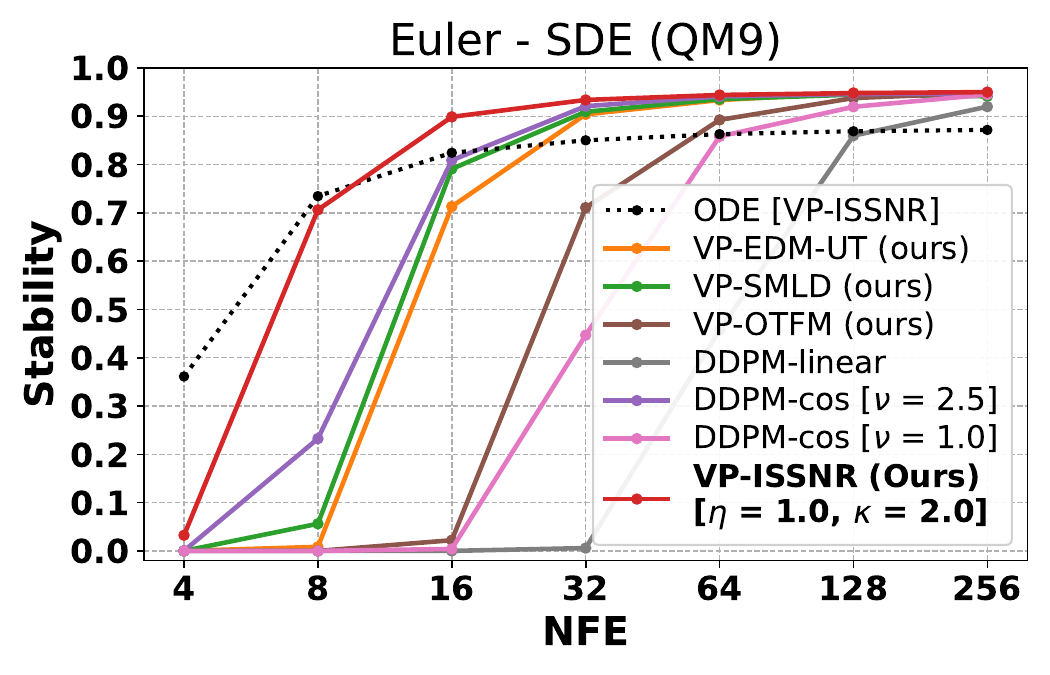}
    \vspace{-1mm}
    \caption{
    Stability rate (\textbf{higher is better}) as a function of the number of function evaluations (NFE) for molecular structure generation on the QM9 dataset. \textbf{Top-right:} Comparison between commonly established non-VP schedules (i.e., $\tau(t)$ non-constant) and their VP counterparts (i.e., $\tau(t)=1$) in our framework. \textbf{Bottom-left:} Comparison of various baselines, including the VP analogs from the top-right plot, against our proposed VP-ISSNR schedule, with fixed parameters $\eta=1.0$ and $\kappa=2.0$ for all NFEs. \textbf{Bottom-right:} Same as the left plot but using the reverse SDE, instead of the reverse ODE. The best-performing ODE schedule, VP-ISSNR, is highlighted in black for reference. For reference, the \textbf{top-left} panel shows the sampled molecular structures obtained from the same prior sample using our approach, compared against the two widely used EDM and OTFM methods, as the number of NFEs increases. 
    \rebuttal{Each curve represents the mean across five independent sampling runs with different random seeds. The standard deviations are too small to be visually perceptible (median $0.6\,\%$) and are provided in Table~\ref{tab:MetricsStability}.}  
    }
    \vspace{-3mm}
    \label{fig:ComparisonMolecule}
\end{figure*}

For training, we adopt the noise model architecture used in Kahouli~\etal~\cite{kahouli2024morered} and minimize the DSM loss \eqref{eq:DSM} using the DDPM-cos schedule with $\nu=1.0$. Unless otherwise stated, for sample generation, we solve the reverse ODE in \eqref{eq:GenRevSDE} for $\lambda = 0$, using first-order Euler integration. 
\rebuttal{To ensure statistical robustness, we perform five independent sampling runs with different random seeds. For each seed, we generate $2\,500$ structures conditioned on compositions $Z$ sampled from the test split (see~\ref{sec:Splits}). We use identical initial noise samples across all evaluated schedules to guarantee a paired comparison.} 
For varying computational budgets, defined by the number of function evaluations (NFEs),
\rebuttal{we report the mean stability rates in Figure~\ref{fig:ComparisonMolecule}. Because the standard deviations are too small to be visually perceptible as error bars on the curves, we summarize the detailed numerical results in Table~\ref{tab:MetricsStability}. Alongside the stability rate, we also report the means and standard deviations for validity, uniqueness, and novelty in Tables~\ref{tab:MetricsValidity}--\ref{tab:MetricsNovelty}.} Further experimental details, \rebuttal{including explicit metric definitions and data splits}, are provided in~\ref{sec:A.AlgorithmDetails}.

We first compare the stability rates of the VE schedules (SMLD and EDM-UT) as well as the VM schedule (OTFM), against their VP counterparts: VP-SMLD, VP-EDM-UT, and VP-OTFM. As shown in figure~\ref{fig:ComparisonMolecule} (top-right), across all schedules, the VP variants outperform or match the performance of their original versions. Specifically, the VP versions lead to substantial enhancements for SMLD and EDM-UT, where the TV increases exponentially, whereas OTFM with a smooth TV modulation performs comparably to its VP analog. These findings suggest that exponentially increasing TV can be detrimental, thereby validating our choice to adopt a constant TV schedule.

Now, we evaluate our proposed VP-ISSNR schedule. Figure~\ref{fig:ComparisonMolecule} (bottom-left) presents results for all schedules, excluding the original VE schedules, which perform worse compared to their VP analogs \rebuttal{in the low-NFE regime}. It also includes the original EDM (with non-uniform time grid) and our proposed VP-ISSNR schedule, using the fixed parameters $\eta = 1.0$, and $\kappa = 2.0$ for all NFE values (see~\ref{sec:Endpoints} for the schedule endpoints). Notably, our VP-ISSNR schedule generates stable molecules in as few as 4 NFEs with the first-order solver Euler. The stability rate reaches \rebuttal{$73.5 \pm 0.6\,\%$} with only 8 steps and \rebuttal{$86.9 \pm 0.5\,\%$} with 128 steps, significantly outperforming all other schedules \rebuttal{in the low-NFE regime.} Figure~\ref{fig:ComparisonMolecule} (top-left) visually confirms this efficiency, where our VP-ISSNR approach generates stable molecular structures much earlier in the reverse process compared to EDM and OTFM.
Additionally, figure~\ref{fig:ComparisonMolecule} (bottom-right) shows that when adding stochasticity and solving the reverse SDE ($\lambda=1$ in \eqref{eq:GenRevSDE}) using our VP-ISSNR schedule, the stability rate increases significantly to \rebuttal{$93.4 \pm 0.5\,\%$} with 32 NFEs and \rebuttal{$94.4 \pm 0.8\,\%$} with 64 NFEs. 
While the ODE achieves a higher stability rate than the SDE at very low NFEs (e.g., \rebuttal{$73.5\,\%$ vs.\ $70.6\,\%$} at 8 NFEs), the SDE achieves superior stability as NFEs increase, suggesting that stochasticity introduces a corrective effect that enhances sample quality in molecules, albeit with a slight increase in sampling time. \rebuttal{Moreover, the seed-to-seed variation of the stability rate (table~\ref{tab:MetricsStability}) is small, at most $1.5\,\%$ and with a median of $0.6\%$ across all schedules and NFEs, which makes the comparison in the low-NFE regime unambiguous. Binomial $95\,\%$ confidence intervals are given in table~\ref{tab:StabilityBinomialCI}, and their half-width is at most $0.9\,\%$.}

In figure~\ref{fig:AllComparisonMolecule_Heun} in \ref{sec:A.AdditionalExperimentalResultsMols}, we present additional experimental results using the second-order Heun's method and the advanced DPM solver \cite{dpm-solver, lu_dpm-solver_2023}, where we observe similar trends. This demonstrates that our noise schedule optimization is orthogonal to the choice of sampler, i.e., it can be combined with advanced sampling methods for even larger overall improvements. We also report the results of training models with different schedules in figure~\ref{fig:AllComparisonMolecule_EulerODE_AnotherTraining}, together with the SNR coverage of the two training schedules in figure~\ref{fig:SNRCoverage}, and assess the quality of samples by running DFT relaxations (see \ref{sec:DFTcalculations} for details) to identify the nearest reference structure in figure~\ref{fig:AllComparisonMolecule_Heun_RMSD}, mirroring the trends in the stability rate results.
\rebuttal{Tables~\ref{tab:MetricsValidity}--\ref{tab:MetricsNovelty} complement the stability rate with validity, uniqueness, and novelty. Validity is a stricter criterion, where only $94.5\,\%$ of the ground-truth DFT-optimized molecules in QM9 are valid but $99.7\,\%$ are stable (\ref{sec:StabilityDefinition}). These values serve as the empirical upper bound for our model's performance because, unlike some previous work, we do not filter them out of the training data. Accordingly, validity lies slightly below the stability rate, but it ranks the schedules essentially in the same order. With the reverse ODE, VP-ISSNR attains the highest validity at every NFE, e.g.\ $66.9 \pm 0.5\,\%$ with only 8 NFEs compared with $47.2 \pm 1.0\,\%$ for the next best schedule, VP-SMLD, and $81.7 \pm 0.7\,\%$ at 128 NFEs. With the reverse SDE, it reaches $84.6 \pm 0.8\,\%$ at 16 NFEs, and $90.0 \pm 0.5\,\%$ at 256 NFEs. The uniqueness is $\geq 94\,\%$ for all schedules and NFEs. The improved stability of VP-ISSNR at low NFEs is thus not obtained by generating duplicates. The novelty is highest at the lowest NFEs, and decreases for all schedules to about $50\,\%$ (ODE) as the NFE grows. This is expected, as very high novelty is not necessarily desirable, when the model should reproduce the target data distribution.}

\rebuttal{Because all schedules are evaluated with the same trained model and the same solver, the only quantity that changes the wall-clock cost is the cost of one NFE. A single evaluation of the denoiser takes $0.05$s for a batch size of 128 on NVIDIA P100, and the total sampling time is that cost multiplied by the NFE.}

\subsection{Generalization to image modalities}
\label{sec:ExperimentsImage}

\begin{figure*}[t]
    \centering
    \hspace{3mm}
    \vspace{2mm}
    \includegraphics[width=.41\linewidth]{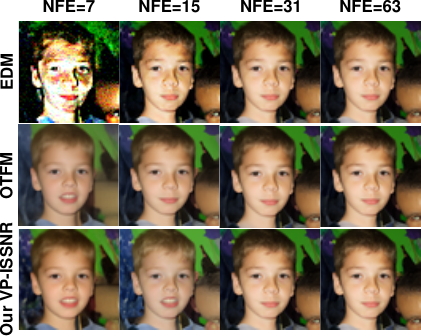}
    \hspace{2mm}
    \includegraphics[width=.47\linewidth]{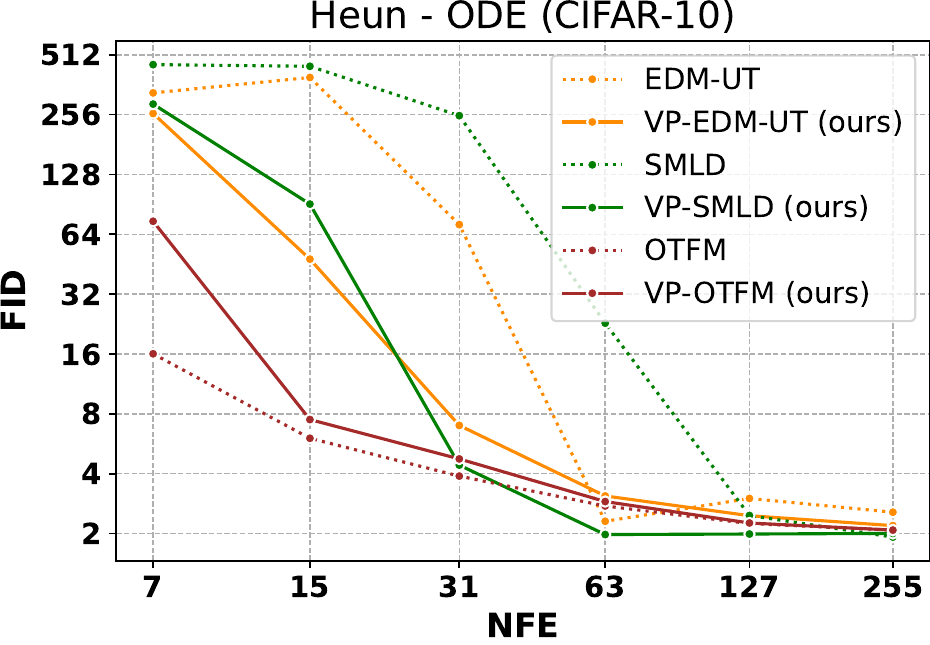}
    \includegraphics[width=.47\linewidth]{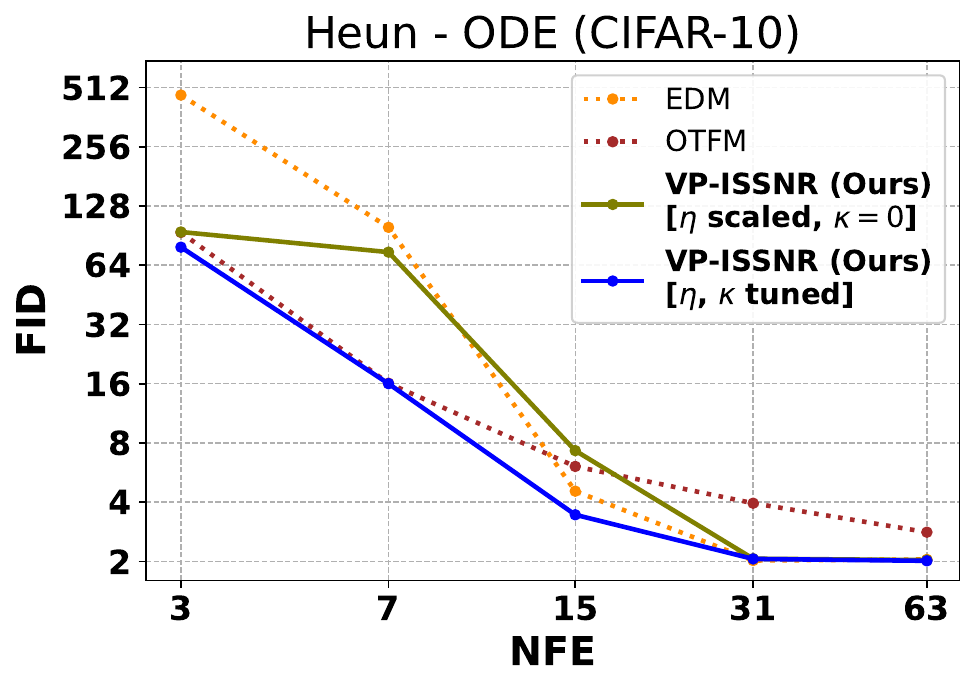}
    \includegraphics[width=.47\linewidth]{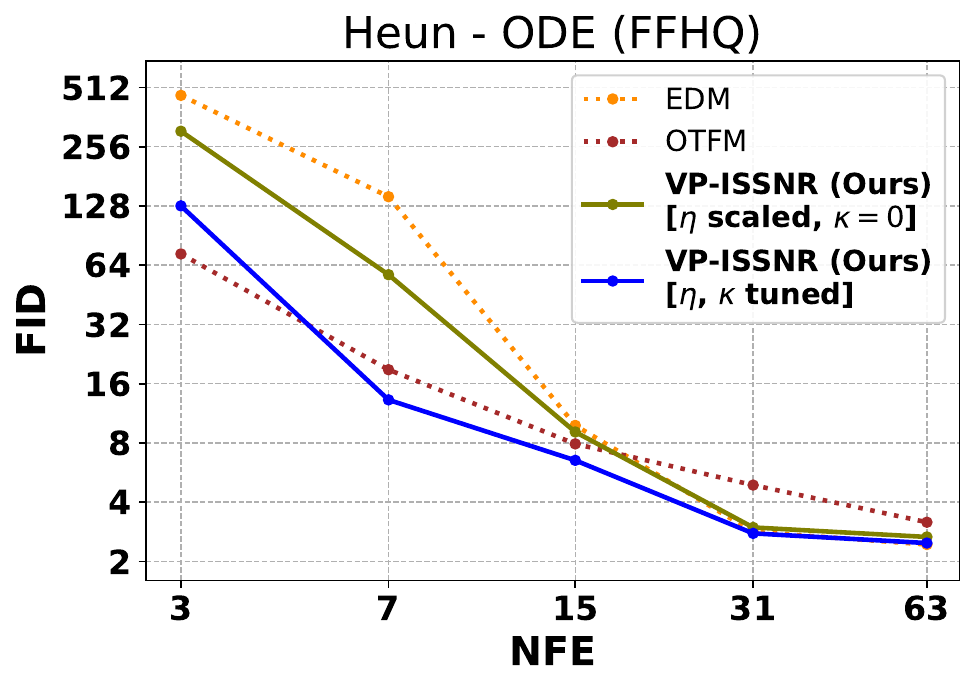}
    \vspace{-1mm}
    \caption{
    FID score (\textbf{lower is better}) as a function of the number of function evaluations (NFE) for image generation on CIFAR-10 and FFHQ. \textbf{Top-right:} Comparison of existing non-VP models and their VP variants on CIFAR-10. \textbf{Bottom-left:} Comparison on CIFAR-10 between baseline methods and two variants of our proposed VP-ISSNR method, one using a scaled $\eta$ in function of the NFE, and the other tuning $\eta$ and $\kappa$ for each NFE via Bayesian optimization, as detailed in section~\ref{sec:ExperimentsImage}. \textbf{Bottom-right:} Same comparison as in the left plot, but on higher-resolution FFHQ images. \rebuttal{Both plots report the mean FID over five independent sampling runs with different random seeds.}
    For reference, the \textbf{top-left} panel shows the generated images obtained from the same prior sample using our approach, compared against the two widely used EDM and OTFM methods, as the number of NFEs increases.
    }
    \vspace{-3mm}
    \label{fig:ComparisonImage}
\end{figure*}

\begin{table}[t]
\centering
\caption{\rebuttal{FID (\textbf{lower is better}) on CIFAR-10 and FFHQ with Heun's method as a function of the number of function evaluations (NFE). Each entry is the mean over five independent sampling runs with different random seeds, each computed from $50\,000$ generated images, with $\pm 1$ standard deviation over the seeds as a subscript. Within each column, every value whose $\pm 1$ standard deviation interval overlaps that of the lowest FID is set in bold.}}
\label{tab:ImageFID}
\footnotesize
\setlength{\tabcolsep}{4pt}
\begin{tabular}{@{}lccccc@{}}
\br
Schedule~/~NFE & \textbf{3} & \textbf{7} & \textbf{15} & \textbf{31} & \textbf{63} \\
\mr
\multicolumn{6}{l}{\emph{CIFAR-10}} \\
EDM & 468.3$_{\pm0.1}$ & 99.66$_{\pm0.25}$ & 4.55$_{\pm0.06}$ & \textbf{2.02}$_{\pm0.03}$ & \textbf{2.05}$_{\pm0.04}$ \\
OTFM & 93.92$_{\pm0.17}$ & \textbf{16.08}$_{\pm0.03}$ & 6.08$_{\pm0.05}$ & 3.95$_{\pm0.05}$ & 2.82$_{\pm0.05}$ \\
VP-ISSNR$[\eta=1.5, \kappa=1.0]$ (ours) & 322.7$_{\pm0.2}$ & 130.2$_{\pm0.3}$ & 4.67$_{\pm0.05}$ & 2.11$_{\pm0.03}$ & \textbf{2.02}$_{\pm0.03}$ \\
VP-ISSNR$[\eta$ scaled$]$ (ours) & 94.39$_{\pm0.28}$ & 74.56$_{\pm0.30}$ & 7.31$_{\pm0.06}$ & \textbf{2.07}$_{\pm0.03}$ & \textbf{2.03}$_{\pm0.03}$ \\
VP-ISSNR$[\eta,\kappa$ tuned$]$ (ours) & \textbf{79.04}$_{\pm0.21}$ & \textbf{16.00}$_{\pm0.09}$ & \textbf{3.45}$_{\pm0.05}$ & \textbf{2.06}$_{\pm0.03}$ & \textbf{2.01}$_{\pm0.03}$ \\
\mr
\multicolumn{6}{l}{\emph{FFHQ}} \\
EDM & 466.8$_{\pm0.1}$ & 142.6$_{\pm0.3}$ & 9.82$_{\pm0.07}$ & \textbf{2.83}$_{\pm0.04}$ & \textbf{2.44}$_{\pm0.03}$ \\
OTFM & \textbf{73.02}$_{\pm0.09}$ & 18.84$_{\pm0.07}$ & 7.91$_{\pm0.06}$ & 4.88$_{\pm0.06}$ & 3.16$_{\pm0.04}$ \\
VP-ISSNR$[\eta=1.5, \kappa=1.0]$ (ours) & 448.4$_{\pm0.2}$ & 213.8$_{\pm0.2}$ & 11.38$_{\pm0.02}$ & 3.46$_{\pm0.04}$ & 2.72$_{\pm0.03}$ \\
VP-ISSNR$[\eta$ scaled$]$ (ours) & 306.9$_{\pm0.2}$ & 57.27$_{\pm0.20}$ & 9.10$_{\pm0.02}$ & 2.98$_{\pm0.03}$ & 2.66$_{\pm0.03}$ \\
VP-ISSNR$[\eta,\kappa$ tuned$]$ (ours) & 128.1$_{\pm0.3}$ & \textbf{13.24}$_{\pm0.07}$ & \textbf{6.53}$_{\pm0.05}$ & \textbf{2.78}$_{\pm0.03}$ & \textbf{2.48}$_{\pm0.03}$ \\
\br
\end{tabular}
\end{table}

While our primary focus is optimizing noise scheduling for scientific applications, we also evaluate the generalization of our proposed schedules on the standard, unconditional image generation task. Following the setup in Karras~\etal~\cite{diff_edm}, we use their pre-trained diffusion models and assess sample quality using the average FID score \cite{fid} computed over $50\,000$ generated images as a function of NFEs on CIFAR-10~\cite{cifar10} and the higher-resolution FFHQ~\cite{Karras2018ASG}. Unless otherwise stated, samples are generated by solving the reverse ODE ($\lambda=0$) using the commonly used second-order Heun's method. 
\rebuttal{The results are summarized in figure~\ref{fig:ComparisonImage}. For the schedules compared in the two lower panels, every FID is averaged over five independent sampling runs with different random seeds, and all schedules start from the same initial noise at a given seed. The standard deviations are reported in table~\ref{tab:ImageFID}.}

Similar to the experiment with molecules in figure~\ref{fig:ComparisonMolecule} (top-right), we first compare the original non-VP schedules to their VP analogs. Figure~\ref{fig:ComparisonImage} (top-right) summarizes the performance on CIFAR-10 (results on FFHQ show similar trends, see \ref{sec:A.AdditionalExperimentalResults}). Consistent with our findings for molecular structure generation, we observe that both SMLD and EDM-UT, which feature exploding TV schedules, benefit significantly from adopting a constant TV schedule. In contrast, the VP-OTFM does not perform as well as the original OTFM with modulated TV, which is different from what we observe for molecular structure generation. \rebuttal{The consistent improvement using a constant TV concerns schedules with an exponentially exploding TV, whereas for schedules with a bounded TV, such as OTFM, the preferable TV schedule may depend on the data modality (see section~\ref{sec:LTE}).} This implies the possibility of further improving fast sampling by optimizing the TV control, which we leave as future work.

Furthermore, we compare our proposed VP-ISSNR schedule with the original EDM and OTFM schedules. Note that our proposed inverse sigmoid (ISSNR) schedule has two free hyperparameters $\eta$ and $\kappa$. Although fixing these to constant values is sufficient to achieve improved performance in molecular structure generation, this is not the case for images, as presented in \ref{sec:AdditionalExperimentalResultsImages}. As expected, our generalized approach with fixed hyperparameters does not outperform highly specialized methods like EDM in unconditional image generation. Baselines such as EDM have been specifically optimized for computer vision modalities, relying strongly on carefully tuned non-uniform time grids, as evidenced by the poorer results of its uniform-time variant, EDM-UT, in figure~\ref{fig:ComparisonImage} top-right. \rebuttal{Our analysis in~\ref{app:LTE} shows that the coefficients through which the TV enters the error bound depend on the variable in which the ODE is integrated, so that EDM and EDM-UT, which describe the same continuous process, can have different error bounds under discretization.}

Therefore, we explore two different approaches. First, we test a simple modification, where we scale $\eta$ logarithmically as $2\eta = 2 + \max(0, \log_2(\texttt{nfe} + 1) - 3)$ with the NFEs and set $\kappa=0$ (labeled as $\eta$ scaled in the plots). \rebuttal{At high enough NFE,} this schedule achieves competitive results, as illustrated in the two bottom panels in figure~\ref{fig:ComparisonImage}. To further enhance performance, it is possible to tune $\eta$ and $\kappa$ for different datasets and NFEs, where we use Bayesian optimization over 32 trials to find optimized values. \rebuttal{The values obtained for every dataset and NFE, together with the fixed and $\eta$-scaled settings, are listed in table~\ref{tbl:hyperopt_values}, and details about the Bayesian optimization procedure, including the search space in table~\ref{tbl:hyperopt_params}, are provided in \ref{sec:BOdetails}.} We refer to this variant as \rebuttal{VP-ISSNR$[\eta,\kappa$ tuned$]$} (the blue curves in the bottom plots in figure~\ref{fig:ComparisonImage}). With the tuned hyperparameters, our \rebuttal{VP-ISSNR} improves the image quality in terms of FID, particularly in the low-NFE regime, matching or improving on EDM and OTFM. This quantitative improvement is visually illustrated in figure~\ref{fig:ComparisonImage} (top-left), where our VP-ISSNR schedule converges to high-quality visual samples faster than EDM, and performs comparably to OTFM at very low NFE budgets. \rebuttal{Note that because re-tuning schedules introduced by previous work from scratch is computationally expensive, we compare our tuned VP-ISSNR variant against their published, optimized hyperparameters rather than conducting a strictly equal-budget optimization. While using fixed hyperparameters for molecular structure generation is sufficient to achieve improved performance, tuning the hyperparameters as done for images may yield further improvements, which we leave as future work.} 

Similar to the experiments on molecules, to further evaluate the orthogonality of our approach to different choices of solvers, we also assess its performance using the first-order Euler, the higher-order Runge–Kutta (RK45), and the advanced DPM solvers \cite{dpm-solver, lu_dpm-solver_2023}. The results are reported in figure~\ref{fig:SolversComparison} in \ref{sec:AdditionalExperimentalResultsImages}. Additionally, we include results on more high-resolution datasets such as AFHQ~\cite{afhq} and ImageNet~\cite{imagenet} in figure~\ref{fig:ImagesDatasetsComparison}. We observe consistent trends across solvers and datasets.

In summary, our empirical evaluation demonstrates the significant advantage of our principled approach for scientific tasks, such as molecular structure generation, on the one hand, and highlights the fundamental differences between data modalities, on the other. The fact that strongly optimized schedules for images underperform on molecular tasks reveals the limitations of adopting computer vision defaults for scientific machine learning.
Despite this, our proposed TV/SNR framework, together with the VP-ISSNR schedule, achieves highly competitive results on image generation, especially when using more flexible, tuned approaches for the hyperparameters, demonstrating its generalizability.

\section{Conclusion}
\label{sec:conclusion}
The performance and sampling efficiency of diffusion models are highly dependent on the chosen noise schedules, which control the trade-off between noise injection and signal preservation. Existing methodologies often control the total variance implicitly, lacking direct control over this crucial balance. To address this, we introduced a novel total-variance/signal-to-noise-ratio (TV/SNR) disentangled framework for designing noise schedules, which allows controlling the TV and SNR independently. 

We empirically demonstrated that established schedules where the TV grows exponentially can be improved by imposing a constant TV, while keeping the SNR schedule unchanged. Building on this insight and using our principled TV/SNR diffusion framework, we proposed the VP-ISSNR schedule as a generalization of the optimal transport flow matching schedule. This approach leads to significant performance improvements in molecular structure generation, where stable molecules can be generated after only 4 steps, representing a {\em speedup of up to an order of magnitude} compared to other approaches. \rebuttal{For image generation, where the baselines are highly optimized, tuning the two hyperparameters of the ISSNR schedule gives results on par with or better than EDM and OTFM.} To provide theoretical insights on these empirical findings, we analyzed the ODE trajectories for a simple toy model to explain the mechanisms that are responsible for the observed increase in sample efficiency\rebuttal{, and derived a truncation error bound that explicitly shows the dependence on the TV schedule.} 

Building on our new TV/SNR framework, future work could follow two main directions. Conceptually, we plan to explore the broader design space of dynamic TV schedules\rebuttal{, including bounded non-constant TV schedules adapted to the data modality,} integrate our TV/SNR framework with distillation and one-shot generation techniques to further accelerate inference, and study the effect of the schedule design during training. On the application level, \rebuttal{ while our current
experiments focused on the small-scale dataset QM9 and serve as a proof of concept, we aim to scale this optimized framework to target large molecular systems, such as proteins and complex drug-like molecules, where sampling efficiency remains a critical bottleneck.}

\section*{\rebuttal{Data and code availability statement}}

\rebuttal{The QM9 dataset~\cite{qm9} and the pre-trained image diffusion models of Karras~\etal~\cite{diff_edm} used in this work are publicly available from their original sources. The code implementing the TV/SNR framework, all schedules reported in table~\ref{tab:ExistingTVSNR}, and the sampling and evaluation pipeline required to reproduce the results is available at \nolinkurl{https://github.com/khaledkah/tv-snr-diffusion}.}

\ack
This work was partly funded by the German Ministry for Education and Research (BMBF) as BIFOLD – Berlin Institute for the Foundations of Learning and Data - under Grants 01IS14013A-E, 01GQ1115, 01GQ0850, 01IS18025A, 031L0207D, and 01IS18037A.
Furthermore, K.R.M. was
supported by Institute of Information \& communications Technology Planning \& Evaluation (IITP) grants funded by the Korean government(MSIT) (No.~RS-2019-II190079, Artificial Intelligence Graduate School Program of Korea University and No.~RS-2024-00457882, AI Research Hub Project) and Hector Fellow Academy.
S.G. was supported by the Postdoc.Mobility fellowship by the Swiss National Science Foundation (project no. 225476).

\section*{References}
\bibliographystyle{unsrt}
\bibliography{tv_snr}

\newpage
\appendix
\section{Derivation of TV/SNR expressions of existing schedules}
\label{sec:A.SCLSNRExpressionExisting}

% Appendix float numbering is left to iopart.cls, which numbers floats per
% appendix section (A1, B1, C1, D1, ...). Previously \thefigure was forced to
% A\arabic{figure} while the counter was still reset at every appendix section,
% which produced two different figures both numbered "A1".

In this appendix, we show the derivations of the TV, $\tau(t)$, and SNR, $\gamma(t)$, for existing diffusion model schedules, which are summarized in table~\ref{tab:ExistingTVSNR}.

\subsection{Variance-exploding (VE) schedules}

Given is the VE perturbation kernel, originally introduced by Song~\etal~\cite{song2019score} as Score Matching with Langevin Dynamics (SMLD) and then reframed in the SDE framework by \cite{song2021score} as VE-SDE,
\begin{align}
    p(\rvx(t)|\rvx(0)) = \mathcal{N}\left(\rvx(t); \rvx(0), b^2(t) \mathbf{I}\right)
\end{align}
with
\begin{align}
    a(t) &= 1, \label{eq:VE_a} \\
    b^2(t) &= \sigma_{\mathrm{min}}^2 \left( \frac{\sigma_{\mathrm{max}}}{\sigma_{\mathrm{min}}} \right)^{2t}.
\end{align}
Here, $\sigma_{\mathrm{min}}$ and $\sigma_{\mathrm{max}}$ are the minimum and maximum noise scales, respectively.

\paragraph{SNR:}
\begin{align}
    \gamma^2(t)
    = \frac{a^2(t)}{b^2(t)}
    = \frac{1}{\sigma_{\mathrm{min}}^2 \left( \frac{\sigma_{\mathrm{max}}}{\sigma_{\mathrm{min}}} \right)^{2t}}
    = \sigma_{\mathrm{min}}^{-2} \left( \frac{\sigma_{\mathrm{min}}}{\sigma_{\mathrm{max}}} \right)^{2t}.
\end{align}

\paragraph{TV:}
\begin{align}
    \tau^2(t) &= a^2(t) + b^2(t) \notag\\
    &=1 + \sigma_{\mathrm{min}}^2 \left( \frac{\sigma_{\mathrm{max}}}{\sigma_{\mathrm{min}}} \right)^{2t}.
\end{align}

\subsection{Elucidating design space of diffusion models (EDM)}

EDM \cite{diff_edm} introduces a noise schedule with a scaling factor $s(t)$ and noise level $\sigma(t)$.
The perturbation kernel is
\begin{align}
    p(\rvx(t)|\rvx(0)) = \mathcal{N}\left(\rvx(t); s(t) \rvx(0), [s(t) \sigma(t)]^2 \mathbf{I}\right)
\end{align}
where they use,
\begin{align}
    a(t) &= s(t) = 1, \\
    b^2(t) &= \sigma^2(t).
\end{align}

\begin{figure}[t]
    \centering
    \includegraphics[width=.49\linewidth]{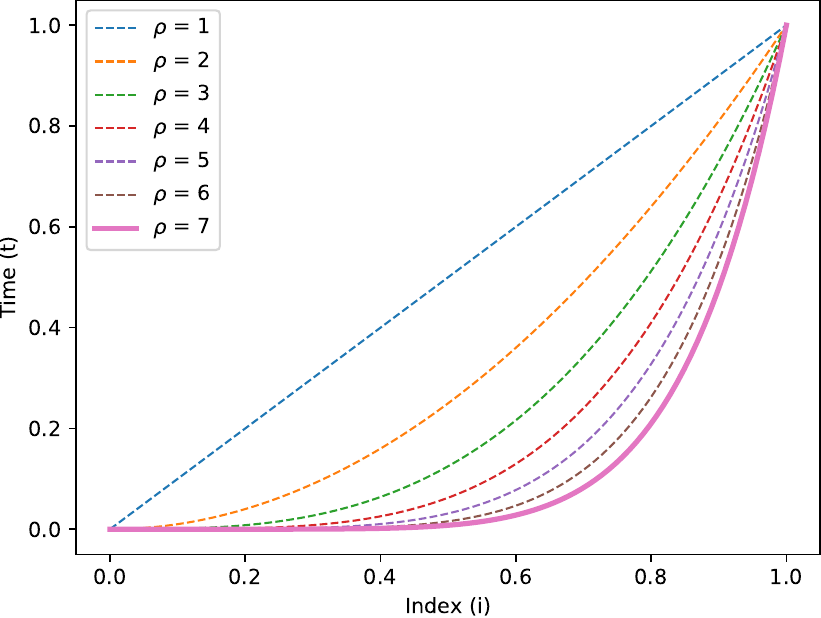}
    \caption{
    The time discretization in equation~\ref{eq:EDMTimeGrid} of the original EDM sampler \cite{diff_edm} for various parameters $\rho$. At $\rho = 1$, linear time discretization is recovered, i.e. $t(i) = i$. For larger $\rho$, time steps get considerably shorter close to the data distribution. EDM uses $\rho = 7$.}
    \label{fig:EDMTimeGrid}
\end{figure}

The time discretization in equation~\ref{eq:EDMTimeGrid} of the original EDM is illustrated in figure~\ref{fig:EDMTimeGrid}.

\paragraph{SNR:}
\begin{align}
    \gamma^2(t) = \frac{a^2(t)}{b^2(t)} = \frac{1}{\sigma^2(t)}.
\end{align}

\paragraph{TV:}
\begin{align}
    \tau^2(t) &= a^2(t) + b^2(t) \notag\\
    &= 1 + \sigma^2(t).
\end{align}

\subsection{Optimal transport flow matching (OTFM)}
The perturbation kernel of OTFM \cite{liu2023flow, lipman2023flow} is given by:
\begin{align}
    p_t(\rvx_t|\rvx_0) =  \mathcal{N}\left(\rvx(t); (1-t) \rvx(0) ,  ((1 - t) \sigma_\mathrm{min} + t)^2 \mI \right)
\end{align}
where $\sigma_\mathrm{min}$ is sufficiently small, resulting in a Gaussian distribution concentrated around $x(0)$. Assuming $\sigma_\mathrm{min} = 0$ and incorporating the boundary constraint on $t \in [t_{\mathrm{min}},1]$, we can define:

\paragraph{SNR:}
\begin{align}
    \gamma^2(t) = \frac{(1-t)^2}{t^2} = \left(\frac{1}{t} -1\right)^2 .
\end{align}

\paragraph{TV:}
\begin{align}
    \tau^2(t) = (1-t)^2 + t^2 = 1-2t(1-t).
\end{align}

\subsection{Denoising diffusion probabilistic models (DDPM)}
The original DDPM model \cite{DDPM_Ho} used the perturbation kernel:
\begin{align}
    p(\rvx(t)|\rvx(0)) = \mathcal{N}\left(\rvx(t); \sqrt{\bar{\alpha}(t)} \rvx(0), (1-\bar{\alpha}(t)) \mathbf{I}\right)
\end{align}
\paragraph{SNR}
\begin{align}
    \gamma^2(t) = \frac{a^2(t)}{b^2(t)} = \frac{\bar{\alpha}(t)}{1-\bar{\alpha}(t)}.
\end{align}

\paragraph{TV}
\begin{align}
    \tau^2(t) &= a^2(t) + b^2(t) \notag\\
    &= 1.
\end{align}

While the TV is always constant in the VP case, different schedules were adopted for $\bar{\alpha}(t)$. The most common are:
\begin{itemize}
    \item \textbf{DDPM-linear} the original linear schedule introduced by Ho~\etal~\cite{DDPM_Ho} and adopted by Song~\etal~\cite{song2021score} as VP-SDE, where:\\
    \begin{align*}
        \bar{\alpha}(t) =  e^{ -\frac{1}{2} t^2 (\beta_\mathrm{max} - \beta_\mathrm{min}) - t \beta_\mathrm{min}},
    \end{align*}
    and therefore
    \begin{align}
        \gamma^2(t) &= \frac{ e^{ -\frac{1}{2} t^2 (\beta_\mathrm{max} - \beta_\mathrm{min}) - t \beta_\mathrm{min}}}{1 -  e^{ -\frac{1}{2} t^2 (\beta_\mathrm{max} - \beta_\mathrm{min}) - t \beta_\mathrm{min}}} \notag \\
        &= \left( \frac{1 - e^{ -\frac{1}{2} t^2 (\beta_\mathrm{max} - \beta_\mathrm{min}) - t \beta_\mathrm{min}}}{e^{ -\frac{1}{2} t^2 (\beta_\mathrm{max} - \beta_\mathrm{min}) - t \beta_\mathrm{min}}} \right)^{-1} \notag \\
        &= \left( e^{ \frac{1}{2} t^2 (\beta_\mathrm{max} - \beta_\mathrm{min}) + t \beta_\mathrm{min}} -1 \right)^{-1}.
    \end{align}
    \item \textbf{DDPM-cos} First introduced by Nichol~\etal~\cite{imp_ddpm} as a better alternative to the linear schedule:
    \begin{align*}
        \bar{\alpha}(t) = \frac{u(t)}{u(0)}, \quad \text{where} \quad u(t)=\cos\left(\frac{t^\nu + s}{1 + s} \frac{\pi}{2}\right)^2.
    \end{align*}
    Note that the parameter $\nu$ does not exist in the original formulation \cite{imp_ddpm} but we adopt it from Vignac~\etal~\cite{vignac2023midi}. The SNR is therefore:
    \begin{align}
        \gamma^2(t) &= \frac{u(t)}{u(0) \left( 1 - \frac{u(t)}{u(0)} \right)} \notag \\
        &= \frac{u(t)}{u(0)-u(t)} \notag \\
        &= \left(\frac{u(0)}{u(t)} -1\right)^{-1} \notag \\
        &= \left( \left( \frac{\cos\left(\frac{s}{1 + s} \frac{\pi}{2}\right)}{\cos\left(\frac{t^\nu + s}{1 + s} \frac{\pi}{2}\right)} \right)^2 -1 \right)^{-1}.
    \end{align}
\end{itemize}

\section{Derivation of the SDE}
\label{app:SDE}
\subsection{Derivation of the perturbation kernel for a given affine SDE}
\label{app:SDEtoKernel}
As shown by Song~\etal~\cite{song2021score}, a diffusion process can be described by a continuous stochastic differential equation (SDE) describing an Itô process:
\begin{equation}
    \dv \rvx =  \mathbf{f}(\rvx, t) \, \dv t + \mathbf{g}(\rvx, t) \, \dv \rvw,
    \label{eq:A.GenForwSDE}
\end{equation}
where $\rvx$ is the state variable, $\rvw$ is the standard Wiener process, and $\mathbf{f}(\rvx, t)$ and $\mathbf{g}(\rvx, t)$ are predefined functions describing the drift and diffusion coefficients, respectively.

While Song~\etal~\cite{song2021score} derived the perturbation kernel using the differential equations for the mean and covariance of an SDE, we take a different approach. By exploiting the affine nature of the SDE in our case, we first solve the SDE and then derive the perturbation kernel parameters, arriving at the same solution. Specifically, for the affine case where
\begin{equation}
    \dv \rvx =  \textcolor{black}{f(t)} \rvx \, \dv t + \textcolor{black}{g(t)} \, \dv \rvw,
    \label{eq:AffineForwSDE_si}
\end{equation}
the integral of the Itô process is described by equation (4.28) in \cite{sarkka_applied} as
\begin{equation}
    \rvx(t) = \phi(t, 0) \rvx(0) + \int_0^t \phi(t, s) g(s) \, \dv \rvw(s)
    \label{eq:solutionForwSDE}
\end{equation}
with an integrating factor $\phi(t, s)$ and initial condition $\rvx(0)$. In the following we use the integrating factor $\phi(t, s) = \exp\left( \int_s^t f(u) \, \dv u \right)$.

The solution $\rvx(t)$ involves a deterministic part dependent on $\rvx(0)$ and a stochastic part independent of $\rvx(0)$ and involving a standard Wiener process. Thus, we can derive a Gaussian perturbation kernel for this process following the general form defined in equation~\ref{eq:PerturbationKernel}:

\begin{align*}
    a(t) \rvx(0) & = \mathbb{E}[\rvx(t)|\rvx(0)] \\
                 & =  \mathbb{E}[\phi(t, 0) \rvx(0) |\rvx(0)] + \underbrace{\mathbb{E}\left[\int_0^t \phi(t, s) g(s) \, \dv \rvw(s) |\rvx(0)\right]}_{=0, \text{ since } \mathbb{E}[\dv \rvw] = 0} \\
                 & = \phi(t,0) \rvx(0),
\end{align*}
and
\begin{align*}
    b^2(t) & = \text{Var} \left( \rvx(t) |  \rvx(0)\right) \\
           & = \underbrace{\text{Var} \left( \phi(t, 0) \rvx(0) |\rvx(0) \right) }_{=0} + \text{Var} \left( \int_0^t \phi(t, s) g(s) \, \dv \rvw(s) |\rvx(0)\right) \\
           & = \E \left[ \left( \int_0^t \phi(t, s) g(s) \, \dv \rvw(s) \right)^2 | \rvx(0)\right] +  \underbrace{\left( \E \left[ \int_0^t \phi(t, s) g(s) \, \dv \rvw(s) | \rvx(0)\right] \right)^2}_{=0, \text{ since } \mathbb{E}[\dv \rvw] = 0} \\
           & \overset{\text{(Itô isometry)}}{=} \E \left[ \int_0^t \phi(t, s)^2 g(s)^2 \, \dv s | \rvx(0) \right] \\
           & = \int_0^t \phi(t, s)^2 g(s)^2 \, \dv s.
\end{align*}
Summarizing, the perturbation kernel parameters are given by:
\begin{align}
    \textcolor{black}{a(t)} & = \phi(t,0) = \exp\left( \int_0^t f(u) \, \dv u \right),
    \label{eq:MeankernelSDE}\\
    \textcolor{black}{b^2(t)} & = \int_0^t \phi(t, s)^2 g(s)^2 \, \dv s = \int_0^t \exp\left( 2 \int_s^t f(u) \, \dv u \right) g(s)^2 \, \dv s.
    \label{eq:VarkernelSDE}
\end{align}

\subsection{Derivation of the SDE for a given perturbation kernel}
\label{app:KerneltoSDE}
In the previous section in ~\ref{app:SDEtoKernel}, we derived a perturbation kernel with the general form specified in equation~\ref{eq:PerturbationKernel} from a given SDE. In this section, we do the opposite and derive the SDE that results in a given perturbation kernel, where we use the results from the previous section. Starting from the definition of $a(t)$ in equation~\ref{eq:MeankernelSDE} and $b^2(t)$ in equation~\ref{eq:VarkernelSDE}, we can first derive the drift $f(t)$ of the affine SDE (equation~\ref{eq:AffineForwSDE_si}):
\begin{align*}
    \exp\left( \int_0^t f(u) \, \dv u \right) & =    a(t) \\
\therefore \quad    f(t) & = \frac{\dv[\log a(t)]}{\dv t}  = \frac{\dot{a}(t)}{a(t)}.
\end{align*}
where $\dot{a}(t) = \frac{\dv [a(t)]}{\dv t}$ denotes the derivative of $a(t)$ with respect to time. Next, we derive the diffusion coefficient $g(t)$:
\begin{align*}
    b^2(t) & = \int_0^t \exp\left( 2 \int_s^t f(u) \, \dv u \right) g(s)^2 \, \dv s \\
           & = \int_0^t \exp\left( 2 \int_s^t \frac{\dv[\log{a(u)}]}{\dv u} \, \dv u \right) g(s)^2 \, \dv s \\
           & = \int_0^t \exp \left( 2 \left(\log a(t) -  \log a(s) \right) \right) g(s)^2 \, \dv s \\
           & = \int_0^t \frac{a(t)^2}{a(s)^2} \, g(s)^2 \, \dv s \\
           & = a^2(t) \int_0^t \frac{g(s)^2}{a(s)^2} \, \dv s \\
    \therefore \quad \left(\frac{b(t)}{a(t)}\right)^2 & = \int_0^t \frac{g(s)^2}{a(s)^2} \, \dv s.
\end{align*}
Differentiating both sides with respect to $t$ and solving for $g(t)$, we get
\begin{align*}
    \frac{g(t)^2}{a(t)^2} & = 2 \, \frac{b(t)}{a(t)} \, \frac{\dv}{\dv t} \left(\frac{b(t)}{a(t)}\right)  \\
\therefore \quad    g(t) & = \sqrt{2 \, a(t) \, b(t) \, \frac{\dv}{\dv t} \left(\frac{b(t)}{a(t)}\right)}.
\end{align*}
Thus, the SDE parameters are given by
\begin{align}
    \textcolor{black}{f(t)} &= \frac{\dv[\log a(t)]}{\dv t} = \frac{\dot{a}(t)}{a(t)},
    \label{eq:DriftSDE} \\
    \textcolor{black}{g(t)} &= \sqrt{2 \, a(t) \, b(t) \, \frac{\dv}{\dv t} \left(\frac{b(t)}{a(t)}\right)}.
    \label{eq:DiffusionSDE}
\end{align}
When defining the perturbation kernel to explicitly include a scaling factor, i.e., when  $b^2(t) = a^2(t) \, c(t)^2 $, $ p(\rvx(t)|\rvx(0)) = \mathcal{N}(\rvx(t);  a(t) \rvx_0, a^2(t) \, c(t)^2 \mI )$ and therefore $\tilde{\rvx}(t) = \left(\rvx(t) / a(t)\right) \sim \mathcal{N}(\rvx_0, c(t)^2 \mI )$, we get the special case of equation~\ref{eq:DiffusionSDE}:
\begin{align}
    g(t) = \sqrt{2 \, a(t) \, a(t) \, c(t) \, \frac{\dv}{\dv t} \left(\frac{a(t) \, c(t)}{a(t)}\right) } = a(t) \sqrt{2 \, c(t) \, \dot{c}(t)}
    \label{eq:ScaledDiffusionSDE}
\end{align}

\subsection{Derivation of our TV/SNR SDE}
\label{app:SNR-SDE}

Using the results from sections \ref{app:SDEtoKernel} and \ref{app:KerneltoSDE}, we derive our TV/SNR SDE. To this end, we first define the perturbation kernel as
 \begin{align}
  p(\rvx(t)|\rvx(0))
  &=
    \mathcal{N}\left(\rvx(t);  \sqrt{\frac{\tau^2(t) \gamma^2(t)}{1 + \gamma^2(t)} }\rvx(0) ,  \frac{\tau^2(t)}{1 + \gamma^2(t)} \mI \right).
    \label{eq:A.ProposedMarginal}
 \end{align}
Given the TV/SNR perturbation kernel in equation~\ref{eq:A.ProposedMarginal} and the results from \ref{app:KerneltoSDE}, we can derive the SDE that results in this perturbation kernel. First, using equation~\ref{eq:DriftSDE} we derive $f(t)$, where
\begin{equation*}
    a^2(t) = \frac{\tau^2(t) \gamma^2(t)}{1 + \gamma^2(t)}.
\end{equation*}

Abbreviating as $\tau$ = $\tau(t)$ and $\gamma = \gamma(t)$ to avoid clutter, we have
\begin{align*}
    \frac{\dv}{\dv t}a(t)^2 = 2 \dot{a}(t) a(t)
\end{align*}
and
\begin{align*}
\frac{\dv}{\dv t}a(t)^2 = \frac{\dv}{\dv t}  \left( \frac{\tau^2 \gamma^2}{1 + \gamma^2} \right) &=
\frac{(1 + \gamma^2)(2 \tau \dot{\tau} \gamma^2 + 2 \gamma \dot{\gamma} \tau^2) - \tau^2 \gamma^2 (2 \gamma \dot{\gamma})}{(1 + \gamma^2)^2} \\
&= \frac{2 \tau \dot{\tau} \gamma^2 (1 + \gamma^2) + 2 \gamma \dot{\gamma} \tau^2}{(1 + \gamma^2)^2}.
\end{align*}
Therefore
\begin{align*}
    \dot{a}(t) &= \frac{\frac{\dv}{\dv t} a^2(t)}{2 a(t)} \\
&= \frac{\tau \dot{\tau} \gamma^2 (1 + \gamma^2) + \gamma \dot{\gamma} \tau^2}{a(t) (1 + \gamma^2)^2}
\end{align*}
and
\begin{align*}
f(t) &= \frac{\dot{a}(t)}{a(t)} \\
&= \frac{\tau \dot{\tau} \gamma^2 (1 + \gamma^2) + \gamma \dot{\gamma} \tau^2}{a^2(t) (1 + \gamma^2)^2} \\
&= \frac{\tau \dot{\tau} \gamma^2 (1 + \gamma^2) + \gamma \dot{\gamma} \tau^2}{\tau^2 \gamma^2 (1 + \gamma^2)}.
\end{align*}
Consequently we have
\begin{align}
    f(t) &= \frac{\dot{\tau}(t)}{\tau(t)} + \frac{\dot{\gamma}(t)}{\gamma(t) \left( 1 + \gamma^2(t) \right)}.
\end{align}
Now, we can derive the diffusion coefficient $g(t)$, where we can use the special case of equation~\ref{eq:DiffusionSDE}, when the variance is explicitly scaled by the mean factor $a(t)$:
\begin{align*}
    b^2(t) = a^2(t) \, c(t)^2 & = \frac{\tau^2(t) \gamma^2(t)}{1 + \gamma^2(t)} \, \frac{1}{\gamma^2(t)} \\
    \Rightarrow     c(t) & = \gamma(t)^{-1} .
\end{align*}
Thus, we use equation~\ref{eq:ScaledDiffusionSDE} to solve for $g(t)$, where
\begin{align*}
    \dot{c}(t) = \frac{\mathrm{d}}{\mathrm{d}t} \left( \gamma(t)^{-1} \right) = - \frac{\dot{\gamma}(t)}{\gamma^2(t)}
\end{align*}
and hence
\begin{align}
    g(t) &= a(t) \sqrt{2 \, c(t) \, \dot{c}(t)} \notag \\
    &= \sqrt{ \frac{\tau^2(t) \gamma^2(t)}{1 + \gamma^2(t)}} \sqrt{-2 \frac{1}{\gamma(t)}\frac{\dot{\gamma}(t)}{\gamma^2(t)}} \notag \\
    &= \sqrt{\frac{-2 \tau^2(t) \dot{\gamma}(t)}{\gamma(t) \left( 1 + \gamma^2(t) \right)}}.
\end{align}
Note that $\gamma(t)$ needs to be differentiable, monotonically decreasing and positive for all $t \in [0, 1]$ to ensure that the SDE is well-defined, i.e., the square root in the diffusion coefficient $g(t)$ is well-defined and the dominant term in the drift $f(t)$ is non-zero.

\subsection{\rebuttal{The TV schedule and truncation error}}
\label{app:LTE}

\rebuttal{This section provides a decomposition of the trajectory acceleration in our disentangled TV/SNR coordinates in order to relate the TV schedule to the truncation error of a first-order solver. We assume the exact score, a uniform time grid, and twice continuously differentiable trajectories.}

\rebuttal{\paragraph{1. The local and global truncation error.} When generating samples, we integrate the probability flow ODE, by setting $\lambda = 0$ in equation~\ref{eq:GenRevSDE}:}
\begin{align}
    \dot{\rvx}(t) := f(t) \rvx - \tfrac{1}{2} g^2(t) \, \rvs(\rvx, t),
    \qquad \rvs(\rvx,t) := \nabla_{\rvx} \log p_t(\rvx).
    \label{eq:A.PFODE}
\end{align}
\rebuttal{Using Taylor expansion with an integral remainder for an Euler step of size $h > 0$ backward from $t$ to $t-h$, we have:}
\begin{align}
    \rvx(t - h) = \rvx(t) - h \dot{\rvx}(t) + \int_{t-h}^{t} \bigl(s - t + h\bigr) \, \ddot{\rvx}(s) \dv s .
\end{align}
\rebuttal{The local truncation error $\bm{\ell}(t)$ of this single step is bounded by the maximum trajectory acceleration: $\lVert \bm{\ell}(t) \rVert \leq \frac{h^2}{2} \sup_{s \in [t-h, t]} \lVert \ddot{\rvx}(s) \rVert$. For $N$ uniform steps ($h = 1/N$), the sum of these local errors is bounded by $\frac{1}{2N}\int_0^1 \lVert\ddot{\rvx}(t)\rVert \dv t$ at leading order in $1/N$.}

\rebuttal{Considering that local errors are propagated and potentially magnified by the remaining reverse process, and denoting the Jacobian of the exact reverse flow from $t$ to $0$ by $\mJ(0,t) = \partial \rvx(0) / \partial \rvx(t)$, the global error between the exact ground truth $\rvx(0)$ and its $N$-step numerical approximation $\rvx_N$ satisfies:}
\begin{align}
    \lVert \rvx_N - \rvx(0) \rVert \;\lesssim\; \frac{1}{2N} \int_0^1 \lVert \mJ(0,t) \rVert \, \lVert \ddot{\rvx}(t) \rVert \dv t .
    \label{eq:A.GlobalError}
\end{align}
\rebuttal{This shows that the truncation error bound grows with the acceleration of the trajectory. As discussed in section~\ref{sec:Discussion}, mitigating the acceleration near the data manifold ($t \approx 0$) is crucial, because errors made at the very end of the generative process are directly imprinted onto the final sample, as there is no remaining flow to correct them.}

\rebuttal{\paragraph{2. Decoupling the dynamics via the normalized state.} To understand how the TV schedule $\tau(t)$ drives this acceleration, and therefore the error bound, independently of the SNR $\gamma(t)$ in our disentangled framework, we factor out the TV by using the variance normalized state $\hat{\rvx}(t) = \rvx(t)/\tau(t)$. In the TV/SNR framework, the drift is $f(t) = f_\tau(t) + f_\gamma(t)$, and the diffusion coefficient is $g^2(t) = -2\tau^2(t) f_\gamma(t)$, where $f_\tau = \dot{\tau}/\tau$, and $f_\gamma = \dot{\gamma} / \left(\gamma \left( 1 + \gamma^2 \right)\right)$. Substituting these into the reverse ODE gives:}
\begin{align}
    \dot{\rvx} = (f_\tau + f_\gamma)\rvx + \tau^2 f_\gamma \rvs .
\end{align}
\rebuttal{Taking the time derivative of the normalized state $\hat{\rvx}$:}
\begin{align}
    \dot{\hat{\rvx}} = \frac{\dv}{\dv t}\left(\frac{\rvx}{\tau}\right) &= \frac{\dot{\rvx}}{\tau} - \frac{\dot{\tau}}{\tau^2}\rvx = \frac{\dot{\rvx}}{\tau} - f_\tau \hat{\rvx} \notag \\
    &= \frac{1}{\tau} \bigl[ (f_\tau + f_\gamma)\tau\hat{\rvx} + \tau^2 f_\gamma \rvs \bigr] - f_\tau \hat{\rvx} \notag \\
    &= f_\gamma \hat{\rvx} + \tau f_\gamma \rvs,
\end{align}
\rebuttal{and defining the normalized score $\hat{\rvs}(\hat{\rvx}, t) = \tau \rvs(\rvx, t)$, the ODE for the normalized state simplifies to:}
\begin{align}
    \dot{\hat{\rvx}}(t) = f_\gamma(t) \bigl( \hat{\rvx} + \hat{\rvs} \bigr) .
    \label{eq:A.NormalizedODE}
\end{align}
\rebuttal{Crucially, the normalized score does not depend on $\tau(t)$. By the change of variables $\hat{\rvx} = \rvx/\tau$, the density of the normalized state is $\hat{p}_t(\hat{\rvx}) = \tau^d p_t(\tau\hat{\rvx})$, so that $\hat{\rvs} = \nabla_{\hat{\rvx}} \log \hat{p}_t(\hat{\rvx})$ is the score of the normalized marginal. By equation~\ref{eq:A.ProposedMarginal}, the perturbation kernel of $\hat{\rvx}(t)$ is $\mathcal{N}\bigl(\hat{\rvx}(t); \gamma(1+\gamma^2)^{-1/2} \rvx(0), (1+\gamma^2)^{-1} \mI\bigr)$, which is the exact kernel of a constant TV process with an identical SNR schedule. Consequently, the dynamics of $\hat{\rvx}(t)$ are governed \emph{only} by the SNR schedule $\gamma(t)$ and the data distribution. A schedule and its constant TV counterpart generate identical \emph{normalized} trajectories. In particular, the SNR-induced acceleration $\ddot{\hat{\rvx}}$ is strictly independent of the TV schedule.}

\rebuttal{\paragraph{3. Exact acceleration decomposition.} The acceleration of the actual state $\rvx = \tau \hat{\rvx}$ is:}
\begin{align}
    \ddot{\rvx} &= \frac{\dv}{\dv t}\bigl( \dot{\tau}\hat{\rvx} + \tau\dot{\hat{\rvx}} \bigr) \notag \\
                &= \ddot{\tau}\hat{\rvx} + 2\dot{\tau}\dot{\hat{\rvx}} + \tau\ddot{\hat{\rvx}} .
\end{align}
\rebuttal{Substituting $\hat{\rvx} = \rvx/\tau$ gives the acceleration formula for all $t$:}
\begin{align}
    \ddot{\rvx}(t) = \underbrace{\frac{\ddot{\tau}}{\tau}\rvx(t) + 2f_\tau \tau \dot{\hat{\rvx}}(t)}_{\text{TV-induced acceleration}} \; + \; \underbrace{\tau \ddot{\hat{\rvx}}(t)}_{\text{SNR-induced acceleration}}
    \label{eq:A.ExactDecomposition}
\end{align}
\rebuttal{ which is decomposed into a TV-induced acceleration and an SNR-induced acceleration. Although the variables mix, this grouping isolates the origin of the acceleration: the first two terms exist if and only if the TV is dynamic ($\dot{\tau} \neq 0$ or $\ddot{\tau} \neq 0$), whereas $\tau \ddot{\hat{\rvx}}$ is the intrinsic baseline acceleration required by the SNR, which is independent of the TV schedule (equation~\ref{eq:A.NormalizedODE}), and merely scaled to the correct spatial dimension.}

\rebuttal{Now, we can observe how errors propagate in these normalized coordinates. Since $\rvx(0) = \tau(0) \hat{\rvx}(0)$, the Jacobian of the reverse flow factorizes as $\mJ(0,t) = \frac{\tau(0)}{\tau(t)} \hat{\mJ}(0,t)$. Inserting this factorization and equation~\ref{eq:A.ExactDecomposition} into the global error bound (equation~\ref{eq:A.GlobalError}) gives:}
\begin{align}
    \lVert \rvx_N - \rvx(0) \rVert \;\lesssim\; \frac{\tau(0)}{2N} \int_0^1 \lVert \hat{\mJ}(0,t) \rVert \left( \left\lvert \frac{\ddot{\tau}}{\tau} \right\rvert \lVert \hat{\rvx} \rVert + 2 \lvert f_\tau \rvert \, \lVert \dot{\hat{\rvx}} \rVert + \lVert \ddot{\hat{\rvx}} \rVert \right) \dv t .
    \label{eq:A.NormalizedGlobalError}
\end{align}
\rebuttal{Notice that the TV $\tau(t)$ within the acceleration perfectly cancels the $1/\tau(t)$ term originating from the Jacobian. The consequences of this bound are summarized below.}

\rebuttal{Equations~\ref{eq:A.ExactDecomposition} and~\ref{eq:A.NormalizedGlobalError} lead to the following conclusions:}

\rebuttal{\emph{(i) The TV schedule enters the error bound only through its logarithmic growth rate and its curvature.} In equation~\ref{eq:A.NormalizedGlobalError}, the TV schedule appears only through the dimensionless coefficients $f_\tau = \frac{\dv}{\dv t} \log \tau$ and $\ddot{\tau}/\tau = \dot{f}_\tau + f_\tau^2$, whereas the remaining factors $\lVert \hat{\mJ} \rVert$, $\lVert \hat{\rvx} \rVert$, $\lVert \dot{\hat{\rvx}} \rVert$ and $\lVert \ddot{\hat{\rvx}} \rVert$ are identical for a schedule and its constant TV counterpart. For a constant TV, both coefficients vanish identically. For VE schedules, $\tau^2 = 1 + \sigma^2$, and for the exponential noise level of SMLD, $\sigma(t) = \sigma_{\mathrm{min}} e^{rt}$ with $r = \log(\sigma_{\mathrm{max}}/\sigma_{\mathrm{min}})$, one finds $f_\tau = r u$ and $\ddot{\tau}/\tau = r^2 u(2-u)$, where $u = \sigma^2/(1+\sigma^2) \in (0,1)$. Both coefficients thus approach $r$ and $r^2$ once $\sigma(t) \gtrsim 1$, which increases the error bound. For the bounded TV of OTFM, $\tau^2(t) = (1-t)^2 + t^2$, in contrast, $f_\tau = (2t-1)/\tau^2 \in [-1, 1]$ and $\ddot{\tau}/\tau = \tau^{-4} \in [1, 4]$ are bounded independently of the endpoints.}

\rebuttal{\emph{(ii) In continuous time, the TV schedule does not affect the generated samples.} By paragraph 2, a schedule and its constant TV counterpart generate identical normalized trajectories. For the same normalized initial state $\hat{\rvx}(1)$, both therefore result in $\rvx(0) = \tau(0) \hat{\rvx}(0)$ with the same $\hat{\rvx}(0)$, where $\tau(0) \approx 1$ for all schedules in table~\ref{tab:ExistingTVSNR}. Any difference in sample quality between a schedule and its constant TV counterpart is thus a discretization effect.}

\rebuttal{\emph{(iii) Limitations.} Equation~\ref{eq:A.NormalizedGlobalError} is an upper bound. While the large coefficients of exploding TV schedules show that such schedules can admit a much larger error bound than their constant TV counterparts, especially if the bound is tight, the bound itself does not determine whether a \emph{bounded} non-constant TV, such as that of OTFM, leads to a larger error than a constant TV, because this comparison may depend on the data distribution, and the TV-induced and SNR-induced accelerations in equation~\ref{eq:A.ExactDecomposition} may partially cancel. A constant TV is therefore a simple, modality-agnostic default, whereas a suitable non-constant bounded TV may have to be carefully adapted to the data distribution. Therefore, we further assess this using the empirical analysis. Moreover, $f_\tau$ and $\ddot{\tau}/\tau$ are derivatives with respect to the variable in which the ODE is integrated and change under a reparameterization of time, so that the same continuous process, such as EDM and its uniform-time variant EDM-UT, can have different error bounds under discretization.}

\rebuttal{\paragraph{4. Relation to the velocity deviation metric in section~\ref{sec:Discussion}.} Let $\bar{\rvv} = \rvx(1) - \rvx(0) = \int_0^1 \dot{\rvx}(t) \dv t$ denote the mean velocity over the unit time interval. The integrand of our empirical metric $\mathcal{C}(t)$ (equation~\ref{eq:CurvatureMetric}) represents the deviation of the instantaneous velocity $\dot{\rvx}(t)$ from this mean. Defining the zero-mean function $\rvu(t) = \dot{\rvx}(t) - \bar{\rvv}$ on $[0,1]$ allows us to apply the Poincar\'e--Wirtinger inequality ($\int_0^1 \lVert \rvu \rVert^2 \dv t \leq \pi^{-2} \int_0^1 \lVert \dot{\rvu} \rVert^2 \dv t$). With $\dot{\rvu} = \ddot{\rvx}$, we obtain:}
\begin{align}
    \int_0^1 \lVert \dot{\rvx}(t) - \bar{\rvv} \rVert^2 \dv t \;\leq\; \frac{1}{\pi^2} \int_0^1 \lVert \ddot{\rvx}(t) \rVert^2 \dv t
\end{align}
\rebuttal{Taking expectations over the prior results in equation~\ref{eq:CurvatureLTEBound}, whenever the right-hand side is finite. Since $\int_0^1 \lVert \ddot{\rvx} \rVert \dv t \leq \bigl( \int_0^1 \lVert \ddot{\rvx} \rVert^2 \dv t \bigr)^{1/2}$ by the Cauchy--Schwarz inequality, a small mean squared acceleration implies both a small bound on the sum of the local errors (paragraph 1) and a small integrated velocity deviation, while a large integrated velocity deviation implies a large mean squared acceleration. 
Moreover, if the acceleration remains moderate along the trajectory, a large integrated velocity deviation also implies a large bound on the sum of the local errors, which translates into a large truncation error to the extent that the error bound is tight.
Note that the above inequality relates time-integrated quantities and does not localize $\mathcal{C}(t)$ at a particular $t$. Therefore, $\mathcal{C}(t)$ serves as a qualitative diagnostic in section~\ref{sec:Discussion}.}

\section{Experimental details}
\label{sec:A.AlgorithmDetails}

\subsection{\rebuttal{Training details}}
For training a diffusion model, we use the loss in equation~\ref{eq:DSM} to keep a unit variance of the model output for all $t$, and adopt the same noise model architecture $\rvepsilon_{\theta}$ from Kahouli~\etal~\cite{kahouli2024morered}, but use 9 interaction blocks, train on continuous time and condition the model on a scaled SNR instead of the time $t$, i.e., $\rvepsilon_{\theta}(\hat{\rvx}(t), c_\mathrm{snr}(\gamma^2(t)))$, where $\hat{\rvx}(t)$ is a scaled version of $\rvx(t)$ to unit variance. This is achieved by first scaling the training data by $\sigma_\mathrm{data}$, which is approximately $\sqrt{2}$ for the QM9 dataset, and always setting $\tau(t)=1$ during training, independent of the training SNR schedule. This has the benefit of making the model compatible with various TV and SNR schedules during sampling without retraining, and avoiding model stability issues due to large cutoff distances in the Graph Neural Network when using non-constant $\tau(t)$. We define $c_\mathrm{snr}(\gamma^2(t)) = \omega \log(\gamma^2(t)) + \xi$ to linearize the SNR input, keeping it in a stable, normalized range, with $\omega=-0.125$ and $\xi=0.35$ providing good performance.
During sampling with a TV schedule $\tau(t) \neq 1$, we scale the model input to $\hat{\rvx}(t) = \tau(t)^{-1} \rvx(t)$ to maintain unit variance for all $t$. Note that the reverse trajectory itself will not become constant. The generated samples $\rvx(0)$ are then scaled back to the target data variance by multiplying by $\sigma_\mathrm{data}$.

We tuned $t_\mathrm{max}$ such that $\gamma(t)^{-1}$ approximates the dataset's maximum pairwise Euclidean distance between molecules. This ensures that all the modes of the distribution are mixed at $t_\mathrm{max}$. We tune $t_\mathrm{min}$ to the largest value producing almost noiseless samples, avoiding extra reverse steps near the data manifold. \rebuttal{In~\ref{sec:Endpoints}, we provide a detailed discussion of the choice of $t_\mathrm{min}$ and $t_\mathrm{max}$}.

We trained two models using different schedules: (i) DDPM-cos with $\nu=1.0$ and (ii) the EDM SNR schedule with $\tau(t)=1$ for the reasons discussed before. We then sampled from each model using all schedules and found that the model trained with DDPM-cos consistently outperformed the EDM-trained model, even when using the EDM schedule for sampling, as depicted in figure~\ref{fig:AllComparisonMolecule_EulerODE_AnotherTraining}. Therefore, we report only the results using the model trained on the DDPM-cos in the main text. \rebuttal{In~\ref{sec:A.AdditionalExperimentalResults}, we provide results for the model trained on the EDM schedule, and a possible explanation for the performance difference between the two models.}

\subsection{\rebuttal{Definition of the performance metrics}}
\label{sec:StabilityDefinition}

For stability, uniqueness and novelty, we adopt the criteria introduced by Gebauer~\etal~\cite{gebauer2022inverse} and use their released implementation without modification, so that our numbers are directly comparable to previous works. For validity, we follow the RDKit-based criterion of Hessmann~\etal~\cite{hessmann2026generative}:

\rebuttal{\emph{Stability.} Gebauer~\etal~\cite{gebauer2022inverse} first assign bonds to each generated 3D point cloud. Instead of using distance cutoffs, the point cloud is converted into an Open Babel molecule, and connectivity together with bond orders are then inferred by Open Babel, where bonds perceived between two hydrogen atoms are discarded.}
\rebuttal{After inferring bonds, valency is checked. A structure satisfies the valence rule if the perceived bond orders saturate the target valence of every atom exactly, where hydrogen atoms are counted like any other atom, without implicit-hydrogen treatment or tolerance parameter. Since Open Babel's bond assignment is sensitive to the order in which atoms are supplied, the procedure is retried up to 10 times with randomly permuted atom orderings before a structure is rejected. Moreover, a breadth-first search over the perceived bond graph must reach every atom, structures that decompose into more than one fragment are rejected.}

\rebuttal{\emph{Uniqueness.} Two stable structures are considered duplicates if their Open Babel FP2 fingerprints have Tanimoto similarity $1$ and their canonical SMILES strings agree. The comparison is also made against the mirror image of each canonical SMILES, so enantiomers are counted as duplicates.}

\rebuttal{\emph{Novelty.} Each unique, stable structure is compared with every molecule of the same composition in the QM9 training and validation splits, using the duplicate criterion above, i.e.\ identical FP2 fingerprints (Tanimoto similarity $1$) and identical canonical SMILES, with the mirror image counted as a match. A structure is novel if it matches no molecule of the training or validation split.}

\rebuttal{\emph{Validity.} Connectivity and bond orders are assigned from the coordinates by RDKit's bond perception, and a structure is valid if such an assignment exists, no atom is left with a radical electron, and the structure forms a single fragment. Stability and validity therefore ask the same question, namely whether a point cloud corresponds to a connected molecule in which every atom has a chemically sensible valence, but they answer it with different bond perception: stability checks the valence of each atom on the bond orders perceived by Open Babel, whereas validity requires RDKit to find a globally consistent assignment of bond orders. RDKit's connectivity perception is a stricter criterion. This is apparent on the molecules of the QM9 test split, which are DFT-optimized equilibrium structures: $99.7\,\%$ of them are stable, but only $94.5\,\%$ are valid, and in each of the molecules that are stable but not valid, RDKit perceives a bond that Open Babel does not. Conversely, every valid test molecule is also stable. Validity values should therefore be read against a ceiling of about $95\,\%$ rather than $100\,\%$, as these structures were not removed from training in our case.}

\rebuttal{Because this geometry-based criterion is a necessary but not always sufficient condition for physical plausibility, the DFT relaxations reported in~\ref{sec:DFTcalculations} provide the complementary, physically grounded assessment.}

\subsection{\rebuttal{Data splits and molecular compositions}}
\label{sec:Splits}

\rebuttal{We use the train/validation/test split of QM9 established by previous work~\cite{gebauer2022inverse, kahouli2024morered}: a random split at the level of molecules, with $55\,000$ training, $10\,000$ validation and $10\,000$ test structures. The three subsets are drawn by a random permutation of the first $75\,000$ entries of the database obtained after removing the $3\,054$ molecules flagged as uncharacterized in the original release~\cite{qm9}, which should not be confused with the unstable and invalid molecules discussed above. All models compared here use the identical split, and the maximum pairwise distance used to set the schedule endpoints (\ref{sec:Endpoints}) is computed on the training split alone.}

\rebuttal{Because the split is random over molecules and not over compositions, molecular formulas are shared between the splits. Our task is the prediction of an equilibrium geometry, and the stability and RMSD numbers reported here should not be read as evidence for generalization to unseen compositions. A formula-disjoint evaluation would be required to assess that, but would test a different capability than the one this work is concerned with.}

\subsection{\rebuttal{Schedule endpoints and mode coverage}}
\label{sec:Endpoints}

\rebuttal{The rationale for the choice of $\gamma_{\mathrm{min}}$ follows the standard argument of Song~\etal~\cite{song2021score}: the terminal noise level must be large enough that the Gaussian kernels placed on the individual data points overlap into a single mode, so that the prior is well approximated by an isotropic Gaussian. It therefore suffices that the terminal noise standard deviation $1/\gamma_{\mathrm{min}}$ be at least of the order of the largest distance between data points. For $\gamma_{\mathrm{max}}$ we require that the initial noise level be small enough that the reverse process can reach the data manifold in a finite number of steps.}

\rebuttal{For the QM9 training split, the largest distance between training molecules is almost $30$\,\AA. In the standardized coordinates in which the model operates (the data are scaled by $\sigma_{\mathrm{data}} = \sqrt{2}$, see~\ref{sec:A.AlgorithmDetails}) this corresponds to $21$. For $\gamma_{\mathrm{max}}$, we empirically found that $1/\gamma_{\mathrm{max}} = 10^{-3}$, which is small enough that further decreasing it does not change the properties of the structure. From $\gamma_{\mathrm{min}}$ and $\gamma_{\mathrm{max}}$, we compute the corresponding $t_\mathrm{min}$ and $t_\mathrm{max}$ for each schedule.}
\rebuttal{For images, for every setting, $t_{\mathrm{min}}$ and $t_{\mathrm{max}}$ are determined from $(\eta,\kappa)$ so that the schedule spans the noise range of the pre-trained models, $\sigma \in [0.002, 80]$, i.e.\ $\gamma_{\mathrm{max}} = 1/0.002$ and $\gamma_{\mathrm{min}} = 1/80$. The tuned as well as the fixed hyperparameter settings $(\eta,\kappa)$ are detailed in~\ref{sec:BOdetails}.}

\rebuttal{While the comparisons in this paper cannot be attributed to endpoint selection, because they were fixed through schedules, a systematic study of the endpoints themselves, including the quantification of the terminal-SNR mismatch and its interaction with $\eta$ and $\kappa$, is an important axis of analysis that we identify as future work.}

\subsection{\rebuttal{VP-ISSNR hyperparameters and Bayesian optimization}}
\label{sec:BOdetails}

\rebuttal{
This section documents the protocol used to obtain the VP-ISSNR$[\eta,\kappa$ tuned$]$ results reported in Figure~\ref{fig:ComparisonImage}, Figure~\ref{fig:ImagesDatasetsComparison} and Figure~\ref{fig:SolversComparison} for images. Only the two free parameters of the ISSNR schedule in \eqref{eq:SigmoidSNRScheduling}, $\eta>0$ and $\kappa\in\mathbb{R}$, are optimized. Both parameters are searched over a bounded box with a uniform prior as specified in table~\ref{tbl:hyperopt_params}. The ranges were fixed once, before running any of the reported sweeps, and were kept identical for all datasets. We run Bayesian Optimization with 32 trial runs independently for each combination of NFE and dataset.

\begin{table}
\centering
\begin{tabular}{lccl}
\hline
Parameter & Distribution & Range \\
\hline
$\eta$    & uniform & $[0.5,\ 4.0]$  \\
$\kappa$  & uniform & $[-3.0,\ 3.0]$  \\
\hline
\end{tabular}
\caption{Search space for Bayesian Optimization of parameters $\eta$ and $\kappa$}
\label{tbl:hyperopt_params}
\end{table}

We use the Bayesian optimization implementation of Weights \& Biases sweeps, which fits a Gaussian-process surrogate to the observed $(\eta,\kappa)\mapsto\text{FID}$ pairs and selects the next candidate by maximizing expected improvement, with a small probability of a random draw for exploration. The objective is the FID computed from
$50\,000$ generated images against the official EDM reference statistics for the
corresponding dataset\footnote{\url{https://nvlabs-fi-cdn.nvidia.com/edm/fid-refs/}}. This metric is evaluated for each of the trial runs. To avoid selection bias, we use a distinct set of seeds ($100\,000-149\,999$) for Bayesian Optimization, that does not overlap with the seeds used during evaluation of the tuned schedules, and for comparison to baselines. The first trial of every sweep is pinned to $(\eta,\kappa)=(1,0)$, i.e.\ to the VP-OTFM special case of our schedule, to seed the search at a meaningful reference point. The stopping rule is the fixed budget of 32 trial runs. Because each trial requires the generation of $50\,000$ samples, this procedure is computationally expensive, but comparable to hyperparameter tuning in previous work, where for instance the time grid or the noise level $\sigma$ were tuned for each dataset. Afterwards, we choose the configuration with the lowest FID score for further evaluation. The selected configurations were re-evaluated over several runs on seeds disjoint from the tuning seeds (table~\ref{tab:ImageFID}).}

\rebuttal{Table~\ref{tbl:hyperopt_values} lists the resulting hyperparameters for every dataset, solver and NFE, alongside the two untuned settings used elsewhere in this work: the $\eta$-scaled variant and the fixed-hyperparameter variant. For every setting, $t_{\mathrm{min}}$ and $t_{\mathrm{max}}$ are then determined from $(\eta,\kappa)$ so that the schedule spans the noise range of the pre-trained models, $\sigma \in [0.002, 80]$, i.e.\ $\gamma_{\mathrm{max}} = 1/0.002$ and $\gamma_{\mathrm{min}} = 1/80$.}

\begin{table}
\centering
\caption{\rebuttal{Hyperparameters of the ISSNR schedule for images. The upper block gives the values found by Bayesian optimization for each dataset, solver and NFE (VP-ISSNR$[\eta,\kappa$ tuned$]$). The middle block gives the two untuned settings, which use the same values for every dataset. The lower separate block gives the values for the adaptive RK45 solver, whose NFE is set by the solver and is therefore reported as the average NFE over the generated samples.}}
\label{tbl:hyperopt_values}
\footnotesize
\setlength{\tabcolsep}{3.5pt}
\begin{tabular}{@{}llcccccccccc@{}}
\br
 &  & \multicolumn{2}{c}{NFE 3} & \multicolumn{2}{c}{NFE 7} & \multicolumn{2}{c}{NFE 15} & \multicolumn{2}{c}{NFE 31} & \multicolumn{2}{c}{NFE 63} \\
Dataset & Solver & $\eta$ & $\kappa$ & $\eta$ & $\kappa$ & $\eta$ & $\kappa$ & $\eta$ & $\kappa$ & $\eta$ & $\kappa$ \\
\mr
\multicolumn{12}{l}{\emph{Bayesian optimization} (VP-ISSNR$[\eta,\kappa$ tuned$]$)} \\
CIFAR-10 & Heun  & 1.04 & $-0.06$ & 0.82 & 0.35 & 1.38 & 0.66 & 2.17 & $-0.47$ & 2.96 & 2.61 \\
CIFAR-10 & Euler & 2.15 & 1.05 & 0.97 & 0.26 & 1.00 & 0.00 & 0.88 & 0.13 & 0.92 & 0.29 \\
CIFAR-10 & DPM-Solver & 2.24 & $-3.00$ & 2.24 & $-2.35$ & 3.86 & $-2.40$ & 4.00 & $-0.22$ & 3.95 & 2.97 \\
FFHQ & Heun & 1.76 & $-0.78$ & 0.85 & $-0.05$ & 1.31 & 0.68 & 1.76 & 0.01 & 1.79 & $-0.33$ \\
AFHQv2 & Heun & 3.21 & $-2.58$ & 1.03 & $-0.69$ & 1.57 & $-0.04$ & 2.15 & $-0.82$ & 2.15 & $-0.82$ \\
ImageNet & Heun & 3.15 & $-2.44$ & 0.85 & $-0.05$ & 0.90 & $-0.09$ & 1.82 & 0.04 & 1.93 & $-0.39$ \\
\mr
\multicolumn{12}{l}{\emph{Untuned settings} (all image datasets and solvers)} \\
$\eta$ scaled & --- & 1.00 & 0.00 & 1.00 & 0.00 & 1.50 & 0.00 & 2.00 & 0.00 & 2.50 & 0.00 \\
fixed & --- & 1.50 & 1.00 & 1.50 & 1.00 & 1.50 & 1.00 & 1.50 & 1.00 & 1.50 & 1.00 \\
\br
\\
\br
 &  & \multicolumn{2}{c}{NFE 14} & \multicolumn{2}{c}{NFE 51} & \multicolumn{2}{c}{NFE 60} & & & & \\
Dataset & Solver & $\eta$ & $\kappa$ & $\eta$ & $\kappa$ & $\eta$ & $\kappa$ & & & & \\
\mr
\multicolumn{12}{l}{\emph{Bayesian optimization, adaptive solver} (VP-ISSNR$[\eta,\kappa$ tuned$]$)} \\
CIFAR-10 & RK45 & 2.45 & 1.90 & 2.53 & 2.20 & 1.21 & 0.16 & & & & \\
\br
\end{tabular}
\end{table}

\section{Additional experimental results}
\label{sec:A.AdditionalExperimentalResults}

\subsection{Molecular structure generation}
\label{sec:A.AdditionalExperimentalResultsMols}

% ---- Stability rate ----
\begin{table}
\centering
\caption{\rebuttal{Stability rate (\%) on QM9 as a function of the NFE, for the reverse ODE and the reverse SDE with the Euler solver. Each entry is the mean over five independent sampling runs with different random seeds, with $\pm 1$ standard deviation over those seeds as a subscript. Within each column and ODE/SDE block, every value whose $\pm 1$ standard deviation interval overlaps that of the column's largest value is set in bold. In the low NFE regime, our method is always outperforming the baselines. Binomial confidence intervals for these values are given in table~\ref{tab:StabilityBinomialCI}.}}
\label{tab:MetricsStability}
\footnotesize
\setlength{\tabcolsep}{3pt}
\begin{tabular}{@{}lccccccc@{}}
\br
Schedule & \textbf{4} & \textbf{8} & \textbf{16} & \textbf{32} & \textbf{64} & \textbf{128} & \textbf{256} \\
\mr
\multicolumn{8}{l}{\emph{Reverse ODE}} \\
EDM & 0.3$_{\pm0.1}$ & 15.8$_{\pm0.6}$ & 55.5$_{\pm1.0}$ & 76.2$_{\pm1.2}$ & 82.6$_{\pm1.1}$ & 85.2$_{\pm0.8}$ & 86.2$_{\pm0.5}$ \\
EDM-UT & 0.0$_{\pm0.0}$ & 0.0$_{\pm0.0}$ & 43.0$_{\pm0.9}$ & 81.3$_{\pm0.5}$ & \textbf{86.1}$_{\pm0.8}$ & \textbf{87.6}$_{\pm0.7}$ & \textbf{87.8}$_{\pm0.6}$ \\
SMLD & 0.0$_{\pm0.0}$ & 0.1$_{\pm0.0}$ & 0.9$_{\pm0.1}$ & 75.8$_{\pm0.6}$ & \textbf{86.8}$_{\pm0.6}$ & \textbf{88.0}$_{\pm0.4}$ & \textbf{88.3}$_{\pm0.6}$ \\
OTFM & 2.5$_{\pm0.7}$ & 28.3$_{\pm0.5}$ & 65.3$_{\pm0.8}$ & 79.9$_{\pm0.6}$ & 84.9$_{\pm0.6}$ & 86.3$_{\pm0.9}$ & \textbf{86.9}$_{\pm0.9}$ \\
VP-EDM-UT & 0.0$_{\pm0.0}$ & 42.9$_{\pm0.6}$ & 72.4$_{\pm0.7}$ & 81.1$_{\pm0.6}$ & 84.6$_{\pm1.0}$ & 86.1$_{\pm0.5}$ & 86.7$_{\pm0.5}$ \\
VP-SMLD & 0.2$_{\pm0.1}$ & 53.3$_{\pm1.0}$ & 74.9$_{\pm0.3}$ & 82.0$_{\pm0.6}$ & 85.0$_{\pm0.8}$ & 86.2$_{\pm0.5}$ & 86.8$_{\pm0.5}$ \\
VP-OTFM & 1.3$_{\pm0.2}$ & 30.0$_{\pm0.7}$ & 66.9$_{\pm0.8}$ & 80.4$_{\pm1.2}$ & 84.7$_{\pm0.6}$ & 86.4$_{\pm0.4}$ & 86.8$_{\pm0.6}$ \\
DDPM-linear & 0.0$_{\pm0.0}$ & 0.9$_{\pm0.2}$ & 18.2$_{\pm1.0}$ & 53.6$_{\pm0.3}$ & 74.1$_{\pm0.8}$ & 81.8$_{\pm0.6}$ & 85.1$_{\pm0.7}$ \\
DDPM-cos [$\nu=2.5$] & 1.6$_{\pm0.1}$ & 37.9$_{\pm1.1}$ & 67.7$_{\pm0.6}$ & 79.7$_{\pm1.3}$ & 84.6$_{\pm0.6}$ & 86.2$_{\pm0.6}$ & 86.9$_{\pm0.7}$ \\
DDPM-cos [$\nu=1.0$] & 0.0$_{\pm0.0}$ & 4.4$_{\pm0.2}$ & 38.4$_{\pm1.5}$ & 68.3$_{\pm1.1}$ & 80.3$_{\pm0.6}$ & 84.2$_{\pm0.3}$ & 86.1$_{\pm0.8}$ \\
\textbf{VP-ISSNR (ours)} & \textbf{36.1}$_{\pm0.5}$ & \textbf{73.5}$_{\pm0.6}$ & \textbf{82.4}$_{\pm0.6}$ & \textbf{85.0}$_{\pm0.6}$ & \textbf{86.3}$_{\pm0.6}$ & 86.9$_{\pm0.5}$ & \textbf{87.2}$_{\pm0.7}$ \\
\mr
\multicolumn{8}{l}{\emph{Reverse SDE}} \\
VP-EDM-UT & 0.0$_{\pm0.0}$ & 0.8$_{\pm0.1}$ & 71.3$_{\pm0.9}$ & 90.4$_{\pm0.4}$ & \textbf{93.4}$_{\pm0.5}$ & \textbf{94.8}$_{\pm0.2}$ & \textbf{94.6}$_{\pm0.4}$ \\
VP-SMLD & 0.0$_{\pm0.0}$ & 5.6$_{\pm0.4}$ & 79.1$_{\pm0.4}$ & 90.9$_{\pm0.4}$ & \textbf{93.6}$_{\pm0.5}$ & \textbf{94.5}$_{\pm0.5}$ & 94.1$_{\pm0.3}$ \\
VP-OTFM & 0.0$_{\pm0.0}$ & 0.0$_{\pm0.0}$ & 2.2$_{\pm0.5}$ & 71.1$_{\pm1.1}$ & 89.3$_{\pm0.3}$ & 93.8$_{\pm0.6}$ & \textbf{94.9}$_{\pm0.5}$ \\
DDPM-linear & 0.0$_{\pm0.0}$ & 0.0$_{\pm0.0}$ & 0.0$_{\pm0.0}$ & 0.6$_{\pm0.1}$ & 45.7$_{\pm1.1}$ & 86.0$_{\pm0.9}$ & 92.0$_{\pm0.3}$ \\
DDPM-cos [$\nu=2.5$] & 0.0$_{\pm0.0}$ & 23.3$_{\pm0.7}$ & 80.9$_{\pm1.4}$ & 92.1$_{\pm0.8}$ & \textbf{94.1}$_{\pm0.6}$ & \textbf{94.6}$_{\pm0.4}$ & \textbf{94.7}$_{\pm0.4}$ \\
DDPM-cos [$\nu=1.0$] & 0.0$_{\pm0.0}$ & 0.0$_{\pm0.0}$ & 0.4$_{\pm0.1}$ & 44.7$_{\pm0.6}$ & 85.8$_{\pm0.7}$ & 92.0$_{\pm0.4}$ & \textbf{94.5}$_{\pm0.4}$ \\
\textbf{VP-ISSNR (ours)} & \textbf{3.2}$_{\pm0.2}$ & \textbf{70.6}$_{\pm0.3}$ & \textbf{89.9}$_{\pm0.6}$ & \textbf{93.4}$_{\pm0.5}$ & \textbf{94.4}$_{\pm0.8}$ & \textbf{94.8}$_{\pm0.3}$ & \textbf{95.0}$_{\pm0.3}$ \\
\br
\end{tabular}
\end{table}

% ---- Validity ----
\begin{table}
\centering
\caption{\rebuttal{Validity (\%) on QM9 as a function of the NFE, for the reverse ODE and the reverse SDE with the Euler solver. Each entry is the mean over five independent sampling runs with different random seeds, with $\pm 1$ standard deviation over those seeds as a subscript. Within each column and ODE/SDE block, every value whose $\pm 1$ standard deviation interval overlaps that of the column's largest value is set in bold. In the low NFE regime, our method is always outperforming the baselines.}}
\label{tab:MetricsValidity}
\footnotesize
\setlength{\tabcolsep}{3pt}
\begin{tabular}{@{}lccccccc@{}}
\br
Schedule & \textbf{4} & \textbf{8} & \textbf{16} & \textbf{32} & \textbf{64} & \textbf{128} & \textbf{256} \\
\mr
\multicolumn{8}{l}{\emph{Reverse ODE}} \\
EDM & 0.0$_{\pm0.0}$ & 7.4$_{\pm0.5}$ & 46.7$_{\pm0.9}$ & 69.2$_{\pm1.3}$ & 76.8$_{\pm1.1}$ & 79.7$_{\pm1.0}$ & \textbf{81.0}$_{\pm0.7}$ \\
EDM-UT & 0.0$_{\pm0.0}$ & 0.0$_{\pm0.0}$ & 31.6$_{\pm0.5}$ & 72.0$_{\pm0.3}$ & 78.7$_{\pm0.7}$ & \textbf{80.9}$_{\pm0.8}$ & \textbf{81.8}$_{\pm0.7}$ \\
SMLD & 0.0$_{\pm0.0}$ & 0.0$_{\pm0.0}$ & 0.5$_{\pm0.1}$ & 64.3$_{\pm0.6}$ & 78.1$_{\pm0.6}$ & \textbf{80.7}$_{\pm0.5}$ & \textbf{81.8}$_{\pm0.5}$ \\
OTFM & 1.0$_{\pm0.2}$ & 20.1$_{\pm0.4}$ & 56.8$_{\pm0.7}$ & 73.2$_{\pm0.6}$ & 78.7$_{\pm0.5}$ & \textbf{80.5}$_{\pm0.7}$ & \textbf{81.3}$_{\pm0.6}$ \\
VP-EDM-UT & 0.0$_{\pm0.0}$ & 38.2$_{\pm0.7}$ & 65.8$_{\pm0.7}$ & 75.7$_{\pm0.6}$ & 79.2$_{\pm1.0}$ & \textbf{81.0}$_{\pm0.6}$ & \textbf{81.5}$_{\pm0.6}$ \\
VP-SMLD & 0.1$_{\pm0.0}$ & 47.2$_{\pm1.0}$ & 68.7$_{\pm0.6}$ & 76.4$_{\pm0.6}$ & \textbf{79.7}$_{\pm0.9}$ & \textbf{81.1}$_{\pm0.7}$ & \textbf{81.6}$_{\pm0.6}$ \\
VP-OTFM & 1.1$_{\pm0.3}$ & 26.2$_{\pm0.4}$ & 60.5$_{\pm0.5}$ & 74.1$_{\pm0.8}$ & 79.1$_{\pm0.4}$ & \textbf{80.9}$_{\pm0.4}$ & \textbf{81.4}$_{\pm0.4}$ \\
DDPM-linear & 0.0$_{\pm0.0}$ & 0.7$_{\pm0.1}$ & 15.9$_{\pm0.8}$ & 47.7$_{\pm0.8}$ & 67.5$_{\pm0.9}$ & 75.4$_{\pm0.7}$ & 79.2$_{\pm0.7}$ \\
DDPM-cos [$\nu=2.5$] & 1.3$_{\pm0.2}$ & 35.9$_{\pm0.8}$ & 63.6$_{\pm0.9}$ & 74.4$_{\pm1.3}$ & 79.3$_{\pm0.5}$ & \textbf{80.8}$_{\pm0.5}$ & \textbf{81.5}$_{\pm0.4}$ \\
DDPM-cos [$\nu=1.0$] & 0.0$_{\pm0.0}$ & 4.5$_{\pm0.4}$ & 35.1$_{\pm1.5}$ & 62.6$_{\pm1.2}$ & 74.1$_{\pm0.8}$ & 78.2$_{\pm0.4}$ & 80.2$_{\pm0.7}$ \\
\textbf{VP-ISSNR (ours)} & \textbf{30.2}$_{\pm0.5}$ & \textbf{66.9}$_{\pm0.5}$ & \textbf{76.4}$_{\pm0.7}$ & \textbf{79.4}$_{\pm0.3}$ & \textbf{81.1}$_{\pm0.7}$ & \textbf{81.7}$_{\pm0.7}$ & \textbf{82.1}$_{\pm0.8}$ \\
\mr
\multicolumn{8}{l}{\emph{Reverse SDE}} \\
VP-EDM-UT & 0.0$_{\pm0.0}$ & 0.2$_{\pm0.1}$ & 66.3$_{\pm1.0}$ & 84.9$_{\pm0.4}$ & 88.0$_{\pm0.5}$ & \textbf{89.6}$_{\pm0.2}$ & \textbf{89.4}$_{\pm1.0}$ \\
VP-SMLD & 0.0$_{\pm0.0}$ & 3.1$_{\pm0.2}$ & 73.4$_{\pm0.8}$ & 85.8$_{\pm0.6}$ & \textbf{88.6}$_{\pm0.9}$ & \textbf{89.4}$_{\pm0.4}$ & 88.9$_{\pm0.2}$ \\
VP-OTFM & 0.0$_{\pm0.0}$ & 0.0$_{\pm0.0}$ & 2.1$_{\pm0.2}$ & 67.4$_{\pm1.2}$ & 86.1$_{\pm0.8}$ & 88.8$_{\pm0.5}$ & \textbf{90.1}$_{\pm0.3}$ \\
DDPM-linear & 0.0$_{\pm0.0}$ & 0.0$_{\pm0.0}$ & 0.0$_{\pm0.0}$ & 0.5$_{\pm0.1}$ & 45.0$_{\pm0.9}$ & 82.4$_{\pm1.2}$ & 88.5$_{\pm0.5}$ \\
DDPM-cos [$\nu=2.5$] & 0.0$_{\pm0.0}$ & 22.7$_{\pm0.3}$ & 78.3$_{\pm1.2}$ & 87.1$_{\pm0.8}$ & \textbf{89.2}$_{\pm0.9}$ & \textbf{89.5}$_{\pm0.6}$ & \textbf{89.6}$_{\pm0.6}$ \\
DDPM-cos [$\nu=1.0$] & 0.0$_{\pm0.0}$ & 0.0$_{\pm0.0}$ & 0.4$_{\pm0.2}$ & 44.0$_{\pm0.8}$ & 82.7$_{\pm0.4}$ & 88.2$_{\pm0.5}$ & 89.4$_{\pm0.3}$ \\
\textbf{VP-ISSNR (ours)} & \textbf{1.2}$_{\pm0.2}$ & \textbf{66.1}$_{\pm1.0}$ & \textbf{84.6}$_{\pm0.8}$ & \textbf{88.7}$_{\pm0.5}$ & \textbf{89.6}$_{\pm0.5}$ & \textbf{89.9}$_{\pm0.5}$ & \textbf{90.0}$_{\pm0.5}$ \\
\br
\end{tabular}
\end{table}

% ---- Uniqueness ----
\begin{table}
\centering
\caption{\rebuttal{Uniqueness (\%) on QM9 as a function of the NFE, for the reverse ODE and the reverse SDE with the Euler solver. Each entry is the mean over five independent sampling runs with different random seeds, with $\pm 1$ standard deviation over those seeds as a subscript. Within each column and ODE/SDE block, every value whose $\pm 1$ standard deviation interval overlaps that of the column's largest value is set in bold. Entries with fewer than 50 stable structures per seed are excluded from the comparison.}}
\label{tab:MetricsUniqueness}
\footnotesize
\setlength{\tabcolsep}{3pt}
\begin{tabular}{@{}lccccccc@{}}
\br
Schedule & \textbf{4} & \textbf{8} & \textbf{16} & \textbf{32} & \textbf{64} & \textbf{128} & \textbf{256} \\
\mr
\multicolumn{8}{l}{\emph{Reverse ODE}} \\
EDM & --- & \textbf{99.9}$_{\pm0.2}$ & \textbf{99.6}$_{\pm0.3}$ & \textbf{99.3}$_{\pm0.1}$ & \textbf{99.4}$_{\pm0.1}$ & \textbf{99.3}$_{\pm0.1}$ & \textbf{99.3}$_{\pm0.0}$ \\
EDM-UT & --- & --- & 95.1$_{\pm0.9}$ & 97.5$_{\pm0.3}$ & 98.8$_{\pm0.2}$ & \textbf{99.2}$_{\pm0.1}$ & \textbf{99.3}$_{\pm0.1}$ \\
SMLD & --- & --- & --- & 93.9$_{\pm0.6}$ & 98.3$_{\pm0.3}$ & \textbf{99.1}$_{\pm0.3}$ & \textbf{99.2}$_{\pm0.1}$ \\
OTFM & \textbf{99.8}$_{\pm0.5}$ & \textbf{99.4}$_{\pm0.3}$ & \textbf{99.4}$_{\pm0.2}$ & 99.2$_{\pm0.2}$ & \textbf{99.2}$_{\pm0.1}$ & \textbf{99.3}$_{\pm0.1}$ & \textbf{99.2}$_{\pm0.1}$ \\
VP-EDM-UT & --- & \textbf{99.8}$_{\pm0.2}$ & \textbf{99.5}$_{\pm0.1}$ & \textbf{99.3}$_{\pm0.1}$ & \textbf{99.3}$_{\pm0.1}$ & \textbf{99.2}$_{\pm0.1}$ & \textbf{99.3}$_{\pm0.1}$ \\
VP-SMLD & --- & \textbf{99.6}$_{\pm0.1}$ & \textbf{99.5}$_{\pm0.1}$ & 99.3$_{\pm0.1}$ & \textbf{99.3}$_{\pm0.0}$ & \textbf{99.3}$_{\pm0.0}$ & \textbf{99.3}$_{\pm0.1}$ \\
VP-OTFM & --- & \textbf{99.6}$_{\pm0.1}$ & \textbf{99.4}$_{\pm0.1}$ & \textbf{99.2}$_{\pm0.2}$ & \textbf{99.2}$_{\pm0.2}$ & \textbf{99.3}$_{\pm0.2}$ & \textbf{99.2}$_{\pm0.1}$ \\
DDPM-linear & --- & --- & \textbf{99.6}$_{\pm0.3}$ & \textbf{99.6}$_{\pm0.1}$ & \textbf{99.4}$_{\pm0.2}$ & \textbf{99.3}$_{\pm0.2}$ & \textbf{99.2}$_{\pm0.1}$ \\
DDPM-cos [$\nu=2.5$] & --- & 98.3$_{\pm0.4}$ & 98.7$_{\pm0.4}$ & 98.9$_{\pm0.2}$ & \textbf{99.2}$_{\pm0.1}$ & \textbf{99.2}$_{\pm0.3}$ & \textbf{99.2}$_{\pm0.1}$ \\
DDPM-cos [$\nu=1.0$] & --- & \textbf{99.4}$_{\pm0.8}$ & \textbf{99.1}$_{\pm0.3}$ & 99.2$_{\pm0.3}$ & \textbf{99.2}$_{\pm0.2}$ & \textbf{99.3}$_{\pm0.1}$ & \textbf{99.2}$_{\pm0.2}$ \\
\textbf{VP-ISSNR (ours)} & \textbf{99.6}$_{\pm0.2}$ & 99.4$_{\pm0.2}$ & \textbf{99.4}$_{\pm0.1}$ & 99.3$_{\pm0.1}$ & \textbf{99.3}$_{\pm0.1}$ & \textbf{99.3}$_{\pm0.1}$ & \textbf{99.3}$_{\pm0.1}$ \\
\mr
\multicolumn{8}{l}{\emph{Reverse SDE}} \\
VP-EDM-UT & --- & --- & 99.2$_{\pm0.1}$ & \textbf{99.1}$_{\pm0.3}$ & \textbf{99.1}$_{\pm0.2}$ & \textbf{98.9}$_{\pm0.2}$ & \textbf{99.1}$_{\pm0.2}$ \\
VP-SMLD & --- & \textbf{100.0}$_{\pm0.0}$ & 99.2$_{\pm0.1}$ & 99.2$_{\pm0.1}$ & 98.9$_{\pm0.1}$ & \textbf{99.1}$_{\pm0.2}$ & \textbf{99.1}$_{\pm0.1}$ \\
VP-OTFM & --- & --- & \textbf{100.0}$_{\pm0.0}$ & \textbf{99.3}$_{\pm0.2}$ & \textbf{99.1}$_{\pm0.2}$ & \textbf{99.0}$_{\pm0.1}$ & \textbf{99.2}$_{\pm0.2}$ \\
DDPM-linear & --- & --- & --- & --- & \textbf{99.4}$_{\pm0.2}$ & \textbf{99.2}$_{\pm0.3}$ & \textbf{99.1}$_{\pm0.1}$ \\
DDPM-cos [$\nu=2.5$] & --- & 99.3$_{\pm0.4}$ & 99.0$_{\pm0.1}$ & 99.1$_{\pm0.2}$ & 99.1$_{\pm0.1}$ & \textbf{99.1}$_{\pm0.2}$ & \textbf{99.0}$_{\pm0.1}$ \\
DDPM-cos [$\nu=1.0$] & --- & --- & --- & \textbf{99.5}$_{\pm0.2}$ & \textbf{99.0}$_{\pm0.3}$ & \textbf{99.2}$_{\pm0.2}$ & \textbf{99.0}$_{\pm0.2}$ \\
\textbf{VP-ISSNR (ours)} & \textbf{100.0}$_{\pm0.0}$ & 99.4$_{\pm0.1}$ & 99.2$_{\pm0.2}$ & \textbf{99.2}$_{\pm0.2}$ & 99.0$_{\pm0.1}$ & \textbf{98.9}$_{\pm0.1}$ & \textbf{99.0}$_{\pm0.1}$ \\
\br
\end{tabular}
\end{table}

% ---- Novelty ----
\begin{table}
\centering
\caption{\rebuttal{Novelty (\%) on QM9 as a function of the NFE, for the reverse ODE and the reverse SDE with the Euler solver. Each entry is the mean over five independent sampling runs with different random seeds, with $\pm 1$ standard deviation over those seeds as a subscript. Within each column and ODE/SDE block, every value whose $\pm 1$ standard deviation interval overlaps that of the column's largest value is set in bold. Entries with fewer than 50 stable structures per seed are excluded from the comparison.}}
\label{tab:MetricsNovelty}
\footnotesize
\setlength{\tabcolsep}{3pt}
\begin{tabular}{@{}lccccccc@{}}
\br
Schedule & \textbf{4} & \textbf{8} & \textbf{16} & \textbf{32} & \textbf{64} & \textbf{128} & \textbf{256} \\
\mr
\multicolumn{8}{l}{\emph{Reverse ODE}} \\
EDM & --- & \textbf{59.1}$_{\pm3.4}$ & 53.9$_{\pm0.7}$ & 52.7$_{\pm0.7}$ & \textbf{52.0}$_{\pm1.0}$ & \textbf{51.4}$_{\pm0.7}$ & \textbf{51.3}$_{\pm0.9}$ \\
EDM-UT & --- & --- & \textbf{58.2}$_{\pm1.6}$ & \textbf{53.4}$_{\pm1.4}$ & \textbf{52.7}$_{\pm0.4}$ & \textbf{52.7}$_{\pm1.4}$ & \textbf{51.7}$_{\pm1.2}$ \\
SMLD & --- & --- & --- & \textbf{55.9}$_{\pm1.2}$ & \textbf{53.0}$_{\pm0.9}$ & \textbf{52.5}$_{\pm0.2}$ & \textbf{52.5}$_{\pm1.6}$ \\
OTFM & \textbf{63.6}$_{\pm4.8}$ & 52.0$_{\pm2.1}$ & 51.6$_{\pm0.8}$ & 51.4$_{\pm1.2}$ & \textbf{51.2}$_{\pm1.2}$ & \textbf{51.2}$_{\pm0.5}$ & \textbf{51.1}$_{\pm0.2}$ \\
VP-EDM-UT & --- & \textbf{56.1}$_{\pm1.3}$ & 54.1$_{\pm0.7}$ & 52.8$_{\pm0.3}$ & \textbf{52.3}$_{\pm0.5}$ & \textbf{51.8}$_{\pm0.8}$ & \textbf{51.3}$_{\pm0.8}$ \\
VP-SMLD & --- & \textbf{55.9}$_{\pm0.9}$ & 54.1$_{\pm0.6}$ & 53.0$_{\pm0.5}$ & \textbf{52.3}$_{\pm0.3}$ & \textbf{51.9}$_{\pm0.9}$ & \textbf{51.4}$_{\pm0.8}$ \\
VP-OTFM & --- & 50.1$_{\pm1.6}$ & 50.7$_{\pm0.8}$ & 51.3$_{\pm1.4}$ & 51.0$_{\pm0.6}$ & \textbf{51.0}$_{\pm0.6}$ & \textbf{51.0}$_{\pm0.5}$ \\
DDPM-linear & --- & --- & 52.3$_{\pm3.8}$ & 50.5$_{\pm0.7}$ & 50.4$_{\pm0.5}$ & \textbf{51.1}$_{\pm0.4}$ & \textbf{51.2}$_{\pm0.6}$ \\
DDPM-cos [$\nu=2.5$] & --- & 50.3$_{\pm0.8}$ & 50.3$_{\pm0.6}$ & 50.0$_{\pm1.3}$ & 50.4$_{\pm0.6}$ & \textbf{50.6}$_{\pm1.0}$ & \textbf{50.8}$_{\pm1.0}$ \\
DDPM-cos [$\nu=1.0$] & --- & 50.5$_{\pm1.3}$ & 46.9$_{\pm1.3}$ & 48.0$_{\pm1.0}$ & 49.3$_{\pm1.1}$ & 50.3$_{\pm0.7}$ & \textbf{51.1}$_{\pm0.6}$ \\
\textbf{VP-ISSNR (ours)} & 57.1$_{\pm1.3}$ & \textbf{55.2}$_{\pm1.7}$ & 53.7$_{\pm0.9}$ & 52.8$_{\pm0.8}$ & \textbf{51.9}$_{\pm0.3}$ & \textbf{51.6}$_{\pm0.8}$ & \textbf{51.2}$_{\pm0.9}$ \\
\mr
\multicolumn{8}{l}{\emph{Reverse SDE}} \\
VP-EDM-UT & --- & --- & 48.4$_{\pm0.9}$ & \textbf{49.1}$_{\pm1.0}$ & \textbf{47.9}$_{\pm1.4}$ & \textbf{47.8}$_{\pm0.6}$ & \textbf{46.8}$_{\pm0.5}$ \\
VP-SMLD & --- & \textbf{66.5}$_{\pm2.9}$ & 50.6$_{\pm0.9}$ & \textbf{50.0}$_{\pm1.6}$ & \textbf{47.4}$_{\pm0.9}$ & \textbf{47.4}$_{\pm1.1}$ & \textbf{46.9}$_{\pm0.9}$ \\
VP-OTFM & --- & --- & \textbf{59.5}$_{\pm5.8}$ & \textbf{47.9}$_{\pm0.6}$ & \textbf{47.2}$_{\pm0.2}$ & \textbf{47.5}$_{\pm0.5}$ & \textbf{47.1}$_{\pm1.1}$ \\
DDPM-linear & --- & --- & --- & --- & \textbf{47.8}$_{\pm1.5}$ & \textbf{47.7}$_{\pm1.0}$ & \textbf{47.8}$_{\pm0.9}$ \\
DDPM-cos [$\nu=2.5$] & --- & 49.7$_{\pm1.5}$ & 47.5$_{\pm1.1}$ & \textbf{47.6}$_{\pm1.2}$ & \textbf{46.6}$_{\pm0.6}$ & \textbf{46.1}$_{\pm0.8}$ & \textbf{46.7}$_{\pm1.1}$ \\
DDPM-cos [$\nu=1.0$] & --- & --- & --- & \textbf{47.2}$_{\pm1.6}$ & \textbf{47.7}$_{\pm0.9}$ & \textbf{48.1}$_{\pm1.3}$ & \textbf{47.1}$_{\pm1.3}$ \\
\textbf{VP-ISSNR (ours)} & \textbf{80.2}$_{\pm4.5}$ & 53.9$_{\pm2.2}$ & 50.4$_{\pm1.2}$ & \textbf{48.0}$_{\pm0.8}$ & \textbf{47.5}$_{\pm1.2}$ & \textbf{46.1}$_{\pm1.1}$ & \textbf{46.5}$_{\pm1.1}$ \\
\br
\end{tabular}
\end{table}

% ---- Stability rate, binomial confidence intervals ----
\begin{table}
\centering
\caption{\rebuttal{Binomial $95\,\%$ confidence intervals (Wilson score) of the stability rate (\%) in table~\ref{tab:MetricsStability}.}}
\label{tab:StabilityBinomialCI}
\scriptsize
\setlength{\tabcolsep}{2.5pt}
\begin{tabular}{@{}lccccccc@{}}
\br
Schedule & \textbf{4} & \textbf{8} & \textbf{16} & \textbf{32} & \textbf{64} & \textbf{128} & \textbf{256} \\
\mr
\multicolumn{8}{l}{\emph{Reverse ODE}} \\
EDM & [0.2, 0.4] & [15.2, 16.4] & [54.7, 56.4] & [75.5, 76.9] & [81.9, 83.2] & [84.6, 85.8] & [85.6, 86.8] \\
EDM-UT & [0.0, 0.0] & [0.0, 0.0] & [42.2, 43.9] & [80.6, 81.9] & [85.5, 86.7] & [87.0, 88.1] & [87.2, 88.3] \\
SMLD & [0.0, 0.0] & [0.0, 0.2] & [0.7, 1.1] & [75.0, 76.5] & [86.2, 87.3] & [87.4, 88.5] & [87.8, 88.9] \\
OTFM & [2.3, 2.8] & [27.5, 29.1] & [64.4, 66.1] & [79.2, 80.6] & [84.2, 85.5] & [85.7, 86.9] & [86.3, 87.4] \\
VP-EDM-UT & [0.0, 0.1] & [42.1, 43.8] & [71.6, 73.1] & [80.4, 81.8] & [84.0, 85.2] & [85.5, 86.7] & [86.1, 87.2] \\
VP-SMLD & [0.1, 0.3] & [52.5, 54.2] & [74.2, 75.6] & [81.3, 82.6] & [84.3, 85.6] & [85.6, 86.8] & [86.2, 87.4] \\
VP-OTFM & [1.1, 1.5] & [29.2, 30.8] & [66.1, 67.7] & [79.7, 81.1] & [84.1, 85.3] & [85.8, 87.0] & [86.2, 87.4] \\
DDPM-linear & [0.0, 0.0] & [0.7, 1.1] & [17.5, 18.8] & [52.8, 54.5] & [73.3, 74.8] & [81.1, 82.4] & [84.5, 85.7] \\
DDPM-cos [$\nu=2.5$] & [1.4, 1.8] & [37.0, 38.7] & [66.9, 68.5] & [79.0, 80.4] & [84.0, 85.2] & [85.6, 86.8] & [86.3, 87.5] \\
DDPM-cos [$\nu=1.0$] & [0.0, 0.0] & [4.0, 4.7] & [37.6, 39.2] & [67.5, 69.1] & [79.6, 81.0] & [83.6, 84.8] & [85.5, 86.7] \\
\textbf{VP-ISSNR (ours)} & [35.3, 37.0] & [72.7, 74.2] & [81.8, 83.1] & [84.4, 85.6] & [85.7, 86.9] & [86.3, 87.5] & [86.6, 87.7] \\
\mr
\multicolumn{8}{l}{\emph{Reverse SDE}} \\
VP-EDM-UT & [0.0, 0.0] & [0.7, 1.0] & [70.5, 72.1] & [89.9, 90.9] & [92.9, 93.8] & [94.4, 95.2] & [94.2, 95.0] \\
VP-SMLD & [0.0, 0.0] & [5.2, 6.0] & [78.4, 79.8] & [90.4, 91.4] & [93.1, 94.0] & [94.1, 94.9] & [93.7, 94.5] \\
VP-OTFM & [0.0, 0.0] & [0.0, 0.0] & [2.0, 2.5] & [70.3, 71.9] & [88.7, 89.8] & [93.3, 94.2] & [94.5, 95.3] \\
DDPM-linear & [0.0, 0.0] & [0.0, 0.0] & [0.0, 0.0] & [0.5, 0.7] & [44.8, 46.6] & [85.4, 86.6] & [91.5, 92.5] \\
DDPM-cos [$\nu=2.5$] & [0.0, 0.1] & [22.5, 24.0] & [80.2, 81.6] & [91.6, 92.5] & [93.7, 94.5] & [94.2, 95.0] & [94.3, 95.1] \\
DDPM-cos [$\nu=1.0$] & [0.0, 0.0] & [0.0, 0.0] & [0.3, 0.5] & [43.8, 45.5] & [85.2, 86.4] & [91.5, 92.4] & [94.1, 94.9] \\
\textbf{VP-ISSNR (ours)} & [3.0, 3.6] & [69.8, 71.4] & [89.4, 90.4] & [92.9, 93.8] & [94.0, 94.8] & [94.4, 95.2] & [94.6, 95.4] \\
\br
\end{tabular}
\end{table}

\begin{figure}[t]
    \centering
    \begin{minipage}{0.48\textwidth}
        \centering
        \includegraphics[width=\linewidth]{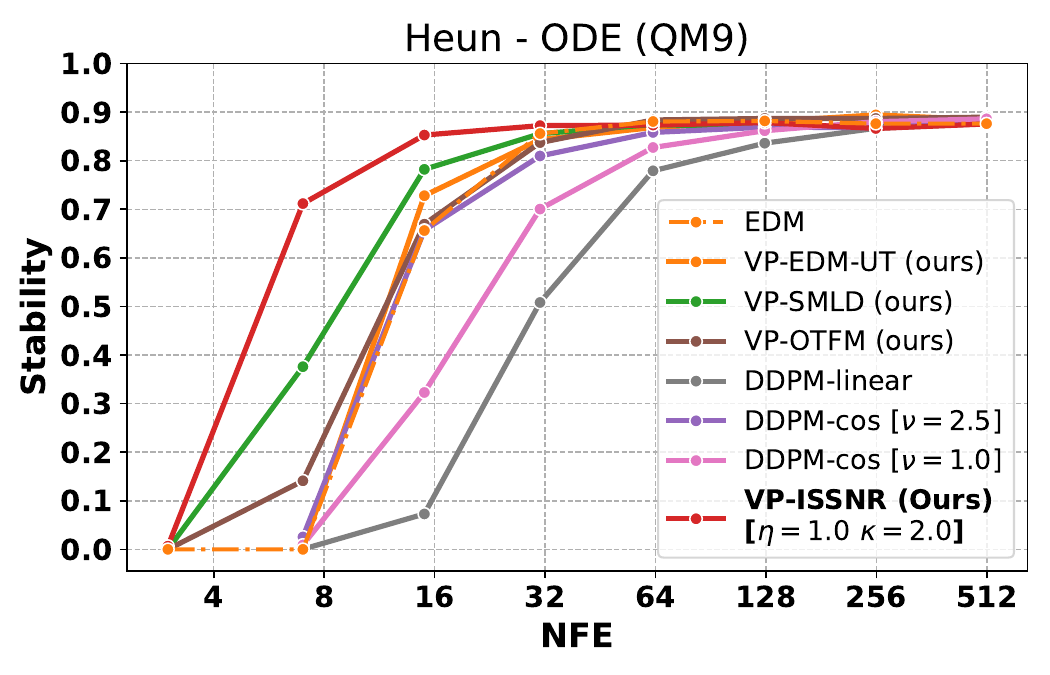}
    \end{minipage}
    \hfill
    \begin{minipage}{0.48\textwidth}
        \centering
        \includegraphics[width=\textwidth]{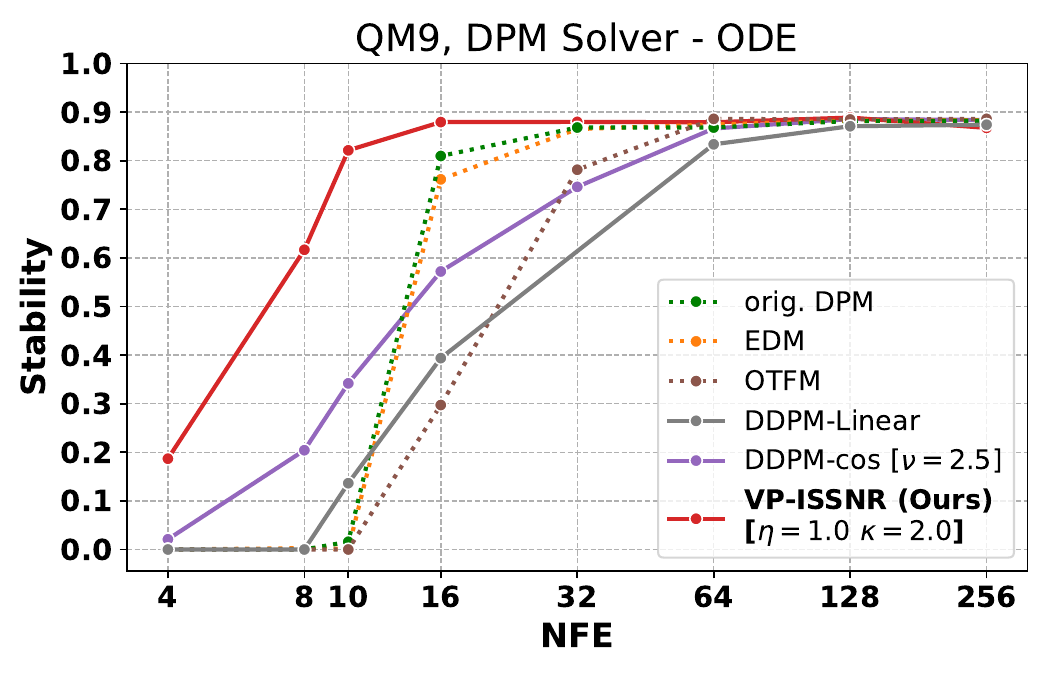}
    \end{minipage}
    \caption{Stability rate (\textbf{higher is better}) as a function of the number of function evaluations (NFE) for molecular structure generation on the QM9 dataset. \textbf{Left:} Second-order Heun sampler. \textbf{Right:} DPM solver \cite{dpm-solver, lu_dpm-solver_2023}.}
    \label{fig:AllComparisonMolecule_Heun}
\end{figure}

\begin{figure}[t]
    \centering
    \includegraphics[width=.49\linewidth]{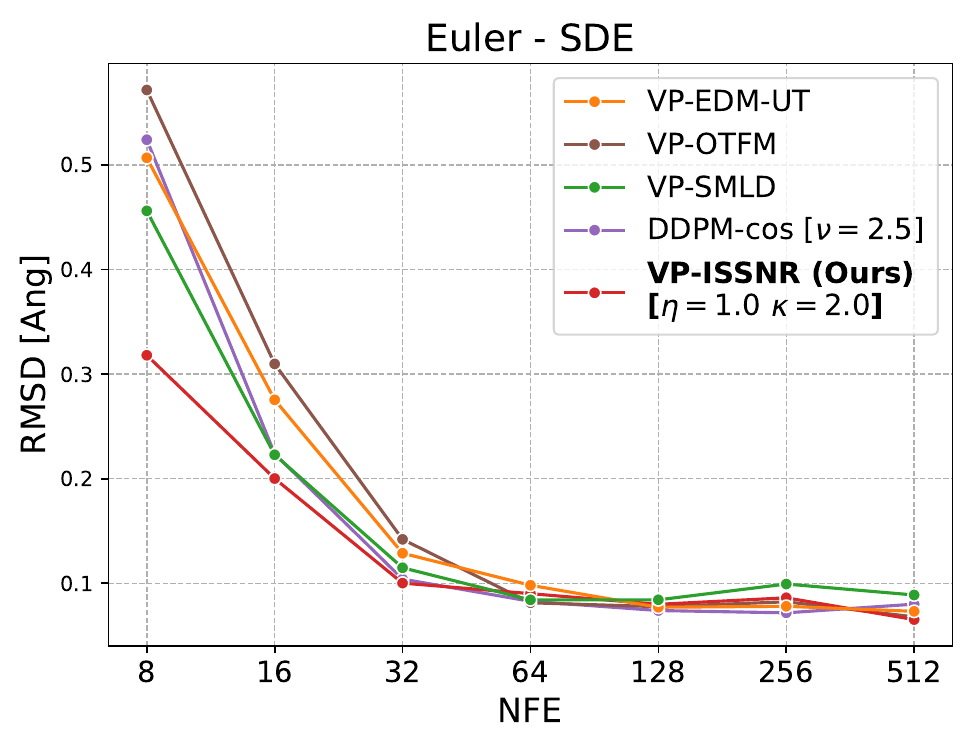}
    \caption{Root mean square deviation (RMSD, \textbf{lower is better}) between the generated structures and reference structures obtained from geometry relaxations using DFT calculations at the B3LYP/6-31G(2df,p) level of theory (see \ref{sec:DFTcalculations} for details), the same method used for generating the structures in QM9 \cite{qm9}, which the model was trained on. We see a similar trend to the stability rate results in figure~\ref{fig:ComparisonMolecule}, where our VP-ISSNR consistently outperforms other approaches at low NFE. This reveals that our method can generate physically plausible molecules that are structurally similar to ground truth reference structures.
    }
    \label{fig:AllComparisonMolecule_Heun_RMSD}
\end{figure}

\begin{figure}[t]
    \centering
    \includegraphics[width=.49\linewidth]{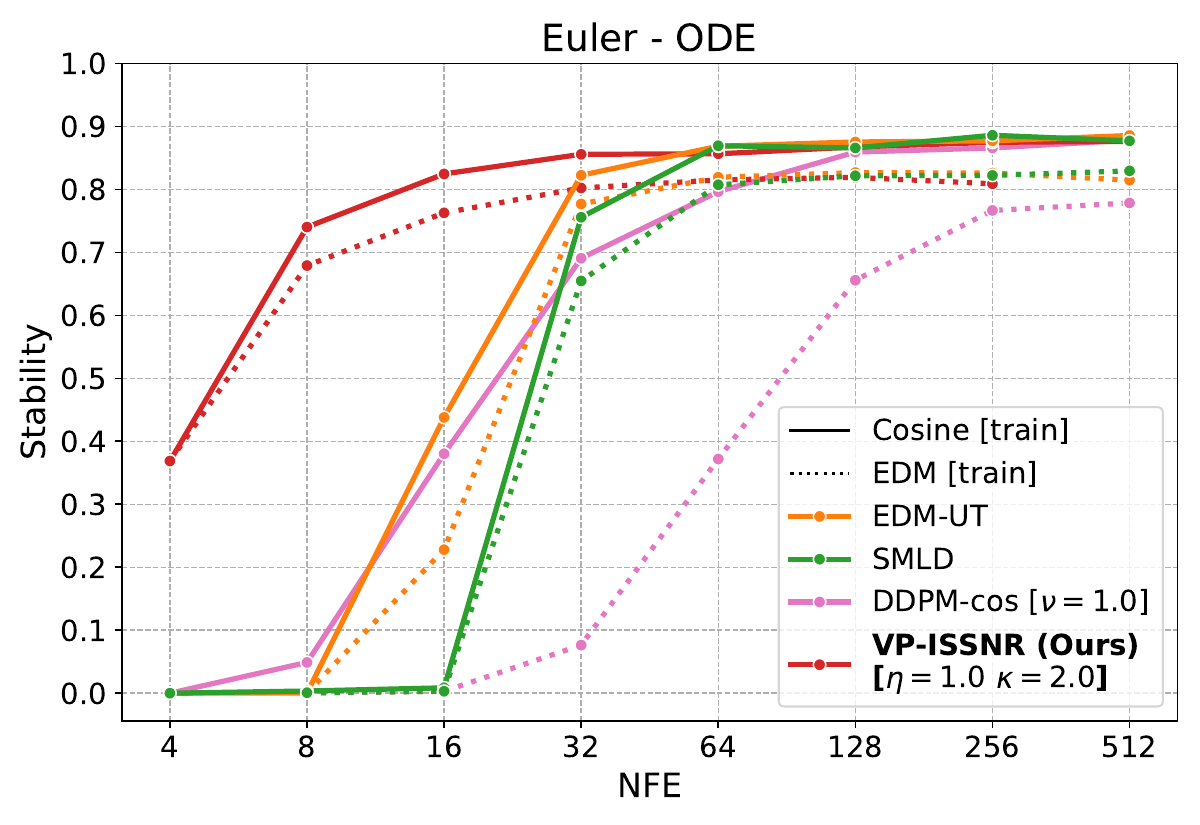}
    \caption{
    Effect of the schedule used during training. Using the same schedule for both training and sampling does not enhance results for EDM, whereas the Cosine schedule emphasizes sampling in more challenging regions during training. Our VP-ISSNR sampling still outperforms other baselines, even when the model is trained using the EDM schedule.}
    \label{fig:AllComparisonMolecule_EulerODE_AnotherTraining}
\end{figure}

Figure~\ref{fig:AllComparisonMolecule_Heun} shows the stability rate of the generated molecular structures with the second-order integration method, Heun, and the advanced DPM solver \cite{dpm-solver, lu_dpm-solver_2023}.
We observe only a marginal improvement with high NFEs, compared to the performance achieved by Euler in figure~\ref{fig:ComparisonMolecule}. However, the same trend between schedules is observed for different samplers, supporting the hypothesis that improvements in sampling methods and noise schedules are orthogonal to each other, i.e., they can be combined for even greater overall improvements.

Figure~\ref{fig:AllComparisonMolecule_Heun_RMSD} shows molecular structure generation performance evaluated by running DFT to relax the generated structures (see \ref{sec:DFTcalculations} for details),
which further validates the stability rate results (we observe a similar trend with this evaluation criterion).

\begin{figure}[t]
    \centering
    \includegraphics[width=.6\linewidth]{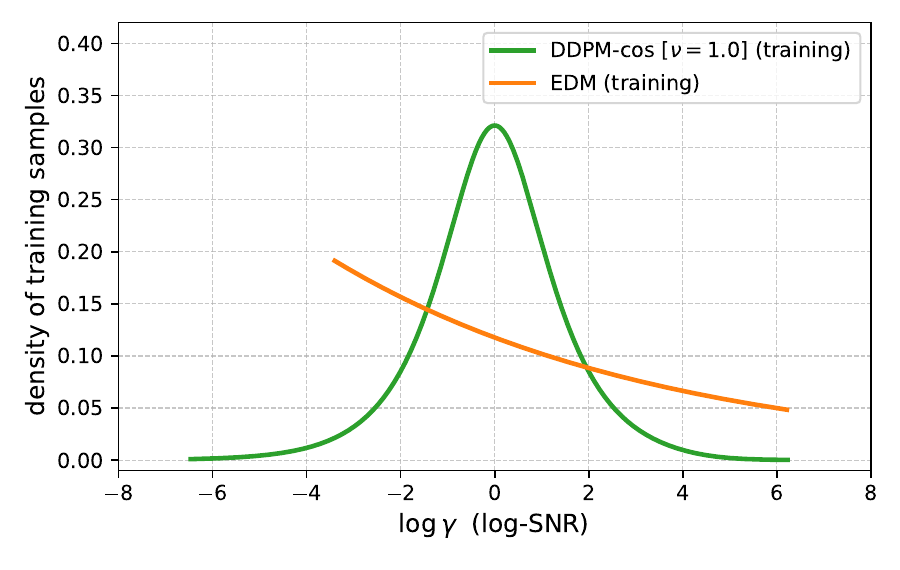}
    \caption{\rebuttal{Density of log-SNR values visited during training under the two training schedules considered in figure~\ref{fig:AllComparisonMolecule_EulerODE_AnotherTraining}. The cosine schedule concentrates the training budget in the intermediate log-SNR region, while the EDM schedule distributes it over a considerably wider range. Since training draws $t$ uniformly on $[0,1]$, the density is $\lvert \mathrm{d}\log\gamma/\mathrm{d}t \rvert^{-1}$, normalized to unit area.}}
    \label{fig:SNRCoverage}
\end{figure}

Figure~\ref{fig:AllComparisonMolecule_EulerODE_AnotherTraining} compares the sample generation performance with the diffusion model trained on different schedules, DDPM-cosine with $\nu=1$ and EDM. We can see that the model trained on the cosine schedule achieves consistently better results than the model trained on the EDM schedule, even when using the EDM schedule during sampling. \rebuttal{A plausible explanation is that the two training schedules allocate the training budget very differently over the SNR axis. Training draws $t$ uniformly on $[0,1]$, so the density of log-SNR values actually visited during training is the push-forward of that uniform density through $\log \gamma(t)$. Figure~\ref{fig:SNRCoverage} shows this density for the two training schedules. The cosine schedule concentrates its samples in the intermediate log-SNR band in which the signal changes most rapidly and the denoising problem is hardest, placing $83.6\,\%$ of its training samples in the range $\lvert\log\gamma\rvert < 2$, whereas the EDM schedule places only $47.7\,\%$ there and spreads the remainder over a much wider range, spending a substantial fraction of it at very low SNR where the optimal denoiser output is close to the data mean and little is learned. A score model that is more accurate in the critical band is then better regardless of which schedule is used at inference time, which is consistent with the ordering observed in figure~\ref{fig:AllComparisonMolecule_EulerODE_AnotherTraining}.}

Tables~\ref{tab:MetricsStability}--\ref{tab:StabilityBinomialCI} provide additional quantitative results for the stability rate, validity, uniqueness, novelty, and binomial confidence intervals of the stability rate, respectively.

\subsection{Image generation}
\label{sec:AdditionalExperimentalResultsImages}

\begin{figure*}[t]
    \centering
    \includegraphics[width=.49\linewidth]{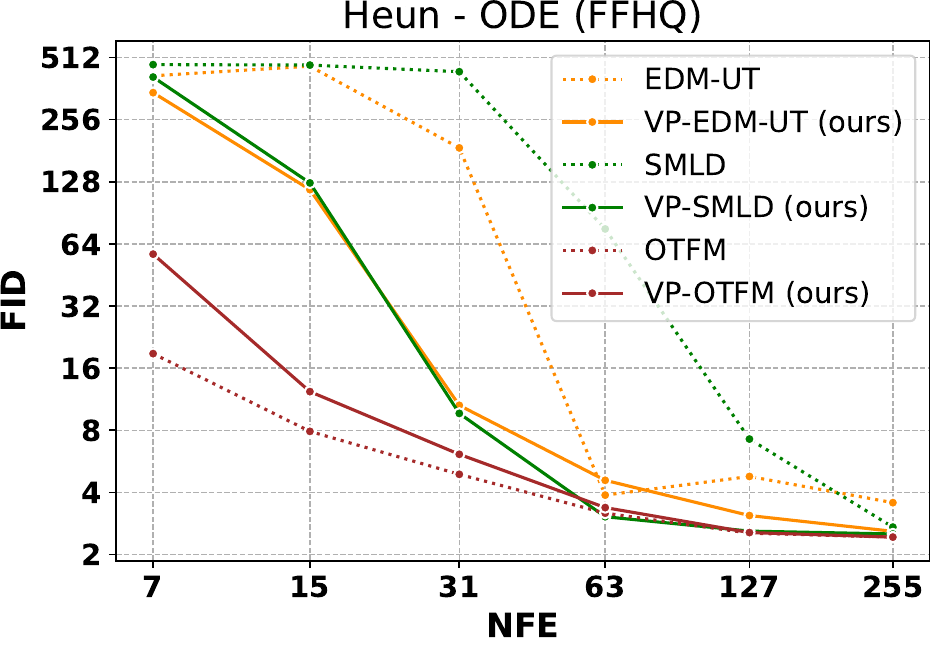}
    \caption{
    FID score \textbf{(lower is better)} as a function of the number of function evaluations (NFE) in image generation on FFHQ, comparing existing non-VP schedules with their VP variants.
    }
    \label{fig:ComparisonFFHQImage.VPComparison}
\end{figure*}

\begin{figure}[t]
    \centering
    \begin{minipage}{0.48\textwidth}
        \centering
        \includegraphics[width=\textwidth]{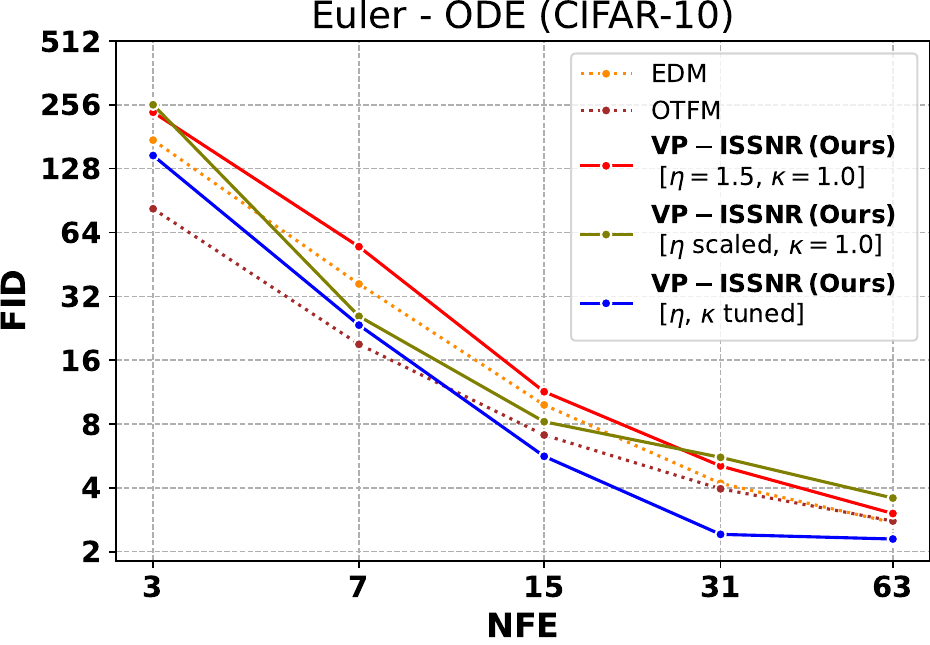}
    \end{minipage}
    \hfill
    \begin{minipage}{0.48\textwidth}
        \centering
        \includegraphics[width=\textwidth]{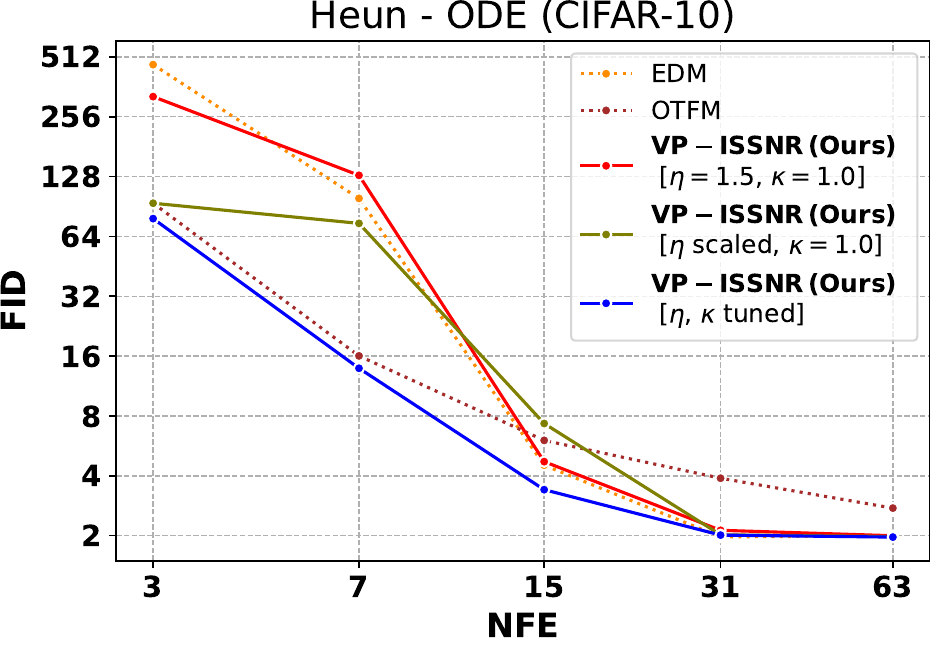}
    \end{minipage}
    \vspace{5pt}
    \begin{minipage}{0.48\textwidth}
        \centering
        \includegraphics[width=\textwidth]{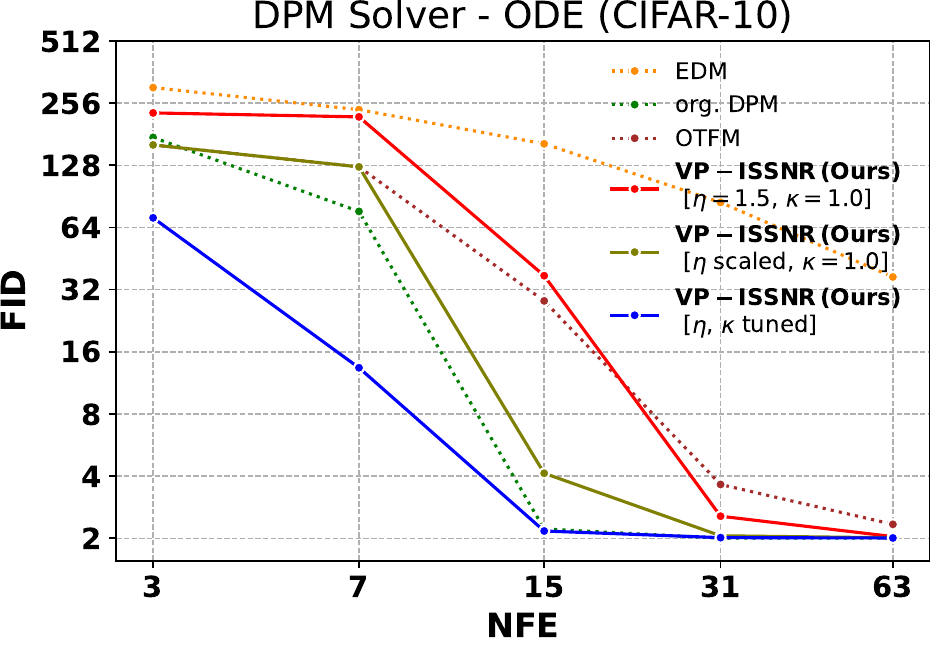}
    \end{minipage}
    \hfill
    \begin{minipage}{0.48\textwidth}
        \centering
        \includegraphics[width=\textwidth]{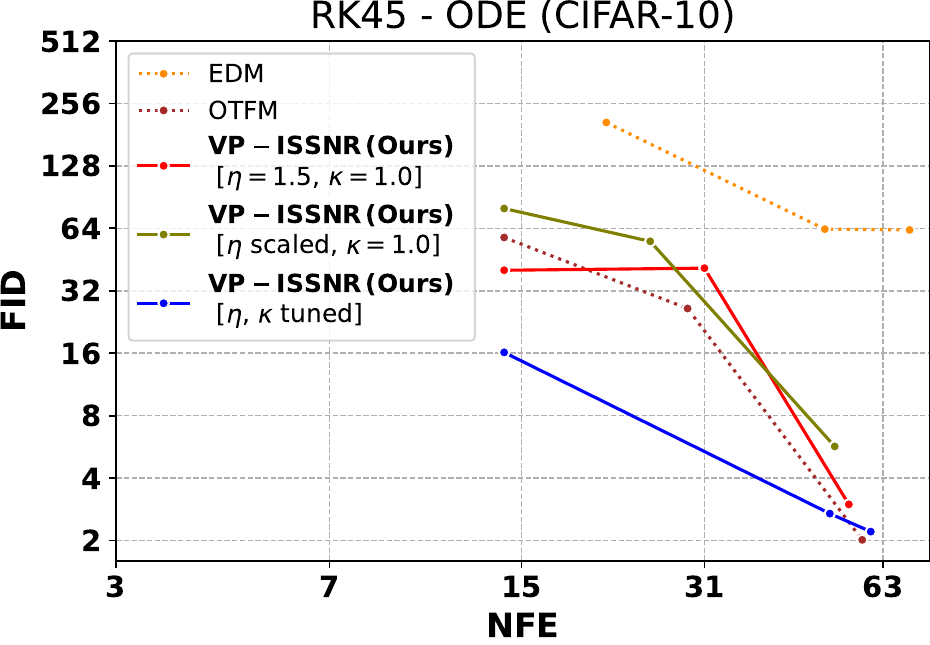}
    \end{minipage}
    \caption{FID score \textbf{(lower is better)} as a function of the number of function evaluations (NFE) for image generation on CIFAR-10. Results are shown for combinations of solvers and noise schedules, including the VP-ISSNR variant with fixed hyperparameters in red. Solvers include Euler (first-order), Heun (second-order), RK45 (high-order), and the advanced DPM solver \cite{dpm-solver, lu_dpm-solver_2023}. For the DPM solver, their proposed schedule is indicated in green. Due to the difficulty of precisely controlling the NFE for RK45, some data points may be missing or slightly shifted in the plot.}
    \label{fig:SolversComparison}
\end{figure}

\begin{figure}[t]
    \centering
    \begin{minipage}{0.48\textwidth}
        \centering
        \includegraphics[width=\textwidth]{figures/CIFAR_heun_fid_vs_nfe4.pdf}
    \end{minipage}
    \hfill
    \begin{minipage}{0.48\textwidth}
        \centering
        \includegraphics[width=\textwidth]{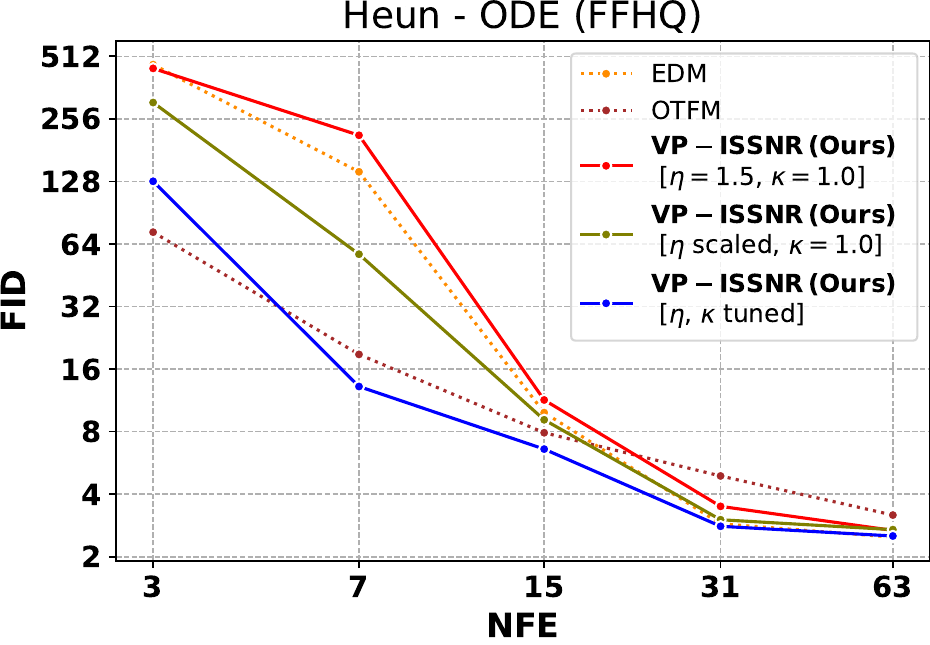}
    \end{minipage}
    \vspace{5pt}
    \begin{minipage}{0.48\textwidth}
        \centering
        \includegraphics[width=\textwidth]{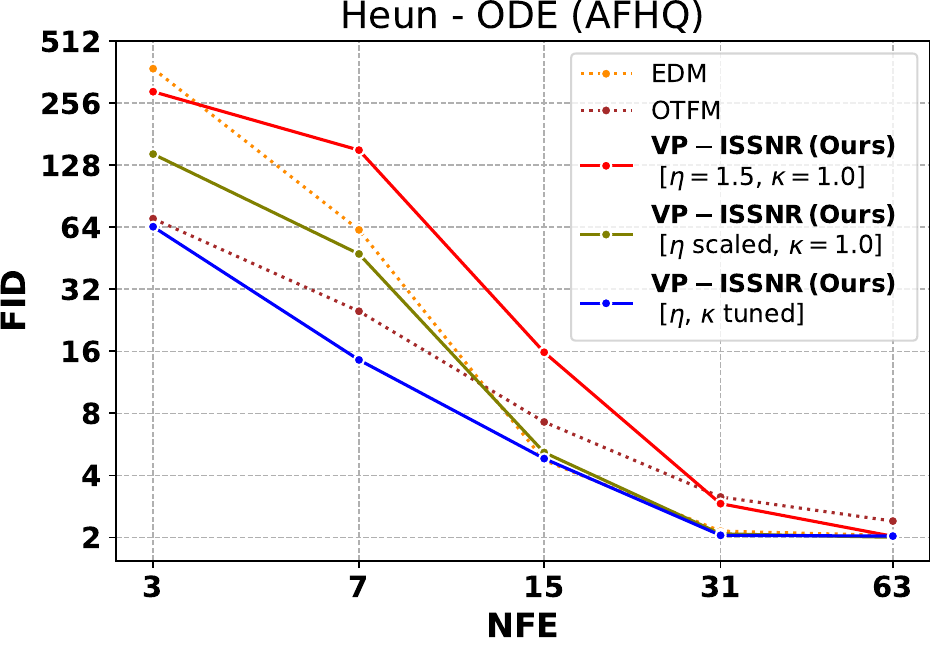}
    \end{minipage}
    \hfill
    \begin{minipage}{0.48\textwidth}
        \centering
        \includegraphics[width=\textwidth]{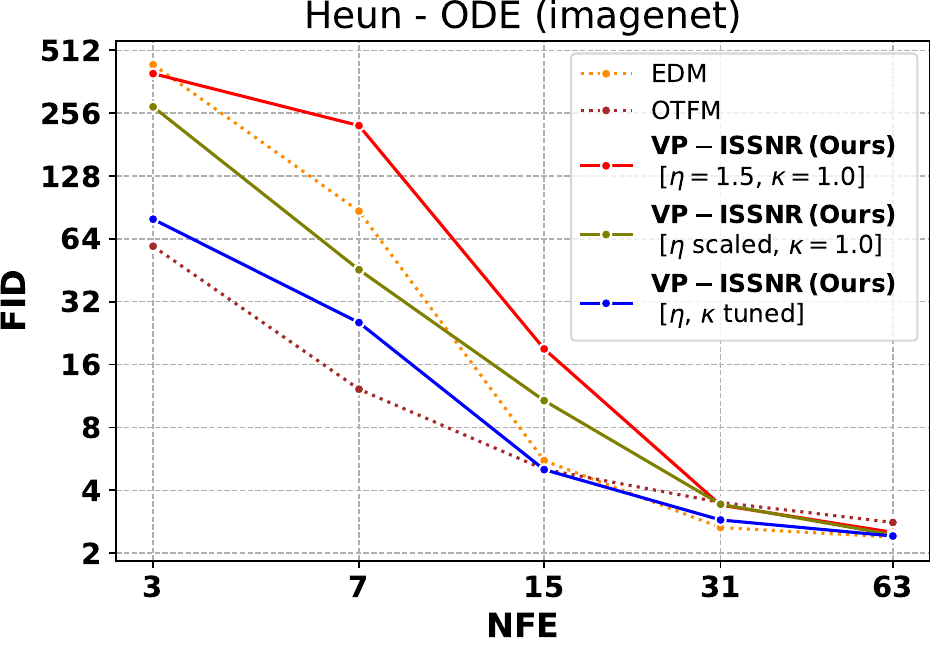}
    \end{minipage}
    \caption{FID score \textbf{(lower is better)} as a function of the number of function evaluations (NFE) for image generation using different datasets with different image resolutions and the second-order solver Heun. For completeness, we add the results of using the fixed hyperparameters variant of our VP-ISSNR schedule in red.}
    \label{fig:ImagesDatasetsComparison}
\end{figure}

For a more rigorous benchmarking of our approach, we further present the following experimental results:

Figure~\ref{fig:ComparisonFFHQImage.VPComparison} compares
existing non-VP with their VP variants
in image generation on FFHQ.

Figure~\ref{fig:SolversComparison} compares the results of using different solvers with different noise schedules, including the VP-ISSNR variant with fixed hyperparameters for completeness. The results further support the hypothesis that improvements in the sampler/solver are orthogonal to improvements in the noise schedule and can be combined for even greater overall improvement.

Figure~\ref{fig:ImagesDatasetsComparison} investigates the robustness and generalizability of different schedules, including the VP-ISSNR variant with fixed hyperparameters, to different image datasets with different resolutions. Overall, we can observe the same trade-off across datasets, with our tuned schedule reaching comparable or lower FID scores. For ImageNet at 3/7 NFE, OTFM outperforms.

\subsection{Trajectory analysis}
\label{sec:TorajectoryAnalysis}
\begin{figure*}[t]
    \centering
    \includegraphics[width=\linewidth]{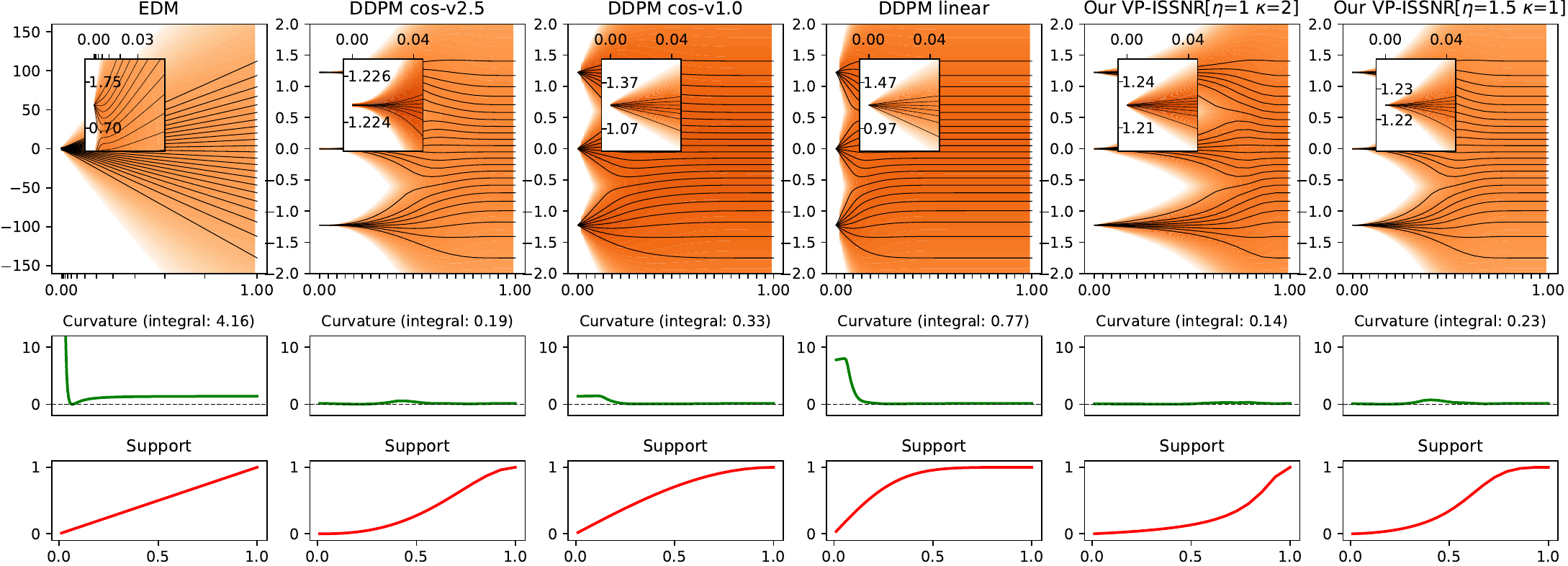}
    \includegraphics[width=\linewidth]{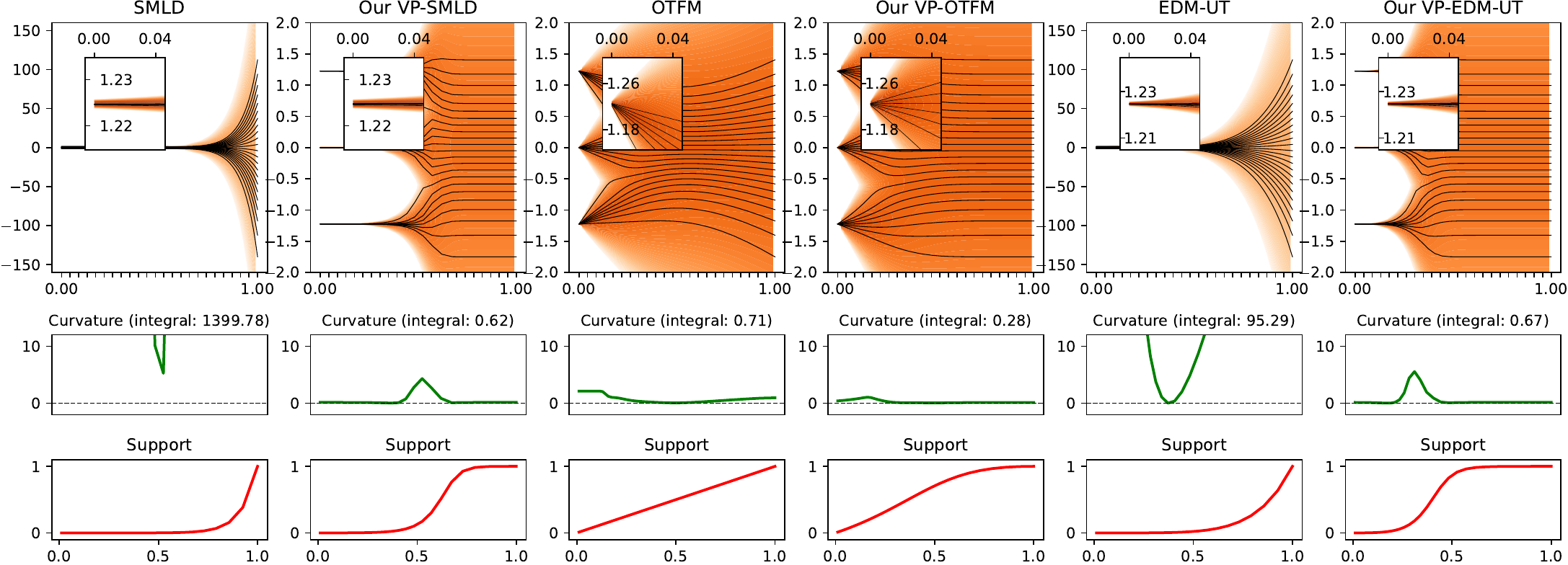}
    \caption{
    ODE trajectories (black solid curves) and the marginal density path (orange shadows) when the data distribution is a mixture of three uniformly spaced delta peaks. The insets focus on the neighborhood of one of the peaks at $x = \sqrt{\frac{3}{2}}$ and $ t \ll 1$, and the green and red curves indicate the local curvature and the support (relative to the one at $t = 1$), respectively. On top of the green curve, we report the global curvature integrated over $t$.
    }
    \label{fig:toy_example_3deltas_ours}
\end{figure*}

Extending the subset of trajectories shown in figure~\ref{fig:overview}, in figure~\ref{fig:toy_example_3deltas_ours}, we show the ODE trajectories for all the noise schedules that were used in section~\ref{sec:Experiments}.

\begin{figure*}[t]
    \centering
    \includegraphics[width=\linewidth]{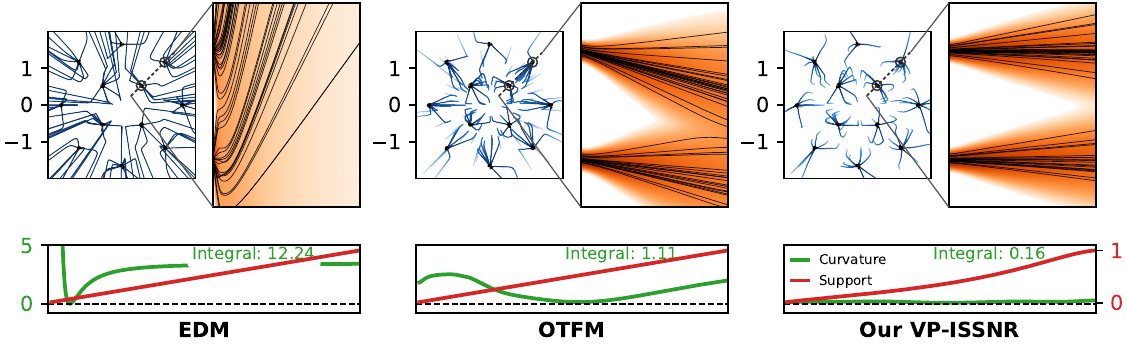}
    \caption{\rebuttal{Two-dimensional version of the toy analysis of figure~\ref{fig:overview}, where
    the target distribution $p_\mathrm{data}$ is a mixture of 12 isotropic Gaussians of equal weight, placed on two concentric rings (4 modes at radius $1$, 8 modes at radius $2.2$, all with standard deviation $0.03$), and rescaled so that the mixture has zero mean and identity covariance.
    \textbf{Left panel of each block:} $96$ trajectories of the reverse ODE (equation~\ref{eq:GenRevSDE} with
    $\lambda = 0$) started from the same prior samples.
    \textbf{Right panel of each block:} The position along the dashed ray, which passes through two picked modes,
    as a function of $t$, over the density of $p_t$ restricted to that ray, i.e. only the trajectories ending in the two circled modes are shown. 
    The trajectories of VP-ISSNR enter the two modes early and along almost straight lines, those of OTFM contract towards the modes over the displayed interval, and those of EDM turn into the modes only at the very end.
    \textbf{Bottom:} Velocity deviation $\mathcal{C}(t)$ (green, equation~\ref{eq:CurvatureMetric}, summed over both dimensions) with its integral over $t$, and the support $b(t)/b(1)$ of the marginal relative to its value at $t=1$ (red), as in figure~\ref{fig:overview}.}}
    \label{fig:toy_example_2d}
\end{figure*}

\rebuttal{While the one-dimensional example of figure~\ref{fig:overview}
makes the trajectories easy to read, Figure~\ref{fig:toy_example_2d} repeats the same analysis but in two dimensions, where the score of the diffused distribution is still available in closed form.}

\subsection{\rebuttal{Geometry relaxations with DFT}}
\label{sec:DFTcalculations}
\rebuttal{All geometry relaxations were performed with ASE\cite{hjorth2017atomic} using the built-in BFGS optimizer with default convergence thresholds. We used PySCF\cite{sun2018pyscf} for performing the DFT calculations and the ASE \texttt{Calculator} class to interface PySCF and ASE. To ensure convergence for all geometries within a reasonable computational budget, we pre-optimized all structures at the B3LYP/STO-3G level and then used those geometries as the starting point for relaxations at the B3LYP/6-31G(2df,p) level (matching the level of theory of the QM9 reference structures the models were trained on). We confirmed that this protocol results in identical final geometries (RMSD < 0.001 \AA) compared to a direct relaxation at the B3LYP/6-31G(2df,p) level for a randomly chosen sample of 100 structures, but at a greatly reduced number of optimization steps with the larger (and more computationally expensive) 6-31G(2df,p) basis set.

RMSD calculations were performed after structural alignment using ASE's \texttt{ase.build.minimize\_rotation\_and\_translation} utility. We note that this utility does \emph{not} consider permutations of chemically equivalent atoms (e.g., the hydrogen atoms of a methyl group) to minimize RMSD, so our reported RMSD values are an upper bound to the true (minimal) RMSD. A manual inspection of 10 optimization trajectories suggests that atoms move little between the initial and final geometries (even for strongly distorted starting structures obtained at low NFEs), so we believe the reported RMSD values to correspond to the minimal RMSD in the vast majority of cases. We report geometric deviations only and do not compare relaxation energies because each generated structure is relaxed into its own nearest local minimum rather than matched to a common reference conformer, energy differences between a generated structure and its relaxed counterpart would mix the geometric error we wish to quantify with the depth of whichever basin the structure happens to fall into, and would make the comparison across structures harder to interpret.}

\end{document}